%% file: main.tex
\documentclass[10pt,twocolumn,letterpaper]{article}

\usepackage[pagenumbers]{cvpr} 

\input{preamble}

\definecolor{cvprblue}{rgb}{0.21,0.49,0.74}
\usepackage[pagebackref,breaklinks,colorlinks,citecolor=cvprblue]{hyperref}

\def\paperID{220} 
\def\confName{3DV\xspace}
\def\confYear{2026\xspace}

\title{Gamma-from-Mono: Road-Relative, Metric, Self-Supervised Monocular Geometry for Vehicular Applications}
\author{
Gasser Elazab$^{1,2}$ \quad
Maximilian Jansen$^{1}$ \quad
Michael Unterreiner$^{1}$ \quad
Olaf Hellwich$^{2}$\\
$^{1}$CARIAD SE \quad
$^{2}$Technische Universit\"at Berlin\\
{\tt\small gasser.elazab@cariad.technology}
}

\begin{document}
\maketitle
\input{sec/0_abstract}
\input{sec/1_intro}
\input{sec/2_formatting}

\input{sec/3_finalcopy}
{
    \small
    \bibliographystyle{ieeenat_fullname}
    \bibliography{main}
}
\input{sec/X_suppl}

\end{document}

%% file: preamble.tex
\usepackage[dvipsnames]{xcolor}

\usepackage[utf8]{inputenc} 
\DeclareUnicodeCharacter{202F}{\,} 

\usepackage{array,booktabs,subcaption}
\usepackage[margin=1in]{geometry} 
\usepackage{graphicx}
\usepackage{array,booktabs}
\usepackage{multirow}
\usepackage{tabularx}
\usepackage[table]{xcolor}
\usepackage{pifont}
\usepackage{cmap}
\usepackage{transparent}  
\usepackage{subcaption} 
\usepackage{float}

\definecolor{refgray}{HTML}{F2F2F2} 

\usepackage{pgfplots}
\pgfplotsset{compat=1.18}  

\usepackage{amsmath}
\usepackage{amssymb}
\usetikzlibrary{arrows.meta}

\usepackage{enumitem}


%% file: sec/0_abstract.tex
\begin{abstract}
Accurate perception of the vehicle’s 3D surroundings, including fine-scale road geometry, such as bumps, slopes, and surface irregularities, is essential for safe and comfortable vehicle control. However, conventional monocular depth estimation often oversmooths these features, losing critical information for motion planning and stability. To address this, we introduce Gamma-from-Mono (GfM), a lightweight monocular geometry estimation method that resolves the projective ambiguity in single-camera reconstruction by decoupling global and local structure. GfM predicts a dominant road surface plane together with residual variations expressed by $\gamma$, a dimensionless measure of vertical deviation from the plane, defined as the ratio of a point’s height above it to its depth from the camera, and grounded in established planar parallax geometry. With only the camera’s height above ground, this representation deterministically recovers metric depth via a closed form, avoiding full extrinsic calibration and naturally prioritizing near-road detail. Its physically interpretable formulation makes it well suited for self-supervised learning, eliminating the need for large annotated datasets. Evaluated on KITTI and the Road Surface Reconstruction Dataset (RSRD), GfM achieves state-of-the-art near-field accuracy in both depth and $\gamma$ estimation while maintaining competitive global depth performance. Our lightweight 8.88M-parameter model adapts robustly across diverse camera setups and, to our knowledge, is the first self-supervised monocular approach evaluated on RSRD.

\end{abstract}

%% file: sec/1_intro.tex
\section{Introduction}
\label{sec:introduction}

\begin{figure}[t]
    \centering
    \begin{subfigure}[b]{\linewidth}
        \centering
        \includegraphics[width=\linewidth]{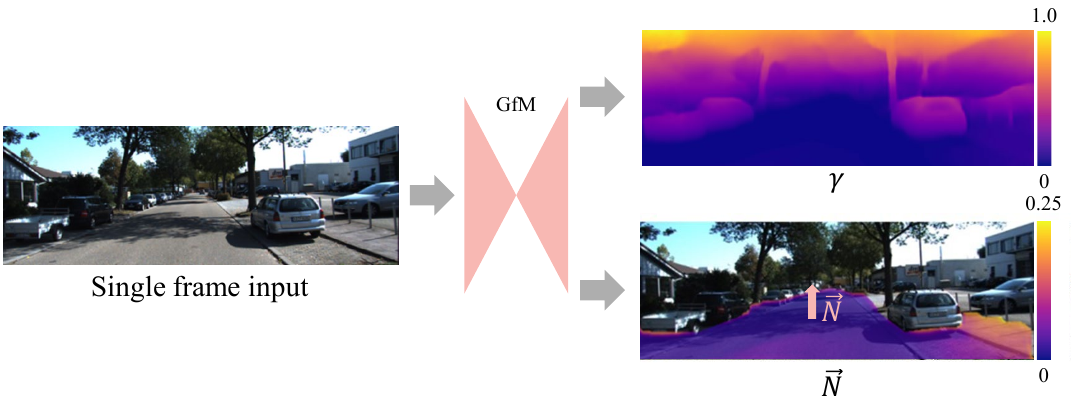}
        \caption{Model overview}
        \label{fig:first_fig}
    \end{subfigure}
    
    \begin{subfigure}[b]{0.45\linewidth}
        \centering
        \includegraphics[scale=0.6]{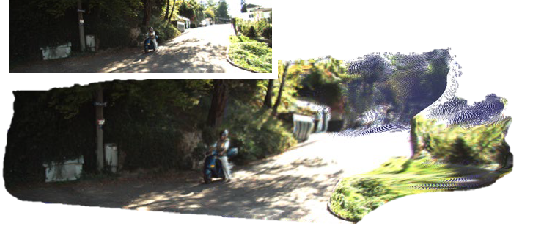}
        \caption{Point cloud (KITTI)}
        \label{fig:second_fig}
    \end{subfigure}
    \begin{subfigure}[b]{0.45\linewidth}
        \centering
        \includegraphics[scale=0.63]{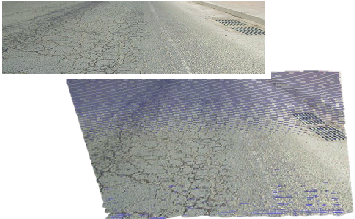}
        \caption{Point cloud (RSRD)}
        \label{fig:third_fig}
    \end{subfigure}
    
    \caption{
    Overview of our self-supervised model.  
    (a) Predicts $\gamma$ (height/depth ratio) and road normal $\vec{N}_{\text{pred}}$. Height visualizations are clipped from 0 to 0.25\,m. 
    (b, c) 3D point clouds from KITTI~\cite{geiger2012kitti} and RSRD~\cite{zhao2023rsrd}, with input images in the top-left.
    }
    \label{fig:combined_fig}
\end{figure}

    



    

Modern robotics and autonomous vehicles demand scalable 3D perception, but achieving it with a single camera remains challenging~\cite{Midas_2022_main}. Monocular Geometry Estimation (MGE) reconstructs per-pixel 3D structure from a single image, enabling cost-effective 3D perception without specialized sensors \cite{afshar2023efficient}. Specifically, accurate 3D reconstruction of the near-road geometry is crucial for navigation in autonomous driving~\cite{TrajectoryTrainning_bumps,AdaptivesuspensionForRoad}, supporting obstacle avoidance and motion planning~\cite{BEV-LANENET-CVPR2023}, and it is also vital for off-road and legged robotics, where elevation and local slope guide traversability and footstep planning~\cite{miki2022elevation}. However, monocular depth estimation methods, which are commonly used in these domains, struggle to accurately capture road topography~\cite{zhao2024roadbev}. Textureless pavements and low-contrast surfaces often lead to oversmoothing and underestimation of slopes, causing small obstacles and surface irregularities to be missed~\cite{yang2023bevheight,zhao2024roadbev,abdusalomov2025breaking}. This limitation is critical, as small height variations, such as bumps or road-level changes, can differentiate drivable regions from hazards and negatively impact vehicle dynamics and safety~\cite{shetty2022slope,pinggera2016lost,AdaptivesuspensionForRoad}.

Metric foundation depth models~\cite{DepthPro2024,depthanythingv2,yin2023metric3d,unidepth1,unidepthv2} generalize well across domains and provide strong scene-level predictions. However, they do not explicitly target road-relative quantities such as height above the ground or local slope. Moreover, they often suffer from residual scale drift when deployed in new environments, typically requiring inference-time scale correction. In contrast, self-supervised monocular methods~\cite{packnet_selfsup,CameraHeightDoesnChange,MonoPP,roadaware_SFM_2021_selfsup_scaled,VADEPTH,GroCo2024,wagstaff_scalerecovery_2021_SELFSUP,dynadepth} attempt to resolve scale ambiguity by incorporating metric anchors such as odometry sensors or ground-plane constraints. While these strategies stabilize global scale, they remain depth-centric, leaving road geometry underconstrained.

\begin{figure}[h]
\centering
\begin{tikzpicture}[scale=1.0,line cap=round,line join=round,
  every node/.style={font=\footnotesize}]

  \def\f{1.7}    
  \def\hc{1.5}   
  \def\d{2.8}    
  \def\h{0.6}    
  \def\xmax{6.0} 

  \coordinate (C)  at (0,0);                         
  \coordinate (G0) at (-0.2,{-\hc});                 
  \coordinate (G1) at (\xmax,{-\hc});                

  \coordinate (P1)   at ({\d},{\h-\hc});             
  \coordinate (Pg1)  at ({\d},{-\hc});               
  \coordinate (P2)   at ({2*\d},{2*\h-\hc});         
  \coordinate (Pg2)  at ({2*\d},{-\hc});             

  \coordinate (I1top)   at ({\f},{\f*(\h-\hc)/\d});        
  \coordinate (I1plane) at ({\f},{\f*(-\hc)/\d});          
  \coordinate (I2top)   at ({\f},{\f*(2*\h-\hc)/(2*\d)});  
  \coordinate (I2plane) at ({\f},{\f*(-\hc)/(2*\d)});      

  \draw[thick] (G0) -- (G1) node[anchor=west]{plane};
  \draw[rounded corners=2pt,fill=black!8] (-0.36,-0.22) rectangle (0.28,0.22);
  \draw[fill=black!15] (0.52,-0.12) -- (0.52,0.12) -- (0.2,0) -- cycle;
  \draw[fill=black!30] (-0.13,0) circle (0.07);
  \fill (C) circle (1.4pt);
  \node[above left=4pt] at (C) {$C$};

  \draw[<->] (0,0) -- (0,{-\hc});
  \node[left] at (0,{-0.5*\hc}) {$h_c$};


  \draw[very thick,blue] (\f,{-\hc-0.25}) -- (\f,{0.5});
  \draw[blue,->] (0,{-\hc+0.35}) -- (\f,{-\hc+0.35}) node[midway,above] {$f$};

  \draw[red,thick] (C) -- (P1);
  \draw[red,thick] (C) -- (P2);

  \draw[red,thick] (C) -- (Pg1);
  \draw[red,thick] (C) -- (Pg2);

  \fill[red] (P1) circle (1.4pt);
  \node[above right] at (P1) {$P_1(d,h)$};
  
  \fill[red] (P2) circle (1.4pt);
  \node[above right] at (P2) {$P_2(2d,2h)$};

  \draw[gray] (Pg1) -- (P1);  \node[anchor=west] at ({\d+0.08},{-\hc+0.45*\h}) {$h$};
  \draw[gray] (Pg2) -- (P2);  \node[anchor=west] at ({2*\d+0.08},{-\hc+\h}) {$2h$};
  \node[below] at (\d,{-\hc}) {$d$};
  \node[below] at ({2*\d},{-\hc}) {$2d$};

  \fill[black] (I1plane) circle (1.5pt);
  \fill[black] (I2plane) circle (1.5pt);
  \fill[red] (I1top) circle (1.7pt);
  \fill[red] (I2top) circle (1.7pt);

  \draw[gray,densely dashed] (\f,{-\hc-0.25}) -- (\f,{0.5});

  \draw[<->,gray] ({\f+0.15},{\f*(-\hc)/\d-0.05})
                  -- ({\f+0.15},{\f*(\h-\hc)/\d-0.05})
      node[midway,right,scale=0.7] {$\Delta v$};
  \draw[<->,gray] ({\f+0.15},{\f*(-\hc)/(2*\d)-0.05})
                  -- ({\f+0.15},{\f*(2*\h-\hc)/(2*\d)-0.05})
      node[midway,right,scale=0.7] {$\Delta v'$};

\end{tikzpicture}
\caption{For points $(d,h)$ and $(2d,2h)$, $\Delta v=f\,h/d$ equals $\Delta v'$, the gap depends only on $h/d$. A known ground reference fixes metric scale, resolving monocular projective ambiguity.}


\label{fig:plane_heights_deltas_single_column}
\end{figure}
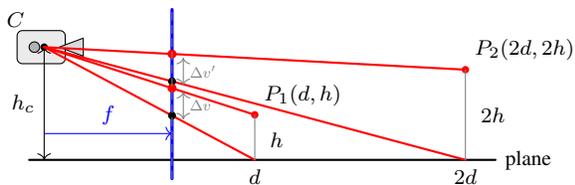

In practice, existing monocular pipelines recover height above the road indirectly via costly post-processing, such as elevation maps~\cite{miki2022elevation}. Recent top-down (BEV) approaches~\cite{zhao2024roadbev} model road surfaces explicitly but rely on discretized ground-plane grids and dense ground-truth supervision, limiting resolution and scalability in unlabeled settings. Since single-view depth is projectively ambiguous~\cite{roussel2019monocular,depthanything}, a road-relative height-to-depth ratio $\gamma = h/d$ offers a complementary representation, making $\gamma$ dimensionless and tied to the ground. As shown in~\cref{fig:plane_heights_deltas_single_column}, doubling the scale of an object relative to the ground leaves the apparent vertical offset unchanged. From a single image the metric configuration is indistinguishable, so depth remains scale ambiguous, whereas in $\gamma$ space, both configurations take the same dimensionless value, consistently tied to the road. With a known ground plane and camera height, this $\gamma$ value converts to absolute height and depth in a simple closed-form way.


Accordingly, we introduce Gamma from Mono (\textbf{GfM}), a single-frame approach that reframes monocular geometry around the dominant road plane, mitigating projective ambiguity and reducing scale to a single camera-height parameter. Our key contributions are:

\begin{itemize}

\item We propose a model that directly predicts a road-relative representation, comprising a global road-plane normal and a per-pixel height-to-depth ratio ($\gamma$), preserving near-road detail.

\item To our knowledge, this is the first single-frame, self-supervised method that directly regresses $\gamma$ for road-relative geometry, enabling explicit, interpretable estimates of road topography.

\item We resolve metric scale from a dimensionless prediction, converting to metric depth using only known camera height and avoiding full extrinsic calibration or test-time fitting.

\end{itemize}

\section{Related Work}

Most prior work in monocular geometry estimation predicts per-pixel depth~\cite{masoumian2022monocular_areview}. 
By contrast, the height-to-depth ratio $\gamma$ has been a key parameter for multi-view reconstruction via planar parallax~\cite{irani1996parallax,PP_foundation_94_1,sawhney19943d}. 
For example, MonoPP~\cite{MonoPP} uses $\gamma$ only as a training-time scale cue distilled from multi-frame planar-parallax constraints. On the other hand, Yuan~\etal~\cite{RoadPlanarParallax} computes per-pixel $\gamma$ from multi-frame homography alignment with LiDAR supervision. Beyond these, monocular methods remain depth-centric. Both approaches rely on multi-view cues or explicit ground-truth depth, and neither regresses $\gamma$ directly from a single image in a self-supervised setting.

\paragraph{Scale ambiguity.} Monocular depth estimation (MDE) has advanced greatly, yet models trained only on monocular images recover depth only up to an unknown scale factor~\cite{depthanything,wang2024moge}. However, metric-scale depth is crucial for autonomous driving safety~\cite{wang2021research_lidarvscamera}. Recent foundational models such as Metric3D~\cite{yin2023metric3d}, UniDepth~\cite{unidepth1,unidepthv2}, MoGe~\cite{wang2024moge,wang2025moge2}, DepthAnything~\cite{depthanything,depthanythingv2}, and DepthPro~\cite{DepthPro2024} include metric-depth variants, yet still exhibit small but persistent scale drift in novel environments. Moreover, these methods are trained on millions of images and use large transformer backbones, making them less suitable for resource-constrained real-time deployment on limited hardware. Traditional solutions recover scale through post-processing, such as fitting a known ground plane, or by integrating explicit cues such as vehicle speed, IMU data during training.

\paragraph{Motion-based supervision.}
Motion cues from vehicle sensors offer effective scale recovery for monocular depth. Velocity from odometry or GPS can directly constrain scale by aligning estimated motion with true displacement. Guizilini~\etal~\cite{packnet_selfsup} (PackNet) achieved metric-scaled depth by incorporating odometry-derived travel distances during training, supervising translation magnitude to align predicted scene motion with real-world displacement and reduce scale drift. Similarly, Zhang~\etal~\cite{dynadepth} leveraged IMU-derived gravity and acceleration measurements to enforce vertical and horizontal consistency constraints, improving metric accuracy and robustness across scenes. Wagstaff and Kelly~\cite{wagstaff_scalerecovery_2021_SELFSUP} extended this idea using full inter-frame pose supervision. These approaches demonstrate how motion-based cues, whether velocity, IMU, or pose, can yield scale-consistent monocular depth.

\paragraph{Ground plane constraints.}
Camera extrinsics with respect to a flat ground plane provide a valuable scale cue, widely used in autonomous driving. Sui~\etal~\cite{roadaware_SFM_2021_selfsup_scaled} enforced a ground-plane constraint to align predicted depths with known geometry, but faced accuracy limitations due to calibration errors and road irregularities. VADepth~\cite{VADEPTH} improved robustness by using attention-based losses on road pixels. In addition, GroCo~\cite{GroCo2024} further introduced a network to detect the road plane and constrain its depth to calibrated metric values. This improves generalization to camera rotations but still requires precise extrinsics during both training and inference. Moreover, MonoPP~\cite{MonoPP} leveraged planar-parallax geometry in a multi-frame teacher-student framework, achieving state-of-the-art results but requiring computationally intensive training and accurate extrinsics. FUMET~\cite{CameraHeightDoesnChange} recovered metric scale by first estimating camera height from the road plane. They then scale it with a per-frame factor from a learned vehicle-size prior (LSP) via silhouette–size matching. Training enforces sequence-wise camera-height invariance. However, reliance on the LSP (trained on DVM-CAR~\cite{huang2022dvmcar_dataset}) and segmentation quality can potentially bias scale estimates.

\begin{figure}[h]
  \centering
   \includegraphics[width = 1.0 \linewidth]{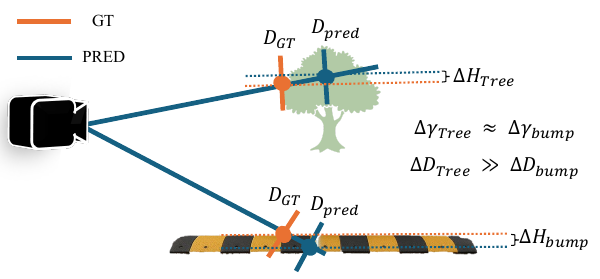}

   \caption{Smoothing a small bump causes negligible depth error compared to a tree, but in $\gamma$-space the errors are similar, revealing sensitivity to small height changes. A numerical example is provided in the supplementary material~\cref{subsec:gamma_vs_depth}.}
   \label{fig:gamma_vs_depth_main_paper}
\end{figure}

\paragraph{Summary and motivation.} 
Monocular geometry estimation faces scale ambiguity, often addressed with external sensors or ground-plane assumptions, though these can be sensitive to calibration and scene geometry. In addition, most such approaches focus only on depth, limiting their ability to reconstruct fine road topography. Inspired by ground-plane priors~\cite{MonoPP,GroCo2024}, we propose a height-sensitive monocular geometry estimation framework that directly predicts the ratio gamma from a single image. In contrast to MonoPP~\cite{MonoPP}, which derives $\gamma$ from two-view optical flow and is therefore view-dependent and prone to alignment errors, our method treats the ratio as a primary prediction target. This representation models extrinsic variations~\cite{irani1996parallax,sawhney19943d}, reduces calibration requirements, and encodes road-relative elevation, as illustrated in~\cref{fig:gamma_vs_depth_main_paper}. Such height-aware perception benefits automotive tasks including slope estimation, elevation-change detection, road-damage identification, and obstacle recognition~\cite{shetty2022slope,abdusalomov2025breaking}.

\begin{figure*}
  \centering
  \includegraphics[width=1\linewidth]{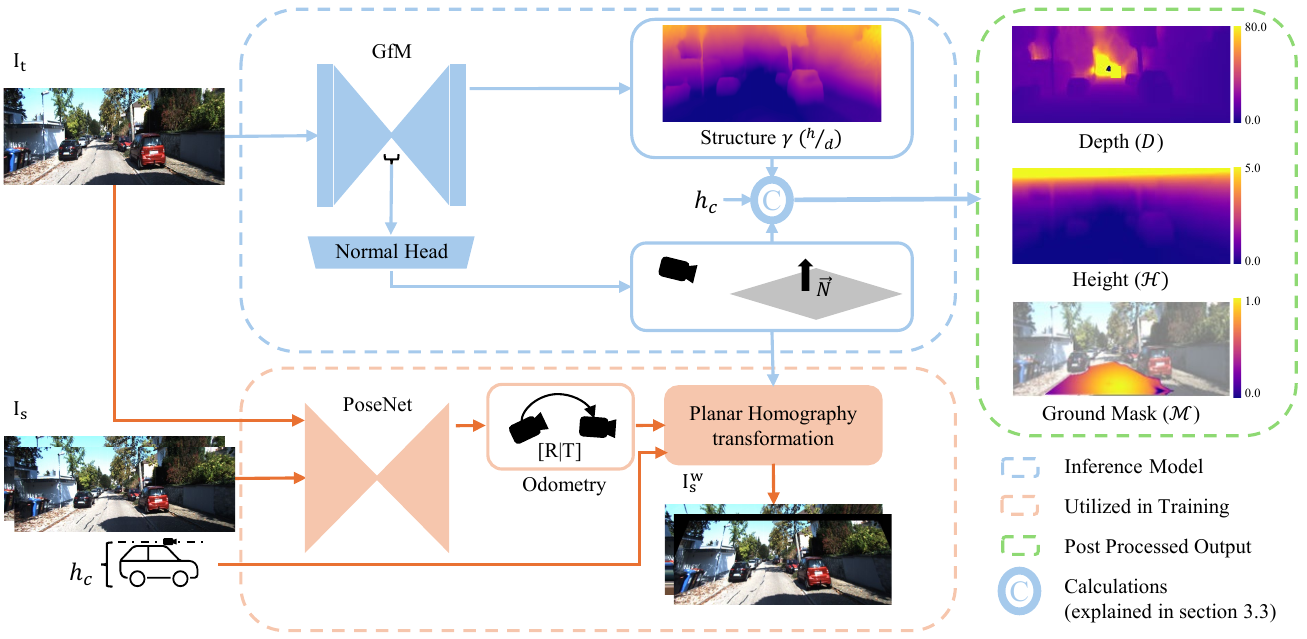}
  \caption{\textbf{Overview of our model architecture.} 
        The main network predicts a per-pixel parameter $\gamma$, while PoseNet 
        estimates the relative pose between $I_{s}$ and $I_{t}$. 
        From the relative pose, we compute a planar homography to align 
        the planar road surface between the two images. 
        Additionally, $\gamma$ is used to infer depth, scene height, 
        and a probabilistic road mask, as detailed in~\cref{sec:postprocessing}.}
  \label{fig:main_fig}
\end{figure*}

\section{Method}
\label{sec:method}

Our method identifies the dominant road plane and estimates \(\gamma\), the per-pixel height above the road plane divided by depth from the camera, in a fully self-supervised manner. In our experiments, direct height prediction in self-supervised training collapsed due to its unconstrained scale. Predicting $\gamma$ instead couples height to depth, yielding a stable, scale-consistent value and sharper close-range geometry, as later shown in~\cref{tab:main_tab}. We show in the inference model in~\cref{fig:main_fig}, how our approach naturally decomposes the scene reconstruction into two interpretable components: a global road plane, described by the normal vector $\vec{N}_{\text{pred}}$, capturing the dominant road geometry and a local per-pixel structure $\gamma$  modeling residual fine-grained variations~\cite{MonoPP,RoadPlanarParallax,irani1996parallax,sawhney19943d}. Consequently, our approach yields a scene-centric geometric representation that effectively captures slopes and recovers uneven terrain with high precision. Our method further uses camera height and intrinsics to recover metric geometry in a simple and fast post-processing step.

\subsection{Problem Setup}
\label{subsec:problem_setup}
Given a monocular target image \(I_t\), the network \(\theta_g\) predicts per-pixel \(\gamma_t(u,v)\) and a single global road-plane normal \(\vec{N}_{\text{pred}}\in\mathbb{R}^3\):
\[
\gamma_t,\,\vec{N}_{\text{pred}} = \theta_g(I_t).
\]
By definition,
\[
\gamma_t(u,v)=\frac{h_t(u,v)}{d_t(u,v)},
\]
where \(d_t(u,v)\) is depth from the camera and \(h_t(u,v)\) is height relative to the road plane. We adopt the convention that \(\vec{N}_{\text{pred}}\) points upward, so heights are measured along \(\vec{N}_{\text{pred}}\) and are positive above the road surface.


For training, we employ a self-supervised approach using neighboring source images $I_s=\{I_{t-1}, I_{t+1}\}$. A Pose Network $\theta_p$ estimates the 6-DoF relative pose $T_{t\to s}$ between $I_t$ and $I_s$ via $T_{t\to s}=\theta_p(I_t,I_s)$. Following~\cite{MONODEPTH_Zhou}, this enables self-supervised learning via novel-view synthesis. If accurate poses from other sensors are available, the pose network can be omitted.

\subsection{Network Architecture}
\label{subsec:network_archi}
GfM is based on LiteMono-8M~\cite{litemono_2023_selfsup} and predicts $\gamma_t$ and $\vec{N}_{\text{pred}}$. The final layer applies a sigmoid and maps its output to a predefined $\gamma$ range using a signed log-space transform that increases precision near zero.~\Cref{eqn:first_gamma_transform} defines the forward transform from $\gamma$ to its log-space representation $\tilde{\gamma}$:
\begin{equation}
    \label{eqn:first_gamma_transform}\tilde{\gamma}=\operatorname{sign}(\gamma)\,\log\!\big(1+\alpha|\gamma|\big),
\end{equation}
where $\alpha>0$ controls the nonlinearity. Apply the same transform to the range endpoints to get $\tilde{\gamma}_{\min}$ and $\tilde{\gamma}_{\max}$. Given the sigmoid output $\sigma_t\in[0,1]$, map it linearly in the transform domain:
\begin{equation}
    \tilde{\gamma}_t=\tilde{\gamma}_{\min}+(\tilde{\gamma}_{\max}-\tilde{\gamma}_{\min})\,\sigma_t.
\end{equation}
Finally, invert the transform to obtain $\gamma_t$:
\begin{equation}
    \gamma_t=\operatorname{sign}(\tilde{\gamma}_t)\,\frac{\exp\!\big(|\tilde{\gamma}_t|\big)-1}{\alpha}.
\end{equation}

To model global scene geometry, we add a lightweight global normal head that predicts a single road-plane normal $\vec{N}_{\text{pred}}$.
It applies attention pooling to the bottleneck feature map $x \in \mathbb{R}^{B \times C \times H \times W}$ to extract a road-focused descriptor, followed by a linear layer and $\ell_2$-normalization to produce a unit vector. Further details are given in the supplementary (\cref{sec_app:global_road_normal_prediction}).
\begin{equation}
    \vec{N}_{\text{pred}} \;=\; 
    \frac{\mathrm{Linear}\!\big(\mathrm{AttnPool}(x)\big)}
         {\big\|\mathrm{Linear}\!\big(\mathrm{AttnPool}(x)\big)\big\|_2}.
\end{equation}
\paragraph{Pose network.}  
Following~\cite{MONODEPTH2,manydepth,litemono_2023_selfsup}, we use a ResNet-18 backbone for pose estimation, taking two concatenated RGB frames (6 channels) as input.


\subsection{Post-processing Network Outputs}
\label{sec:postprocessing}

Using a simple post-processing step, our network outputs are transformed to interpretable representations: metric depth, residual parallax flow, and a road mask for subsequent loss computation and analysis.

\paragraph{Recovering depth from \(\gamma\).}
Given \(\gamma_t(u,v)\), we compute the metric depth \(d_t(u,v)\)~\cite{RoadPlanarParallax} as:
\begin{equation}
\label{eqn:depth_from_gamma}
d_{t}(u,v) 
= 
\frac{h_{c}}
{\gamma_{t}(u,v) + {\vec{N}}_{\text{pred}} \cdot \left(K^{-1} \mathbf{p}(u,v)\right)},
\end{equation}
with \(\mathbf{p}(u,v) = [u,\;v,\;1]^\top\). The height relative to the road follows directly as \(h_t(u,v) = \gamma_t(u,v) \cdot d_t(u,v)\).

\paragraph{Road-plane homography.}  
Assuming the road is the dominant planar surface, we compute the homography between source and target views using the road planar homography computation in~\cite{MonoPP,RoadPlanarParallax}. Warping \(I_s\) with \(H_{s \to t}\) yields the aligned image \(I_s^w\), where the ground plane should ideally coincide with that of \(I_t\).


\paragraph{Residual parallax flow.}
Points not lying on the dominant plane do not align under the homography. Their resulting misalignment, as explained in~\cite{RoadPlanarParallax,MonoPP}, known as the planar parallax residual flow \(\mathbf{u}_{s \to t}^{\text{res}}\)~\cite{PP_foundation_94_1}, is computed directly from \(\gamma_t\):
\begin{equation}
\label{eqn:residual_flow}
u_{s \to t}^{\text{res}} 
= 
\frac{-\gamma_t \,\tfrac{T_z}{h_c}}
{1 - \gamma_t \,\tfrac{T_z}{h_c}} 
\;\bigl(p_t - e_t\bigr),
\end{equation}
where $T_z\neq 0$ is the translation along the $z$-axis, $p_t$ the pixel coordinate, and $e_t$ the epipole in $I_t$.

\paragraph{Probabilistic road mask.}
\label{paragraph:roadmask}
We compute a probabilistic road mask by comparing the predicted global normal \(\vec{N}_{\text{pred}}\) with local surface normals \(\mathbf{n}(u,v)\), obtained by backprojecting depth~\cite{MonoPP,DNet}. To prevent unintended interactions with the loss, we detach the depth before computing \(\mathbf{n}(u,v)\), ensuring that the road mask remains unaffected during optimization. We then measure the angular deviation between \(\mathbf{n}(u,v)\) and \(\vec{N}_{\text{pred}}\):

\begin{equation}
\theta(u,v)\;=\;\cos^{-1}\!\left(
\frac{\mathbf{n}(u,v)\cdot\vec{N}_{\text{pred}}}
     {\|\mathbf{n}(u,v)\|\;\|\vec{N}_{\text{pred}}\|}
\right).
\end{equation}
We convert this deviation into an angle-based probability map $p_{\text{angle}}(u,v)\in[0,1]$. We then multiply $p_{\text{angle}}$ by a Gaussian spatial prior $g(u,v)$. The prior is centered horizontally, and its vertical mean is placed at $0.25$ of the image height measured from the bottom edge to emphasize the near-field road region.

\begin{equation}
\mathcal{M}_{\text{road}}(u,v) \;=\; p_{\text{angle}}(u,v)\cdot g(u,v).
\end{equation}
This weighting improves robustness to uphill/downhill road slopes and other geometric changes. An illustrative example is in \cref{fig:main_fig}, with further details in the supplementary material (\cref{sec:road_mask_appendix}).

\paragraph{Target view synthesis.}
To reconstruct the target view $I_t$ from the source $I_s$, we synthesize two images: $\hat{I}^{d}_{s\to t}$ by depth-based warping~\cite{litemono_2023_selfsup,MONODEPTH2,GroCo2024}, and $\hat{I}^{f}_{s\to t}$ by planar homography plus residual flow~\cite{MonoPP}. 
First, we warp $I_s$ with the homography $H_{s\to t}$:
\begin{equation}
  I_s^{w}=\big\langle I_s,\ \text{warp}(H_{s\to t})\big\rangle,
  \label{eqn:homography_warp}
\end{equation}
then warp off-plane regions with a per-pixel residual flow $\mathbf{u}^{\text{res}}_{s\to t}$:
\begin{equation}
  \hat{I}^{f}_{s\to t}=\big\langle I_s^{w},\ \text{warp}(\mathbf{u}^{\text{res}}_{s\to t})\big\rangle,
  \label{eqn:synthesize_ppflow}
\end{equation}
where $\text{warp}(\cdot)$ denotes the sampling grid and $\langle I,\cdot\rangle$ bilinear resampling~\cite{billinear_sampling}.


\subsection{Losses}
\label{subsec:losses}
All pixel-level losses are averaged over valid pixels.

\paragraph{(1) Photometric loss.}
We measure photometric consistency between synthesized views (\(\hat{I}_{s \to t}^d\), \(\hat{I}_{s \to t}^f\)) and the target view \(I_t\) using photometric error (pe)~\cite{MONODEPTH2}:
\begin{equation}
\mathrm{pe}\bigl(\hat{I}_t, I_t\bigr) 
= 
\alpha\,\frac{1 - \mathrm{SSIM}(\hat{I}_t, I_t)}{2}
+ 
(1 - \alpha)\,\bigl\lVert \hat{I}_t - I_t\bigr\rVert_1,
\end{equation}
where \(\alpha\) balances SSIM~\cite{SSIM2004} and \(L_1\). Following Monodepth2~\cite{MONODEPTH2}, we employ minimum reprojection across source images and use an implicit auto-masking strategy~\cite{MONODEPTH2} to mitigate occlusions and dynamic objects:
\begin{equation}
\mathcal{L}_{\text{photo}} 
=  \mathcal{L}^d + \mathcal{L}^f = 
\min_{s}\mathrm{pe}\bigl(\hat{I}^{d}_{s \to t}, I_t\bigr) 
+ 
\min_{s}\mathrm{pe}\bigl(\hat{I}^{f}_{s \to t}, I_t\bigr).
\end{equation}

While the first photometric loss $\mathcal{L}^d$, which compares warped images synthesized using depth, ensures consistency on large scales, the second term $\mathcal{L}^f$ supervises fine-grained structures near the road surface.

\begin{table*}[t!]
\centering
\footnotesize
\setlength\tabcolsep{3pt}
\renewcommand{\arraystretch}{1}

\definecolor{kittiColor}{HTML}{FFF5E6}
\definecolor{rsrdColor}{HTML}{E6F7FF}
\newcommand{\kittiTag}{\textcolor{kittiColor}{\large\textbullet}}
\newcommand{\rsrdTag}{\textcolor{rsrdColor}{\large\textbullet}}

\begin{subtable}{\textwidth}
\centering
\caption{KITTI~\cite{geiger2012kitti,IMPROVED_KITTI_GT}}
\label{tab:main_tab:kitti}

\begin{tabularx}{\textwidth}{p{0.3cm}|p{2 cm} p{3cm} c c c c c c c c c }
\toprule
 & Published in & Method & Train & \#Params & Abs Rel $\downarrow$ & Sq Rel $\downarrow$ & RMSE $\downarrow$ & RMSE log $\downarrow$ & $\delta_1$ $\uparrow$ & $\delta_2$ $\uparrow$ & $\delta_3$ $\uparrow$ \\
 \midrule
\multirow{6}{*}{\rotatebox[origin=c]{90}{{Depth}}} & WACV 2025 & MonoPP~\cite{MonoPP}  &  M+camH & 34.57M & \underline{0.089}& 0.545  & 3.864  & \textbf{0.134}  & 0.913  & \underline{0.983}  & \underline{0.995}  \\
& RA-L 2022 & VADepth~\cite{VADEPTH}  &  M+camH & 18.8M & 0.091  & 0.555 & 3.871 &\textbf{0.134} & \underline{0.913}  & \underline{0.983} & \underline{0.995} \\
& ECCV 2024 & GroCo~\cite{GroCo2024}  &  M+camH & 34.65M & \underline{0.089} & \underline{0.517} & 3.815 &\textbf{0.134} & 0.910 & \textbf{0.984} & \underline{0.995} \\
& arXiv 2025 & UniDepthV2-L~\cite{unidepthv2}  &  F & 353.22M & 0.123  & 0.576  & \textbf{3.612}  & \underline{0.139}  & \textbf{0.933}  &  {0.991}  & \textbf{0.997} \\
& arXiv 2025 & MoGe2-L~\cite{wang2025moge2}  &  F & 330.90M & 0.180  & 0.732  & 4.122  & 0.217  & 0.633  &  0.969  & 0.994 \\
& ICLR 2025 & DepthPro~\cite{DepthPro2024}  &  F & 951.99M & 0.117  & \textbf{0.502} & \underline{3.721}  & 0.149  & 0.874  &  0.979  & \underline{0.995} \\
& ICLR 2025 & DepthPro~\cite{DepthPro2024}~\textdagger  &  F & 951.99M & 0.072  & 0.349 & 3.489  & 0.112  & 0.941  &  0.990  & 0.998 \\
& - & GfM (ours)  &  M+camH & \textbf{8.88M} & \textbf{0.085} & 0.554 & 4.082 & \underline{0.139} & \underline{0.916} & \underline{0.983} & \underline{0.995} \\
\midrule
\multirow{5}{*}{\rotatebox[origin=c]{90}{{Depth $\leq 20$}}} 
& WACV 2025 & MonoPP~\cite{MonoPP}  &  M+camH & 34.57M & \underline{0.067} & \underline{0.121} & \underline{1.109} & \underline{0.093} & {0.957} & \underline{0.994} & \textbf{0.999} \\
& ECCV 2024 & GroCo~\cite{GroCo2024}  &  M+camH & 34.65M & \underline{0.067} & 0.125 & 1.141 & 0.094 & 0.955 & 0.995 & \textbf{0.999} \\
& arXiv 2025 & UniDepthV2-L~\cite{unidepthv2}  &  F & 353.22M & 0.105  & 0.188  & 1.383  & 0.114  & \underline{0.966}  &  \textbf{0.996}  & \textbf{0.999} \\
& ICLR 2025 & DepthPro~\cite{DepthPro2024}  &  F & 951.99M & 0.105 & 0.202 & 1.362 & 0.122 & 0.913 & 0.990 & \underline{0.997} \\
& ICLR 2025 & DepthPro~\cite{DepthPro2024}~\textdagger  &  F & 951.99M & {0.051} & {0.071} & {0.888} & {0.073} & {0.982} & {0.998} & 0.999 \\
& - & GfM (ours)  &  M+camH & \textbf{8.88M} & \textbf{0.058} & \textbf{0.103} & \textbf{1.036} & \textbf{0.085} & \textbf{0.967} & \textbf{0.996} & \textbf{0.999} \\
\bottomrule
\end{tabularx}

\begin{tabularx}{\textwidth}{p{0.3cm}|p{2 cm} p{3cm} c c c c c c c c c }
\toprule
 & Published in & Method & Train & \#Params & PP & Abs Diff $\downarrow$  & RMSE $\downarrow$ & RMSE log $\downarrow$ & $\delta_1$ $\uparrow$ & $\delta_2$ $\uparrow$ & $\delta_3$ $\uparrow$ \\
\midrule
\multirow{4}{*}{\rotatebox[origin=c]{90}{{Gamma}}} 
& ECCV 2024 & GroCo~\cite{GroCo2024}  &  M+camH & 34.65M & \checkmark  & \textbf{0.012}  & \textbf{0.018} & \underline{0.840} & \underline{0.66} & \underline{0.768} & \textbf{0.822} \\
& ICLR 2025 & DepthPro~\cite{DepthPro2024}  &  F& 951.99M & \checkmark  & \underline{0.014}   & \underline{0.020}  & 0.885  & 0.655  &  0.748  & 0.797 \\
& ICLR 2025 & DepthPro~\cite{DepthPro2024}~\textdagger  &  F & 951.99M & \checkmark  & 0.011   & 0.016  & 0.718  & 0.726  &  0.811  & 0.854 \\

& - & GfM (ours)  &  M+camH & \textbf{8.88M}& -  & \textbf{0.012}  & \textbf{0.018} & \textbf{0.801} & \textbf{0.706} & \textbf{0.780} & \underline{0.820} \\
\bottomrule
\end{tabularx}
\end{subtable}

\begin{subtable}{\textwidth}
\centering
\caption{RSRD~\cite{zhao2023rsrd}}
\label{tab:main_tab:rsrd}

\begin{tabularx}{\textwidth}{p{0.3cm}|p{2 cm} p{3cm} c c c c c c c c c }
\toprule
 & Published in & Method & Train & \#Params & Abs Rel $\downarrow$ & Sq Rel $\downarrow$ & RMSE $\downarrow$ & RMSE log $\downarrow$ & $\delta_1$ $\uparrow$ & $\delta_2$ $\uparrow$ & $\delta_3$ $\uparrow$ \\
\midrule
\multirow{8}{*}{\cellcolor{white}\rotatebox[origin=c]{90}{{Depth}}}
& - & Monodepth2~\cite{MONODEPTH2}~$\ddagger$  &  M+Pose & 14.3M & \underline{0.059}  & \underline{0.032} &  \underline{0.393} & \underline{0.076}  & \underline{0.987}  &  \textbf{0.996}  &  \textbf{0.999}  \\
& ICLR 2025 & DepthPro~\cite{DepthPro2024}  &  F & 951.99M & 0.344  & 0.763 & 1.807  & 0.484  & 0.294  &  \underline{0.588}  & {0.781} \\
& ICLR 2025  & DepthPro~\cite{DepthPro2024}~\textdagger  &  F & 951.99M & 0.065  & 0.064 & 0.600  & 0.086  & 0.971  &  0.996  & 0.999 \\
& arXiv 2025 & UniDepthV2-L~\cite{unidepthv2} & F & 353.22M & 0.627 & 2.339 & 3.417 & 0.478 & 0.112 & 0.380 & \underline{0.880} \\
& arXiv 2025 & UniDepthV2-L~\cite{unidepthv2}~\textdagger & F & 353.22M & 0.043 & 0.027 & 0.417 & 0.058 & 0.994 & 0.999 & 1.000 \\
& arXiv 2025 & MoGe2-L~\cite{wang2025moge2} & F & 330.90M & 0.576 & 1.666 & 2.664 & 0.445 & 0.078 & 0.550 & 0.910 \\
& arXiv 2025 & MoGe2-L~\cite{wang2025moge2}~\textdagger & F & 330.90M & 0.037 & 0.021 & 0.377 & 0.053 & 0.993 & 0.999 & 1.000 \\
& -  & GfM (ours)  &  M+Pose & \textbf{8.88M} & \textbf{0.034} & \textbf{0.028} & \textbf{0.372}  & \textbf{0.055} & \textbf{0.990} & \textbf{0.996} & \textbf{0.999} \\
\bottomrule
\end{tabularx}

\begin{tabularx}{\textwidth}{p{0.3cm}|p{2 cm} p{3cm} c c c c c c c c c }
\toprule
 & Published in & Method & Train & \#Params & PP & Abs Diff $\downarrow$  & RMSE $\downarrow$ & RMSE log $\downarrow$ & $\delta_1$ $\uparrow$ & $\delta_2$ $\uparrow$ & $\delta_3$ $\uparrow$ \\
\midrule
\multirow{8}{*}{\rotatebox[origin=c]{90}{{Gamma}}} 
& - & Monodepth2~\cite{MONODEPTH2}~$\ddagger$  &  M+Pose & 14.3M & \checkmark  & \underline{0.015}   & \underline{0.023}  & \textbf{1.124}  &  \underline{0.459}  & \underline{0.463} &  \underline{0.471}\\
& ICLR 2025 & DepthPro~\cite{DepthPro2024}  &  F& 951.99M & \checkmark  & 0.211   & 0.226  & 2.807  & 0.057  &   0.064  & 0.077 \\
& ICLR 2025 & DepthPro~\cite{DepthPro2024}~\textdagger  &  F & 951.99M & \checkmark  & 0.016  & 0.024  & 1.234   & 0.500  &  0.521   & 0.558 \\
& arXiv 2025 & UniDepthV2-L~\cite{unidepthv2} & F & 353.22M & \checkmark & 0.120 & 0.132 & 2.191 & 0.045 & 0.101 & 0.168 \\
& arXiv 2025 & UniDepth~\cite{unidepthv2}~\textdagger & F & 353.22M & \checkmark & 0.037 & 0.044 & 0.260 & 0.516 & 0.921 & 0.991 \\
& arXiv 2025 & MoGe2-L~\cite{wang2025moge2} & F & 330.90M & \checkmark & 0.142 & 0.155 & 2.216 & 0.018 & 0.034 & 0.056 \\
& arXiv 2025 & MoGe2-L~\cite{wang2025moge2}~\textdagger & F & 330.90M & \checkmark & 0.027 & 0.032 & 0.243 & 0.629 & 0.902 & 0.977 \\
& - & GfM (ours)  &  M+Pose & \textbf{8.88M}& -  & \textbf{0.009}  & \textbf{0.015} & \underline{1.142} & \textbf{0.587} & \textbf{0.591} & \textbf{0.605} \\
\bottomrule
\end{tabularx}
\end{subtable}

\caption{\textbf{Quantitative evaluation on KITTI and RSRD datasets.}
The upper rows present depth metrics evaluated against their respective ground-truth, while the lower rows present the height-to-depth ratio (\(\gamma\)) performance, computed using the improved KITTI ground-truth~\cite{IMPROVED_KITTI_GT} and dense point clouds for RSRD~\cite{zhao2023rsrd}.
Abbreviations:
\emph{M} (monocular-only),
\emph{PP} (RANSAC plane-fitting post-processing),
\emph{camH} (camera-height supervision),
\emph{\textdagger} (median-scaled by ground-truth. Thus, excluded from direct comparison due to scale drift correction),
\emph{V} velocity regression,
\emph{Pose} full GT pose supervision,
\emph{SI} object size priors from internet data,
\emph{F} foundational models trained on massive data,
\emph{$\ddagger$} Monodepth2 baseline trained on RSRD with GT poses.
}
\label{tab:main_tab}
\end{table*}

\paragraph{(2) Homography alignment loss.}
We compute a masked photometric loss using \(I_s^{w}\) and \(I_t\), restricting the loss to the dominant road plane:
\begin{equation}
\mathcal{L}_{\text{homo}} \;=\; \mathcal{M}_{\text{road}} \,\cdot\, \text{pe}\bigl(I_s^{w}, I_t\bigr),
\end{equation}
where \(\mathcal{M}_{\text{road}}\) is the probabilistic road mask defined in~\cref{paragraph:roadmask}. This loss enforces accurate alignment of planar road regions between views~\cite{MonoPP}.

\paragraph{(3) Normal consistency loss.}
To avoid degenerate or inverted normals, we penalize deviations between the predicted normal \({\vec{N}}_{\mathrm{pred}}\) and a reference normal \({\vec{N}}_{\mathrm{ref}}\) using a hinge-based angular loss:
\begin{gather}
\cos(\Delta \theta) = 
\frac{{\vec{N}}_{\mathrm{pred}} \cdot {\vec{N}}_{\mathrm{ref}}}
     {\|{\vec{N}}_{\mathrm{pred}}\|\;\|{\vec{N}}_{\mathrm{ref}}\|}, \\
\mathcal{L}_{\text{norm}} = [\,1 - \cos(\Delta\theta)\,] \;+\; 
\mathrm{ReLU}\bigl(\cos(\theta_{\text{thres}})- \nonumber \\[-0.1em] 
\cos(\Delta \theta)\bigr)^2.
\end{gather}
This formulation encourages physically plausible normals, imposing larger penalties when angular deviations exceed the threshold \(\theta_{\text{thres}}\).

\paragraph{Total loss.}
Including an additional edge-aware smoothness term \(\mathcal{L}_{\text{smooth}}\)~\cite{godard2017unsupervised_monodepth1,MONODEPTH2}, our final objective is:
\begin{equation}
\mathcal{L}_{\text{total}}
=\mathcal{L}_{\text{photo}} + \mathcal{L}_{\text{homo}} + \lambda_{\text{norm}}\mathcal{L}_{\text{norm}} + \lambda_{\text{smooth}}\mathcal{L}_{\text{smooth}},
\end{equation}
where \(\lambda_{\text{norm}}\), \(\lambda_{\text{smooth}}\) balance loss components.

%% file: sec/2_formatting.tex
\begin{figure*}[t]
    \centering
    %
    %
    \begin{minipage}[c]{0.23\textwidth}  
        \centering
        \vspace{4em}
        \includegraphics[width=1\linewidth,height=1.5cm]{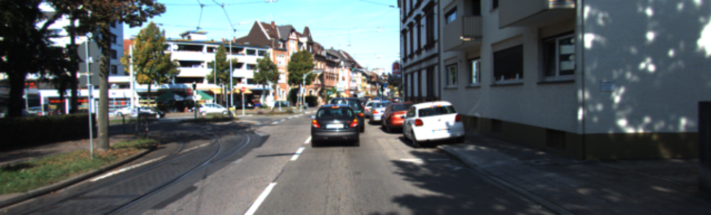}\\
        \vspace{0.3em}
        \small (a)~Input (KITTI)

        \vspace{7.2em}

        \includegraphics[width=\linewidth,height=1.5cm]{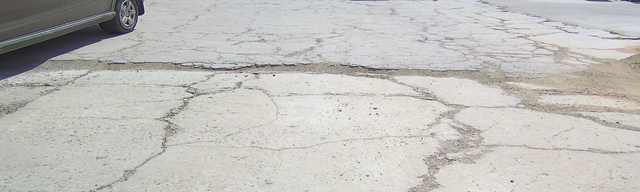}\\
        \vspace{0.3em}
        \small (b)~Input (RSRD)
    \end{minipage}
    %
    %
    \begin{minipage}[c]{0.75\textwidth}
        \centering
        \small
        \setlength{\tabcolsep}{2pt}
        \begin{tabular}{
            >{\centering\arraybackslash}m{0.12\textwidth}  
            >{\centering\arraybackslash}m{0.20\textwidth}  
            >{\centering\arraybackslash}m{0.20\textwidth}  
            >{\centering\arraybackslash}m{0.20\textwidth}  
            >{\centering\arraybackslash}m{0.03\textwidth}  
        }
            \multicolumn{5}{c}{Example (a) KITTI} \\
            \toprule
            \textbf{Description}
             & \textbf{GfM (ours)}
             & \textbf{GroCo~\cite{GroCo2024}}
             & \textbf{DepthPro~\cite{DepthPro2024}}
             &  \\
            \midrule

            $\gamma$ &
            \includegraphics[width=\linewidth,height=1cm]{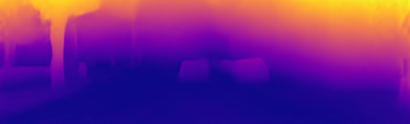} &
            \includegraphics[width=\linewidth,height=1cm]{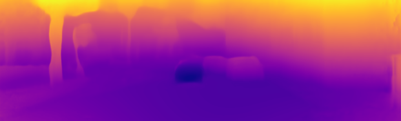} &
            \includegraphics[width=\linewidth,height=1cm]{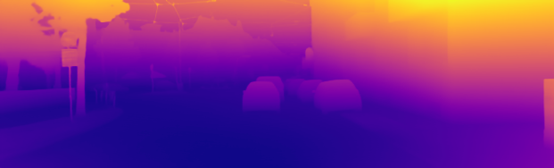} &
            \includegraphics[width=0.9\linewidth,height=1cm]{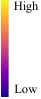} \\

            $\gamma$ error &
            \includegraphics[width=\linewidth,height=1cm]{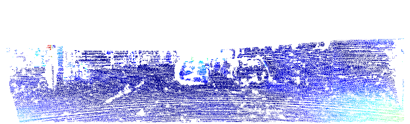} &
            \includegraphics[width=\linewidth,height=1cm]{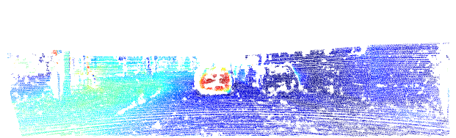} &
            \includegraphics[width=\linewidth,height=1cm]{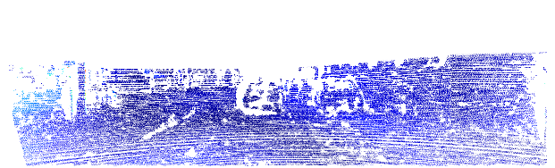} &
            \includegraphics[width=0.9\linewidth,height=1cm]{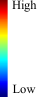} \\

            Depth &
            \includegraphics[width=\linewidth,height=1cm]{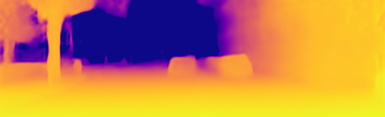} &
            \includegraphics[width=\linewidth,height=1cm]{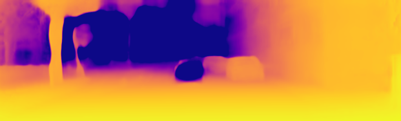} &
            \includegraphics[width=\linewidth,height=1cm]{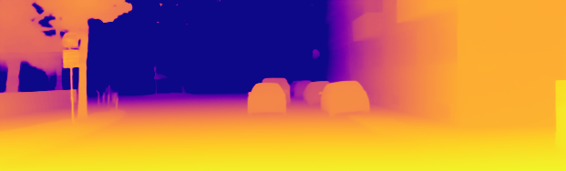} &
            \includegraphics[width=0.9\linewidth,height=1cm]{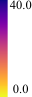} \\

            Depth error &
            \includegraphics[width=\linewidth,height=1cm]{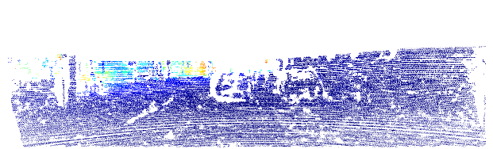} &
            \includegraphics[width=\linewidth,height=1cm]{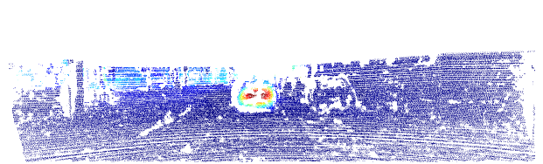} &
            \includegraphics[width=\linewidth,height=1cm]{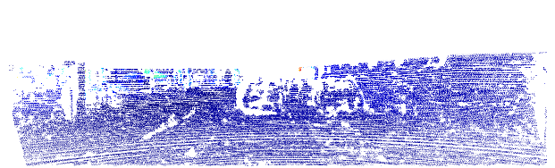} &
            \includegraphics[width=0.9\linewidth,height=1cm]{figures/colormaps/colormaps_E.pdf} \\
            \midrule

            \multicolumn{5}{c}{{Example (b) RSRD}} \\
            \toprule
            \textbf{Description}
             & \textbf{GfM (ours)}
             & \textbf{Monodepth2~\cite{MONODEPTH2}}
             & \textbf{DepthPro~\cite{DepthPro2024}}
             &  \\
            \midrule

            $\gamma$ &
            \includegraphics[width=\linewidth,height=1cm]{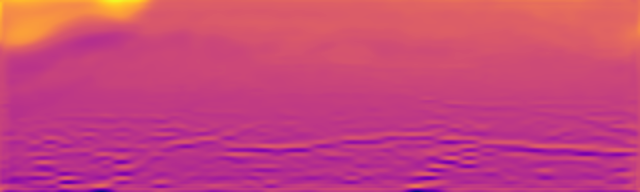} &
            \includegraphics[width=\linewidth,height=1cm]{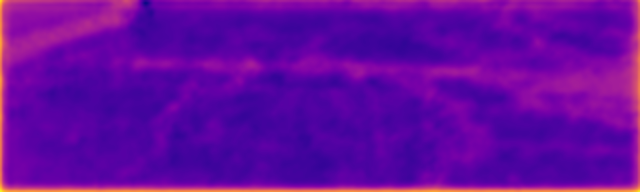} &
            \includegraphics[width=\linewidth,height=1cm]{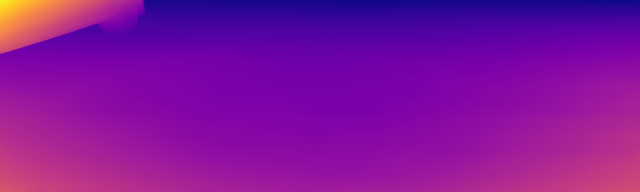} &
            \includegraphics[width=0.9\linewidth,height=1cm]{figures/colormaps/new_colormap.pdf} \\

            $\gamma$ error &
            \includegraphics[width=\linewidth,height=1cm]{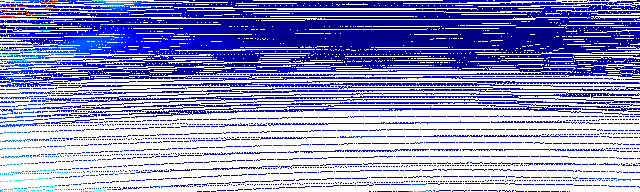} &
            \includegraphics[width=\linewidth,height=1cm]{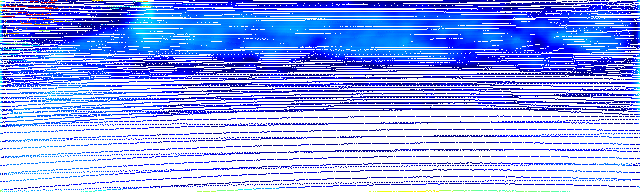} &
            \includegraphics[width=\linewidth,height=1cm]{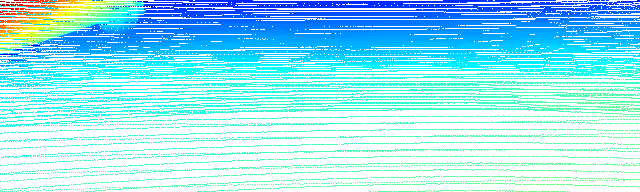} &
            \includegraphics[width=0.9\linewidth,height=1cm]{figures/colormaps/colormaps_E.pdf} \\

            Clipped height  &
            \includegraphics[width=\linewidth,height=1cm]{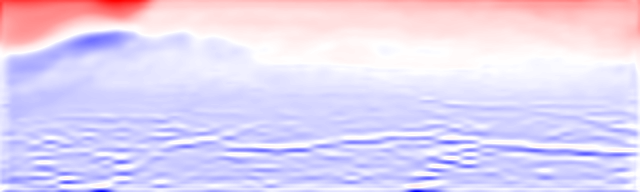} &
            \includegraphics[width=\linewidth,height=1cm]{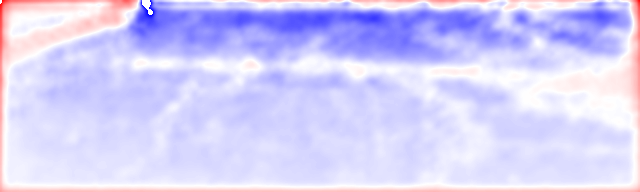} &
            \includegraphics[width=\linewidth,height=1cm]{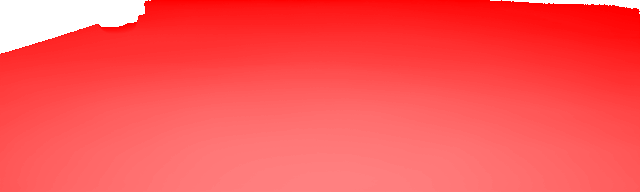} &
            \includegraphics[width=0.9\linewidth,height=1cm]{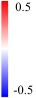} \\
            \bottomrule
        \end{tabular}
    \end{minipage}

    \caption{
    \textbf{Qualitative comparison on KITTI and RSRD.}
    Left: (a) KITTI~\cite{geiger2012kitti} input and (b) RSRD~\cite{zhao2023rsrd} input.
    Right: 
    (a)~KITTI example comparing GfM, GroCo~\cite{GroCo2024}, and DepthPro~\cite{DepthPro2024} on $\gamma$, $\gamma$ error (Abs Diff), depth, and depth error (Abs Rel).
    (b)~RSRD example comparing GfM, Monodepth2~\cite{MONODEPTH2}, and DepthPro on $\gamma$ prediction, $\gamma$ error map (Abs Diff), and height predictions clipped to $[-0.5,0.5]$.
    Colormaps for each metric are shown in the last column.
    }
    \label{fig:Qualitative_big_results}
\end{figure*}

\section{Experiments}
\label{sec:Experiments}

\paragraph{Evaluation overview.}
We evaluate on KITTI~\cite{geiger2012kitti}, the standard monocular depth benchmark for vehicular application, and RSRD~\cite{zhao2023rsrd}, which contains varied height perturbations. KITTI allows benchmarking against prior work, while RSRD tests road-topography sensitivity and enables evaluation on diverse terrain. All experiments use an image resolution of \(640\times192\).

For KITTI~\cite{geiger2012kitti}, we use the Eigen split~\cite{eigen2014depth}: 39{,}180 monocular triplets for training, 4{,}424 images for validation, and 697 for testing. Evaluation follows the improved benchmark~\cite{IMPROVED_KITTI_GT} with a GT camera height of $\approx\!1.65$\,m~\cite{koledic2024gvdepth}. Depth is clipped at 80\,m for both depth and $\gamma$ evaluations. On the other hand, given the textureless surface of RSRD, we rely on the provided ground-truth poses. We train on 7{,}040 sparse triplets and evaluate on the 600 images with dense ground truth.


\paragraph{Implementation details.}
Our method is implemented in PyTorch and trained on a single NVIDIA~V100 GPU with a batch size of 12. For KITTI, training consists of two phases: (1) Pretraining as monocular depth estimation for 10 epochs, jointly learning depth and metric-scaled pose from PoseNet. (2) Switching to \(\gamma\)-prediction, training for 25 additional epochs on KITTI (35 epochs total).  

For RSRD, we use ground-truth poses rather than jointly optimizing pose and structure, as learning both simultaneously on textureless data is unreliable. Consequently, we omit the homography loss (\(\mathcal{L}_{\text{homo}}\)) for RSRD, since its primary function is enforcing metric scale in jointly learned depth and pose.

The network outputs a sigmoid map that is converted to \(\gamma\) via the signed log-space transform in~\cref{eqn:first_gamma_transform} with \(\alpha=0.5\). We set the \(\gamma\) range to \([\,-0.1,\,5.0\,]\) for KITTI and \([\,-0.5,\,2.0\,]\) for RSRD. For the normal-consistency loss we use an angular threshold \(\theta_{\text{thres}}=5^\circ\) and reference normals \(\vec{N}_{\mathrm{ref}}=[0,-1,0]\) for KITTI and \([0,-0.95,-0.29]\) for RSRD. We optimize with AdamW (weight decay \(10^{-2}\)) and a cosine learning-rate schedule initialized at \(5\times10^{-4}\); loss weights are \(\lambda_{\text{norm}}=0.1\) and \(\lambda_{\text{smooth}}=10^{-2}\). During training, we follow the data-augmentation and auto-masking strategies of~\cite{MonoPP,MONODEPTH2}.

\begin{table*}[t]
  \centering
  \footnotesize
  \setlength{\tabcolsep}{3pt}
  \renewcommand{\arraystretch}{0.95}
  \caption{Ablation study on KITTI benchmark (capped at 80\,m). Lower is better except \(\delta\uparrow\). \textbf{Bold} indicates best in column.}
  \label{tab:ablation_study}
  \begin{tabular}{l|ccccc|ccccc}
    \toprule
    & \multicolumn{5}{c|}{\textbf{Gamma}} & \multicolumn{5}{c}{\textbf{Depth}} \\
    \textbf{Variant} &
    Abs Diff $\downarrow$ & RMSE $\downarrow$ & RMSE log $\downarrow$ & $\delta_1 \uparrow$ & $\delta_3 \uparrow$ &
    Abs Rel $\downarrow$ & RMSE $\downarrow$ & RMSE log $\downarrow$ & $\delta_1 \uparrow$ & $\delta_3 \uparrow$ \\
    \midrule
    GfM (full) &
    \textbf{0.012} & \textbf{0.018} & \textbf{0.801} & \textbf{0.706} & \textbf{0.801} &
    \textbf{0.085} & \textbf{4.082} & 0.139 & \textbf{0.916} & \textbf{0.995} \\

    w/o predicted normal (fixed $\vec{N}$) &
    0.013 & 0.020 & 0.909 & 0.658 & 0.796 &
    0.095 & 4.258 & 0.150 & 0.903 & 0.995 \\

    w/o log-space transform &
    0.016 & 0.024 & 1.119 & 0.558 & 0.695 &
    0.095 & 4.302 & 0.144 & 0.907 & 0.994 \\

    w/o ImageNet pretraining &
    0.015 & 0.024 & 1.033 & 0.645 & 0.777 &
    0.111 & 4.648 & 0.168 & 0.872 & 0.989 \\

    w/o probabilistic road mask &
    0.013 & 0.021 & 0.944 & 0.678 & 0.796 &
    0.087 & 4.210 & \textbf{0.138} & 0.908 & 0.995 \\
    \bottomrule
  \end{tabular}
  \vspace{-0.5em}
\end{table*}

\paragraph{Baselines.}  
We focus on metric-scaled self-supervised methods as well as several foundation models, with DepthPro~\cite{DepthPro2024} highlighted in the main tables as a representative foundation model due to its high sensitivity to fine geometric variations. For KITTI, comparisons include methods such as GroCo~\cite{GroCo2024}, using publicly available pretrained weights. Since no prior self-supervised method was trained on RSRD and training code for recent models is unavailable, we construct a baseline by training Monodepth2~\cite{MONODEPTH2} with known poses and identical inputs to GfM. For each depth map (prediction and GT), \(\gamma\) is computed from depth by fitting the ground plane with RANSAC (10{,}000 iterations, \(0.01\,\mathrm{m}\) inlier threshold). More details are in the Supplementary (\cref{sec:ransac_plane_fit}), and further evaluations are in~\cref{sec:app_quantitative_results}.

\paragraph{Evaluation metrics.}
For depth estimation, we use standard metrics: \emph{Abs Rel}, \emph{Sq Rel}, \emph{RMSE}, \emph{RMSE log}, and \(\delta\)-accuracies \((\delta_1, \delta_2, \delta_3)\), following prior work~\cite{MonoPP,MONODEPTH2,manydepth,litemono_2023_selfsup}. For \(\gamma\) (signed height-to-depth), standard relative metrics can become unstable near or below zero. As a result, we define:
\begin{itemize}[leftmargin=1.5em]
    \item \emph{Abs Diff}: mean absolute difference \(\bigl|\gamma_{\mathrm{pred}} - \gamma_{\mathrm{gt}}\bigr|\),
    \item \emph{RMSE} \& \emph{RMSE log}: computed with a small offset to handle sign changes,
    \item Modified \(\delta\)-accuracies: pixels are correct if \(\bigl|\gamma_{\mathrm{pred}} - \gamma_{\mathrm{gt}}\bigr| < 0.01\) or satisfy the ratio-based condition~\cite{MONODEPTH2}.
\end{itemize}
These adjustments ensure a robust evaluation for the signed \(\gamma\) predictions.

\begin{figure}[h!]
    \centering
    \includegraphics[width=\linewidth]{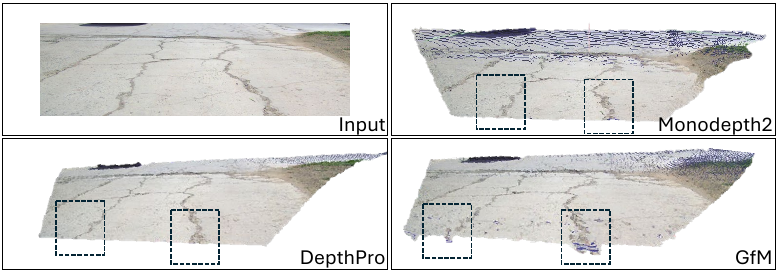}
    \caption{
        \textbf{Point Cloud Comparison.} 
    Computed 3D point clouds from different models on a sample image from RSRD~\cite{zhao2023rsrd}. DepthPro exhibits scale drift and slightly flattens the road surface. Meanwhile, Monodepth2~\cite{MONODEPTH2}, despite being trained on the dataset, still struggles to capture road topography accurately. Additional qualitative examples are in~\cref{fig:rsrd_pointcloud_comparison_appendix}.
    }

    \label{fig:pointcloud_comparison}
\end{figure}

\paragraph{Results.}
As summarized in~\cref{tab:main_tab:kitti}, GfM delivers competitive performance in general depth estimation on KITTI while using significantly fewer parameters and less supervision, and achieves state-of-the-art results in near-range depth and $\gamma$ metrics. Qualitative examples in~\cref{fig:Qualitative_big_results} further illustrate these strengths: in example~(a), our method outperforms prior work GroCo~\cite{GroCo2024} in challenging dynamic-object scenarios, though far-away dynamic objects remain difficult, as shown in the supplementary material (~\cref{tab:qualitative_results_appendix_bad}).
On the new RSRD dataset (\cref{tab:main_tab:rsrd}), GfM surpasses all evaluated methods in both $\gamma$ and depth metrics, and produces more reliable 3D reconstructions, as seen in~\cref{fig:pointcloud_comparison}, where foundation models often misinterpret road slopes without domain adaptation. Example~(b) in~\cref{fig:Qualitative_big_results} shows how our accurate $\gamma$ prediction captures the fine details of road topography with higher fidelity than both conventional depth-based self-supervised models and large-scale foundation models. Importantly, $\gamma$ evaluation is invariant to camera height assumptions: perturbing the height only rescales recovered metric depth, leaving $\gamma$ unchanged. This makes $\gamma$ a physically interpretable indicator of road-relative geometry.

\paragraph{Ablation study.}
\Cref{tab:ablation_study} reports ablations on KITTI. Although omitted from the table due to catastrophic metric failures, some components have particularly strong influence. Removing the normal constraint $\mathcal{L}_{\text{norm}}$ causes $\vec{N}_{\text{pred}}$ to collapse into degenerate or flipped orientations, degrading road-topography estimation and photo-consistency; this results in a $\gamma$ Abs Diff of $2.282$, $\delta_1^{\gamma}=0.00$, depth Abs Rel of $0.937$, and $\delta_1^{\text{depth}}=0.00$. Removing the homography alignment loss $\mathcal{L}_{\text{homo}}$ eliminates the dominant-plane scale cue, leaving $\gamma$ largely unaffected but severely degrading metric depth. Other components contribute as follows (\cref{tab:ablation_study}, rows~1--4).

\begin{enumerate}[label=\arabic*., leftmargin=*, itemsep=0.2em, topsep=0.2em]
    \item \textbf{Learned normal}
    Predicting $\vec{N}_{\text{pred}}$ outperforms using KITTI extrinsics directly, consistent with prior findings that vehicle tilt varies in real driving due to suspension and load~\cite{koledic2024gvdepth}.
    \item \textbf{Signed log-space mapping.}
    Replacing the signed log-space transformation (\cref{subsec:network_archi}) with a linear mapping, as in~\cite{MONODEPTH2,manydepth,MonoPP,litemono_2023_selfsup}, caused training to diverge, unless the learning rate was reduced by $10\times$, even then, performance was worse.

    \item \textbf{ImageNet pretraining.}
    Pretraining yields consistent gains for both $\gamma$ and depth, matching trends in self-supervised depth~\cite{litemono_2023_selfsup,MONODEPTH2}.

    \item \textbf{Probabilistic road mask $\mathcal{M}_{\text{road}}$.}
    Replacing our probabilistic mask with a binary mask, as in~\cite{MonoPP,DNet}, reduces performance; the probabilistic near-field weighting is more robust.
\end{enumerate}


\section{Conclusion}
In this work, we tackled a core limitation of monocular depth estimation: its inability to deliver explicit scene geometry for real-world vehicular tasks due to scale ambiguity, extrinsic calibration sensitivity, and limited structural understanding. We introduced a lightweight model that predicts the \(\gamma\) representation and a global road-plane normal, which unifies depth and height and reduces reliance on extrinsic calibration. In addition, it is trained self-supervised on monocular driving video. Given known camera height, GfM turns a single image into metric, detailed near-field geometry that exposes subtle road irregularities for safer planning.

\paragraph{Limitations.}
Metric scale depends on camera-height accuracy, as miscalibration introduces a global scale factor while preserving relative geometry. As with other self-supervised methods, performance can degrade on dynamic objects, especially distant objects.

\paragraph{Future work.}
We will extend the framework toward a foundation model for vehicular perception by self-supervising \(\gamma\) on large-scale monocular driving videos, recovering metric scale at test time via the known camera height. In addition, we will explore using \(\gamma\) directly for on-road control and planning where full 3D reconstruction is unnecessary.

%% file: sec/3_finalcopy.tex



%% file: sec/X_suppl.tex
\clearpage
\setcounter{page}{1}
\maketitlesupplementary

\begin{figure*}[t]
    \centering
    \includegraphics[width=\textwidth]{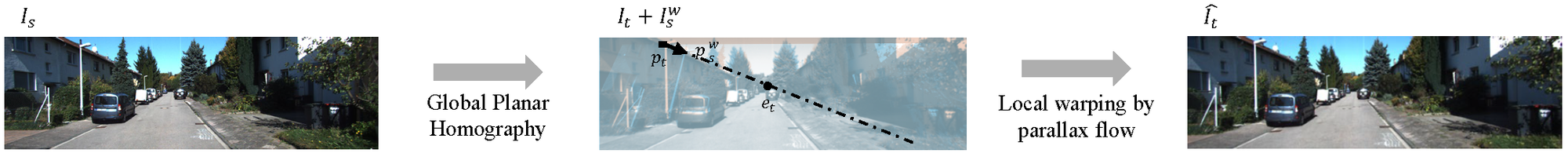}
    \caption{Decomposition of image synthesis into two steps. (Left) Source image. 
    (Middle) After applying the global planar homography, 
    the remaining residual flow is inherently epipolar 
    and directly tied to the per-pixel $\gamma$ parameter. 
    (Right) Synthesized target image by residual-flow warping.}
    \label{fig_app:pp_second_fig}
\end{figure*}

\begin{figure*}[t!]
    \centering
    \includegraphics[width=\linewidth]{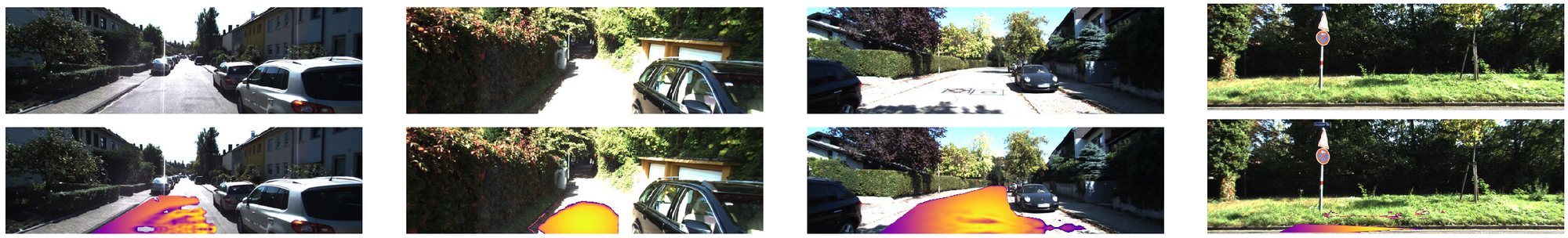}
    \caption{Example of input images with the computed mask during training for identifying the road region.}
    \label{fig:road_mask_ex}
\end{figure*}
\section{Why $\gamma$-Space Exposes Near-Field Road Geometry}
\label{subsec:gamma_vs_depth}

Small vertical variations near the vehicle, such as speed bumps, shallow ramps, induce tiny depth changes and are easily masked by depth noise. In contrast, $\gamma$-space (\(\gamma=h/d\)) directly encodes height relative to distance, making small near-field elevations produce clearer error signals. Because $\gamma$ can approach zero on the road, we report absolute errors for depth, $\gamma$, and height to avoid unstable ratios. In this example, we show how a small near-field deviation (speed bump) is amplified in $\gamma$ compared with a larger depth error on a tall object (tree).

\paragraph{Numerical example of~\cref{fig:gamma_vs_depth_main_paper}.}
Two objects at ground-truth range \(d_{\text{GT}}=3.00\,\mathrm{m}\):
\begin{itemize}[leftmargin=1.1em,itemsep=0.25em]
  \item \textbf{Tree.}
  \begin{flushright}
  \small
  \setlength{\tabcolsep}{2pt}
  \begin{tabular}{@{}l l@{}}
    GT $(d,h,\gamma)$: & $(3.00\,\mathrm{m},\,2.00\,\mathrm{m},\,0.667)$ \\
    Pred $(d',h',\gamma')$: & $(3.50\,\mathrm{m},\,2.20\,\mathrm{m},\,0.629)$ \\
    Abs.\ errors $(|\Delta d|,|\Delta h|,|\Delta\gamma|)$: & $(0.50\,\mathrm{m},\,0.20\,\mathrm{m},\,0.038)$
  \end{tabular}
  \end{flushright}

  \item \textbf{Speed bump.}
  \begin{flushright}
  \small
  \setlength{\tabcolsep}{2pt}
  \begin{tabular}{@{}l l@{}}
    GT $(d,h,\gamma)$: & $(3.00\,\mathrm{m},\,0.150\,\mathrm{m},\,0.050)$ \\
    Pred $(d',h',\gamma')$: & $(3.20\,\mathrm{m},\,0.000\,\mathrm{m},\,0.000)$ \\
    Abs.\ errors $(|\Delta d|,|\Delta h|,|\Delta\gamma|)$: & $(0.20\,\mathrm{m},\,0.150\,\mathrm{m},\,0.050)$
  \end{tabular}
  \end{flushright}
\end{itemize}

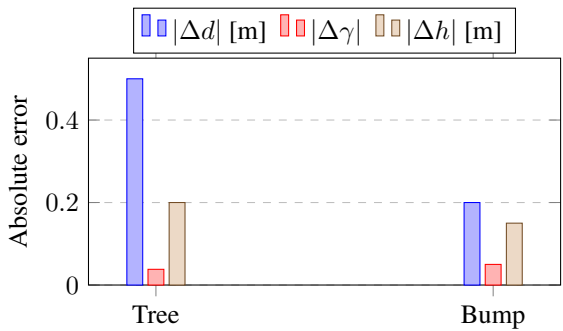
\begin{figure}[H]
\centering
\begin{tikzpicture}
\begin{axis}[
    width=\linewidth,
    height=4.6cm,
    ybar,
    bar width=6pt,
    ymin=0,
    ylabel={Absolute error},
    symbolic x coords={Tree,Bump},
    xtick=data,
    legend style={at={(0.5,1.02)}, anchor=south, legend columns=3,
      /tikz/every even column/.style={column sep=3pt}},
    legend cell align=left,
    enlarge x limits=0.20,
    ymajorgrids=true,
    grid style=dashed,
    tick label style={/pgf/number format/fixed},
]
\addplot coordinates {(Tree,0.50) (Bump,0.20)}; \addlegendentry{$|\Delta d|$ [m]}
\addplot coordinates {(Tree,0.0381) (Bump,0.0500)}; \addlegendentry{$|\Delta \gamma|$}
\addplot coordinates {(Tree,0.20) (Bump,0.150)}; \addlegendentry{$|\Delta h|$ [m]}
\end{axis}
\end{tikzpicture}
\caption{\textbf{Absolute errors for depth, $\gamma$, and height (single example).}
Depth alone makes the near-field bump look acceptable (\(0.20\,\mathrm{m}\) is small), while $\gamma$ and derived height expose the miss clearly.}
\label{fig:abs_error_bars}
\end{figure}

\begin{table}[h]
\centering
\footnotesize
\setlength{\tabcolsep}{3.5pt}
\renewcommand{\arraystretch}{0.95}
\caption{Ground truth vs.\ prediction at \(d_{\text{GT}}=3.00\) m (all \emph{absolute} errors).}
\label{tab:gamma_vs_depth}
\begin{tabular}{l|ccc|ccc}
\toprule
 & \multicolumn{3}{c|}{\textbf{Depth / Height}} & \multicolumn{3}{c}{\textbf{$\gamma$}} \\
\textbf{Object} & $d_{\text{GT}}/d'$ & $h_{\text{GT}}/h'$ & $|\Delta d|$ & $\gamma_{\text{GT}}$ & $\gamma'$ & $|\Delta\gamma|$ \\
\midrule
Tree  & 3.00 / 3.50 & 2.00 / 2.20 & \textbf{0.50} & 0.667 & 0.629 & \textbf{0.038} \\
Bump  & 3.00 / 3.20 & 0.150 / 0.000 & \textbf{0.20} & 0.050 & 0.000 & \textbf{0.050} \\
\bottomrule
\end{tabular}
\vspace{-0.5em}
\end{table}

\noindent\textbf{Takeaway.} In \textbf{depth} space, the bump’s error (\(0.20\,\mathrm{m}\)) appears negligible, but in \(\boldsymbol{\gamma}\)-space (and height) it is unmistakable. Switching the representation from depth to $\gamma$ highlights near-field topology and small obstacles that depth alone tends to hide, while still reflecting height–depth mismatches for tall structures (tree).

\section{Extracting the Local Normal Vectors}
\label{sec:local_normals}

We estimate per-pixel normals in three steps: (i) backproject pixels to 3D, (ii) form centered 3D differences, and (iii) normalize their cross product. Given depth $D(u,v)$ and intrinsics $K$,
$P(u,v)=D(u,v)\,K^{-1}[u,\,v,\,1]^\top$.
Using a small offset $\Delta$ with reflection padding, define~\cite{MonoPP,DNet}
$\mathbf d_x=P(u+\Delta,v)-P(u-\Delta,v)$ and
$\mathbf d_y=P(u,v+\Delta)-P(u,v-\Delta)$.
The unit normal is
\[
\mathbf n(u,v)=\frac{\mathbf d_x\times \mathbf d_y}{\lVert\mathbf d_x\times \mathbf d_y\rVert}.
\]

\section{Gaussian Probabilistic Road Mask}
\label{sec:road_mask_appendix}

As shown in~\cref{fig:road_mask_ex}, using the extracted local surface normals, we define a probabilistic road mask by computing the angular similarity between the local normals and a reference ground normal, as an example \([0,-1,0]\) in camera coordinates for KITTI-style camera mounting position. 

The probability of a pixel belonging to the road is computed using a soft ramp function based on the cosine similarity between the computed normal \( \mathbf{n}(u,v) \) and the reference ground normal \( \mathbf{n}_{ref} \):

\begin{equation}
P_{\text{road}}(u,v)
= \max\!\left(
  0,\ 
  \frac{\bigl|\vec{N}_{\mathrm{pred}}(u,v)\cdot \vec{N}_{\mathrm{ref}}\bigr|}
       {\cos(\theta_{\text{tol}})}
\right),
\label{eq:roadprob}
\end{equation}

where:
\begin{itemize}
    \item \( \theta_{tol} \) is the threshold angle that determines the maximum tolerance, that a normal can have while still being considered part of the road.
    \item The cosine similarity is used since normals are already normalized, making the dot product a direct measure of angular difference.
\end{itemize}

\paragraph{Gaussian mask.}To further refine the road probability map, we apply:
\begin{itemize}
    \item A trapezoidal weighting mask that emphasizes the lower portion of the image, ensuring that the road mask remains spatially consistent.
    \item A bottom-centered Gaussian weighting function \(G(u,v)\) to bias the probability toward the expected road region. This Gaussian mask is defined as:
    \begin{equation}
    G(u,v) = \exp\!\left(
      -\frac{(u - c_x)^2}{2\,\sigma_w^2}
      -\frac{(v - c_y)^2}{2\,\sigma_h^2}
    \right),
    \end{equation}
    where \(c_y\) and \(c_x\) define the center position of the Gaussian, while \(\sigma_h\) and \(\sigma_w\) control the spread in the vertical and horizontal directions.
    The Gaussian is then normalized so that the maximum value is 1 to ensure uniform scaling:
    \begin{equation}
    G(u,v) = \frac{G(u,v)}{\max(G)}
    \end{equation}
\end{itemize}

The final probabilistic road mask is obtained by thresholding \(P_{final}\) at a median-derived threshold, ensuring that detected road pixels are robust to noise and depth variations.

\section{Computing Gamma By Extracting the Road Plane Using RANSAC}
\label{sec:ransac_plane_fit}

To convert depth values into a height-to-depth ratio \( \gamma \) relative to the road plane, usually the ground-plane should be identified. as an example for the image synthesis via planar-parallax process to be computed, as shown in~\cref{fig_app:pp_second_fig}. We employ a RANSAC plane-fitting algorithm. In addition, it was utilized to convert ground-truth depth maps, as well as baselines, into \( \gamma \) representations.

Our approach follows these key steps:

Let \(P\) denote the 3D points obtained from the depth map.
\begin{enumerate}[leftmargin=1.15em,itemsep=0.2em]
  \item \textbf{Pre-filtering.} Keep finite points within a reasonable range, in KTTI 80 meters as an example, and restrict to a lower trapezoid ROI consistent with \cref{sec:road_mask_appendix}.
  \item \textbf{Hypothesis.} Sample three non-collinear points \((P_1,P_2,P_3)\) and form a candidate plane with
        \( \mathbf n \propto (P_2{-}P_1)\times(P_3{-}P_1) \) and \( d = -\mathbf n\!\cdot\!P_1 \).
  \item \textbf{Scoring.} A point \(P\) is an inlier if \( \lvert \mathbf n\!\cdot\!P + d \rvert < \tau \).
        Keep the hypothesis with the most inliers, where $\tau$ is the threshold.
  \item \textbf{Orientation.} Enforce an upward normal: if \( \mathbf n\!\cdot\!\mathbf n_{\mathrm{ref}} < 0 \) (ground-up reference), flip \(\mathbf n,d \leftarrow -\mathbf n,-d\). This is mainly to keep a consistent convention of the height definition to be positive above the road.
  \item \textbf{Height \& gamma.} For each pixel, compute \( H(u,v)=\mathbf n\!\cdot\!P(u,v)+d \) and \( \gamma(u,v)=H(u,v)/D(u,v) \).
\end{enumerate}






\section{Global Road Normal Prediction}
\label{sec_app:global_road_normal_prediction}
In addition to depth, the decoder predicts a single global road-plane normal vector $\vec{N}_{\text{pred}} \in \mathbb{R}^3$ for the entire ground region.  The normal branch is attached to the \emph{bottleneck features} of the encoder. We first apply an \textbf{attention pooling} layer to aggregate the $H'\times W'$ spatial features into a single $C$-dimensional descriptor.
This pooling uses a learnable query vector $\mathbf{q} \in \mathbb{R}^C$ to produce attention weights over all spatial locations, highlighting road-like features and suppressing irrelevant ones:
\[
\alpha_{ij} = \frac{\exp(\mathbf{q}^\top k_{ij})}{\sum_{p,q} \exp(\mathbf{q}^\top k_{pq})}
,\quad
\mathbf{g} = \sum_{i,j} \alpha_{ij} v_{ij}
\]
where $k_{ij}, v_{ij} \in \mathbb{R}^C$ are the projected key/value vectors at location $(i,j)$.

The pooled descriptor $\mathbf{g}$ is then passed through a fully-connected layer producing 3 channels, reshaped to $[B, 3, 1, 1]$ and $\ell_2$-normalized to unit length:
\[
\vec{N}_{\text{pred}} = \frac{W_n \,\mathbf{g} + b_n}{\lVert W_n \,\mathbf{g} + b_n \rVert_2}
\]
This guarantees a valid unit normal at all times.

In practice, we find that detaching the normal branch from the depth gradients slightly improve stability across different trials, but we keep it trainable in our final model.

\paragraph{Integration into the decoder.}
The normal head is independent of the disparity prediction heads, which operate at multiple scales $s \in \{0,1,2,3\}$ from the upsampling tower.  
While depth is predicted at multiple resolutions, the normal is predicted once from the bottleneck, using only $\approx 3C + C^2$ parameters for the \texttt{AttnPool2d} and linear layer.  
This lightweight design adds negligible overhead ($< 0.1\%$ params) and ensures that $\vec{N}_{\text{pred}}$ captures the global road-plane geometry rather than local texture noise.

\section{Extra Quantitative Results}
\label{sec:app_quantitative_results}

Tables~\ref{tab:kitti_eigen_bins} and \ref{tab_app:kitti_eigen_full} report our evaluation on the KITTI Eigen split. 
Table~\ref{tab:kitti_eigen_bins} uses the improved KITTI ground-truth~\cite{IMPROVED_KITTI_GT} and evaluates performance at different maximum depth ranges. 
At the farthest range (\(\leq\)80\,m), our method (GfM) is competitive but does not surpass the strongest self-supervised baselines such as GroCo~\cite{GroCo2024} or MonoPP~\cite{MonoPP} in certain long-range metrics. 
However, our model requires only \textbf{8.88\,M} parameters, less than half the size of most compared networks-, and uses only camera-height supervision, unlike methods relying on full pose or additional priors. 

When the evaluation range is reduced, GfM improves steadily relative to other methods. 
At \(\leq\)60\,m and \(\leq\)40\,m, it often achieves the best or second-best results in most metrics. 
At \(\leq\)20\,m, GfM attains the best performance across all metrics (excluding median-scaled results), confirming its strength in the near range where precise absolute depth is critical. 

Table~\ref{tab_app:kitti_eigen_full} complements this analysis by benchmarking on the full Eigen split with the original sparse KITTI ground-truth, enabling comparison with historical methods not evaluated on dense GT. 
Despite the varied training setups of these baselines, GfM remains competitive while being much smaller and requiring only the camera height as scale recovery cue. This demonstrates a favourable trade-off between accuracy, efficiency, and supervision, particularly for scenarios prioritising short-to-mid-range depth estimation.

\begin{table*}[t!]
\centering
\footnotesize
\setlength\tabcolsep{3pt}
\renewcommand{\arraystretch}{1}

\caption{\textbf{KITTI (Eigen) depth results across evaluation ranges.}
Per-column best is \textbf{bold}, second-best is \underline{underlined}. 
Median-scaled rows (†) are \emph{excluded} from best/second-best consideration. 
All results use the Eigen split, we reference the improved KITTI ground-truth~\cite{IMPROVED_KITTI_GT} to be consistent with our evaluation protocol.}
\label{tab:kitti_eigen_bins}

\begin{subtable}{\textwidth}
\centering
\caption{Eigen benchmark~\cite{IMPROVED_KITTI_GT} \(\leq\)80 m}
\label{tab:eigen_80m}

\begin{tabularx}{\textwidth}{p{0.3cm}|p{2 cm} p{3cm} c c c c c c c c c }
\toprule
 & Published in & Method & Train & \#Params & Abs Rel $\downarrow$ & Sq Rel $\downarrow$ & RMSE $\downarrow$ & RMSE log $\downarrow$ & $\delta_1$ $\uparrow$ & $\delta_2$ $\uparrow$ & $\delta_3$ $\uparrow$ \\
\midrule
\multirow{5}{*}{\rotatebox[origin=c]{90}{Depth}} 
& ECCV 2024 & GroCo~\cite{GroCo2024}  &  M+camH & 34.65M & \underline{0.089} & \underline{0.516} & \underline{3.800} & \textbf{0.134} & \underline{0.910} & \textbf{0.984} & \textbf{0.995} \\
& WACV 2025 & MonoPP~\cite{MonoPP}  &  M+camH & 34.57M & \underline{0.089} & 0.545 & 3.864 & \textbf{0.134} & \underline{0.913} & \underline{0.983} & \textbf{0.995} \\
& ICLR 2025 & DepthPro~\cite{DepthPro2024}  &  F & 951.99M & 0.117 & \textbf{0.502} & \textbf{3.720} & 0.149 & 0.874 & 0.979 & \textbf{0.995} \\
& ICLR 2025 & DepthPro~\cite{DepthPro2024}~\textdagger &  F & 951.99M & 0.072 & 0.346 & 3.465 & 0.112 & 0.941 & 0.990 & 0.998 \\
& - & GfM (ours)  &  M+camH & \textbf{8.88M} & \textbf{0.085} & 0.554 & 4.082 & \underline{0.139} & \textbf{0.919} & \underline{0.983} & \textbf{0.995} \\
\bottomrule
\end{tabularx}

\end{subtable}

\vspace{0.5em}

\begin{subtable}{\textwidth}
\centering
\caption{Eigen benchmark~\cite{IMPROVED_KITTI_GT} @ \(\leq\)60 m}
\label{tab:eigen_60m}

\begin{tabularx}{\textwidth}{p{0.3cm}|p{2 cm} p{3cm} c c c c c c c c c }
\toprule
 & Published in & Method & Train & \#Params & Abs Rel $\downarrow$ & Sq Rel $\downarrow$ & RMSE $\downarrow$ & RMSE log $\downarrow$ & $\delta_1$ $\uparrow$ & $\delta_2$ $\uparrow$ & $\delta_3$ $\uparrow$ \\
\midrule
\multirow{4}{*}{\rotatebox[origin=c]{90}{Depth}} 
& ECCV 2024 & GroCo~\cite{GroCo2024}  &  M+camH & 34.65M & \underline{0.086} & \textbf{0.423} & \underline{3.149} & \textbf{0.129} & \underline{0.917} & \textbf{0.986} & \textbf{0.996} \\
& WACV 2025 & MonoPP~\cite{MonoPP}  &  M+camH & 34.57M & 0.087 & 0.443 & 3.199 & \textbf{0.129} & \underline{0.919} & \underline{0.985} & \textbf{0.996} \\
& ICLR 2025 & DepthPro~\cite{DepthPro2024}  &  F & 951.99M & 0.115 & \underline{0.425} & \textbf{3.057} & 0.144 & 0.881 & 0.982 & \textbf{0.996} \\
& - & GfM (ours)  &  M+camH & \textbf{8.88M} & \textbf{0.082} & 0.437 & 3.283 & \underline{0.134} & \textbf{0.927} & \textbf{0.986} & \underline{0.995} \\
\bottomrule
\end{tabularx}
\end{subtable}

\vspace{0.5em}

\begin{subtable}{\textwidth}
\centering
\caption{Eigen benchmark~\cite{IMPROVED_KITTI_GT} \(\leq\)40 m}
\label{tab:eigen_40m}

\begin{tabularx}{\textwidth}{p{0.3cm}|p{2 cm} p{3cm} c c c c c c c c c }
\toprule
 & Published in & Method & Train & \#Params & Abs Rel $\downarrow$ & Sq Rel $\downarrow$ & RMSE $\downarrow$ & RMSE log $\downarrow$ & $\delta_1$ $\uparrow$ & $\delta_2$ $\uparrow$ & $\delta_3$ $\uparrow$ \\
\midrule
\multirow{4}{*}{\rotatebox[origin=c]{90}{Depth}} 
& ECCV 2024 & GroCo~\cite{GroCo2024}  &  M+camH & 34.65M & \underline{0.081} & \underline{0.288} & \textbf{2.242} & \underline{0.118} & \underline{0.929} & \underline{0.989} & \textbf{0.997} \\
& WACV 2025 & MonoPP~\cite{MonoPP}  &  M+camH & 34.57M & \underline{0.081} & 0.293 & 2.249 & 0.117 & \underline{0.932} & 0.988 & \textbf{0.997} \\
& ICLR 2025 & DepthPro~\cite{DepthPro2024}  &  F & 951.99M & 0.112 & 0.324 & 2.256 & 0.136 & 0.894 & 0.985 & \textbf{0.997} \\
& - & GfM (ours)  &  M+camH & \textbf{8.88M} & \textbf{0.075} & \textbf{0.279} & \underline{2.246} & \textbf{0.114} & \textbf{0.942} & \textbf{0.990} & \textbf{0.997} \\
\bottomrule
\end{tabularx}
\end{subtable}

\vspace{0.5em}

\begin{subtable}{\textwidth}
\centering
\caption{Eigen benchmark~\cite{IMPROVED_KITTI_GT} with depth \(\leq\)20 m}
\label{tab:eigen_20m}

\begin{tabularx}{\textwidth}{p{0.3cm}|p{2 cm} p{3cm} c c c c c c c c c }
\toprule
 & Published in & Method & Train & \#Params & Abs Rel $\downarrow$ & Sq Rel $\downarrow$ & RMSE $\downarrow$ & RMSE log $\downarrow$ & $\delta_1$ $\uparrow$ & $\delta_2$ $\uparrow$ & $\delta_3$ $\uparrow$ \\
\midrule
\multirow{5}{*}{\rotatebox[origin=c]{90}{Depth}} 
& ECCV 2024 & GroCo~\cite{GroCo2024}  &  M+camH & 34.65M & \underline{0.067} & 0.125 & 1.141 & 0.094 & 0.955 & \underline{0.995} & \textbf{0.999} \\
& WACV 2025 & MonoPP~\cite{MonoPP}  &  M+camH & 34.57M & \underline{0.067} & \underline{0.121} & \underline{1.109} & \underline{0.093} & \underline{0.957} & 0.994 & \textbf{0.999} \\
& ICLR 2025 & DepthPro~\cite{DepthPro2024}  &  F & 951.99M & 0.105 & 0.202 & 1.362 & 0.122 & 0.913 & 0.990 & \underline{0.997} \\
& ICLR 2025 & DepthPro~\cite{DepthPro2024}~\textdagger  &  F & 951.99M & 0.051 & 0.071 & 0.888 & 0.073 & 0.982 & 0.998 & 0.999 \\
& - & GfM (ours)  &  M+camH & \textbf{8.88M} & \textbf{0.058} & \textbf{0.103} & \textbf{1.036} & \textbf{0.085} & \textbf{0.967} & \textbf{0.996} & \textbf{0.999} \\
\bottomrule
\end{tabularx}

\end{subtable}

\end{table*}

\begin{table*}[t!]
\centering
\footnotesize
\setlength\tabcolsep{3pt}
\renewcommand{\arraystretch}{1}

\definecolor{kittiColor}{HTML}{FFF5E6}
\newcommand{\kittiTag}{\textcolor{kittiColor}{\large\textbullet}}

\caption{\textbf{KITTI (Eigen split~\cite{eigen2014depth})}: benchmarking against sparse ground-truth~\cite{geiger2012kitti}. 
We report standard depth metrics. Abbreviations: \emph{M} (monocular-only), \emph{camH} (camera-height supervision), \emph{V} (velocity regression), \emph{Pose} (GT pose), \emph{SI} (size priors from internet data).}
\label{tab_app:kitti_eigen_full}

\begin{tabularx}{\textwidth}{p{0.3cm}|p{2 cm} p{3cm} c c c c c c c c c }
\toprule
 & Published in & Method & Train & \#Params & Abs Rel $\downarrow$ & Sq Rel $\downarrow$ & RMSE $\downarrow$ & RMSE log $\downarrow$ & $\delta_1$ $\uparrow$ & $\delta_2$ $\uparrow$ & $\delta_3$ $\uparrow$ \\
\midrule
\multirow{12}{*}{\rotatebox[origin=c]{90}{Depth~\cite{geiger2012kitti}}}
& ICCV 2019 & Monodepth2~\cite{godard2019digging_monodepth2_selfsup,MonoPP}  &  M+camH & 14.3M & 0.126  & 0.973  & 4.880  & 0.198  & 0.864  & 0.957  & 0.980  \\
& IROS 2020 & DNet~\cite{DNet}  &  M+camH & 14.3M & 0.118  & 0.925 & 4.918  & 0.199 & 0.862 & 0.953  & 0.979 \\
& CVPR 2020 & PackNet~\cite{packnet_selfsup}  &  M+V & 128M & 0.111  & 0.829 & 4.788 & 0.199 & 0.864  & 0.954 & 0.980\\
& IROS 2021 & Wagstaff~\etal~\cite{wagstaff_scalerecovery_2021_SELFSUP}  &  M+Pose & 14.3M & 0.123  & 0.996 & 5.253 & 0.213 & 0.840  & 0.947 & 0.978 \\
& IROS 2021 & Wagstaff~\etal~\cite{wagstaff_scalerecovery_2021_SELFSUP} &  M+camH & 14.3M & 0.155  & 1.657 & 5.615 & 0.236 & 0.809  & 0.924 & 0.959 \\
& arXiv 2021 & Sui~\etal~\cite{roadaware_SFM_2021_selfsup_scaled}  &  M+camH & 14.3M & 0.128  & 0.936 & 5.063  & 0.214 & 0.847 & 0.951  & 0.978 \\
& RA-L 2022 & VADepth~\cite{VADEPTH}  &  M+camH & 18.8M & \underline{0.109}  & \textbf{0.785} & \textbf{4.624} & \underline{0.190} & \underline{0.875}  & \underline{0.960} & \textbf{0.982} \\
& ECCV 2022 & DynaDepth~\cite{dynadepth} &  M+Pose & 30M & \underline{0.109}  & \underline{0.787} & \underline{4.705} & {0.195} & 0.869  & 0.958 & \underline{0.981} \\
& ECCV 2024 & FUMET~\cite{CameraHeightDoesnChange}  &  M+SI & 25.6M & 0.108 & \textbf{0.785} & 4.736 & 0.195 & 0.871  & 0.958 & \underline{0.981} \\
& WACV 2025 & MonoPP~\cite{MonoPP}  &  M+camH & 25.6M & \textbf{0.107}  & 0.835 & {4.658}  & \textbf{0.186}  & \textbf{0.891}  &  \textbf{0.962}  & {0.982} \\
& ECCV 2024 & GroCo~\cite{GroCo2024}  &  M+camH & 34M & 0.113  & 0.851 & 4.756  & 0.197  & 0.870  &  0.958  & 0.980 \\
& - & GfM (ours)  &  M+camH & \textbf{8.8M} &   \textbf{0.107}  &   0.808  &   4.918  &   0.194  &   0.870  &   0.957  &   \underline{0.981} \\
\bottomrule
\end{tabularx}
\end{table*}

\section{Extra Qualitative Results}
\label{sec:app_qualitative_results}

\begin{figure*}
  \centering
  \includegraphics[width=0.8\linewidth]{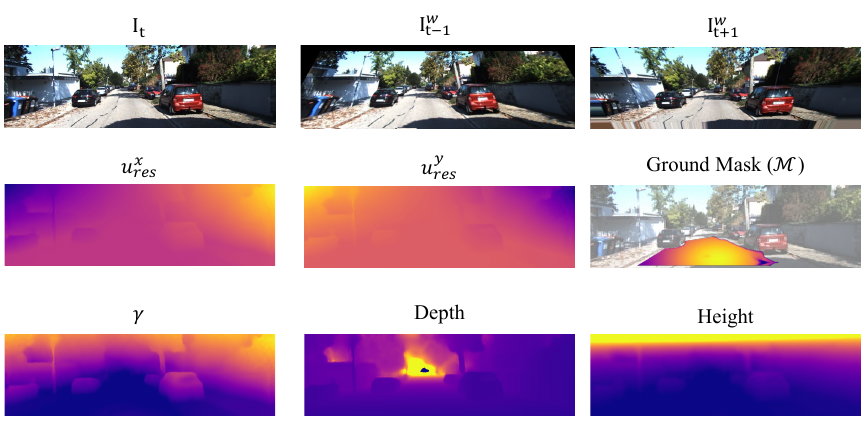}
\caption{Example of the full processing pipeline, from warped images and intermediate parallax flow to the final computed outputs.}
  \label{fig:full_pipeline}
\end{figure*}

\Cref{fig:full_pipeline} illustrates our full processing pipeline, showcasing input images, warped views, intermediate parallax flow, and final outputs. Additional qualitative comparisons in \Cref{tab:qualitative_results_appendix_good} highlight our model's ability to estimate accurate height-to-depth (\(\gamma\)) and depth predictions, with error maps as a visual aid for the correct and incorrect predictions. \Cref{tab:qualitative_results_appendix_bad} presents failure cases, where faraway dynamic objects pose challenges, explaining our strong \textit{Abs. Rel.} depth performance.

\begin{figure}[t]
  \centering
  \begin{subfigure}{0.48\columnwidth}
    \centering
        {\includegraphics[width=\linewidth]{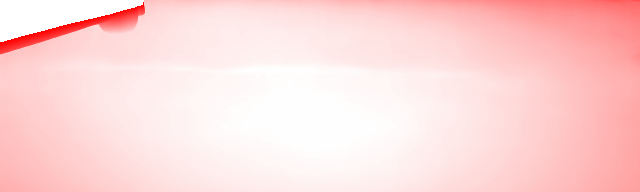}};
    \caption{MoGe2-L~\cite{wang2025moge2}}
    \label{fig:height_moge}
  \end{subfigure}\hfill
  \begin{subfigure}{0.48\columnwidth}
    \centering
        {\includegraphics[width=\linewidth]{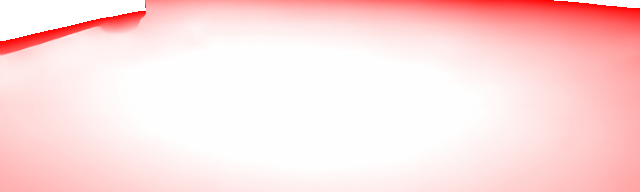}};
    \caption{UniDepthV2-L~\cite{unidepthv2}}
    \label{fig:height_unidepth}
  \end{subfigure}

  \vspace{0.5em}

  \begin{subfigure}{0.48\columnwidth}
    \centering
        {\includegraphics[width=\linewidth]{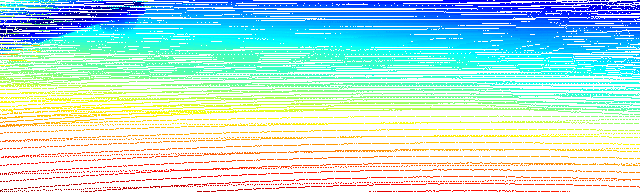}};
    \caption{$\gamma$ error map (MoGe2-L)}
    \label{fig:height_depthpro}
  \end{subfigure}\hfill
  \begin{subfigure}{0.48\columnwidth}
    \centering
        {\includegraphics[width=\linewidth]{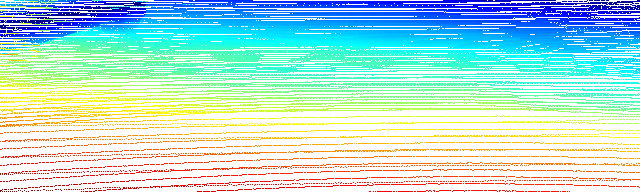}};
    \caption{$\gamma$ error map (UniDepthV2-L)}
    \label{fig:height_gfm}
  \end{subfigure}

  \caption{\textbf{Clipped height visualization on the same RSRD example presented in~\cref{fig:Qualitative_big_results} (b)}
  GfM better preserves fine road topography (local slopes, camber, and shallow undulations), while foundational models exhibit visible scale inconsistencies. This supports our finding that \emph{self‑supervised fine‑tuning remains useful} when moving to new scenes and environments. 
  All panels use the same colormap and clipping range for comparability.}
  \label{fig:height_compare}
\end{figure}

\begin{figure*}[t]
    \centering

    \begin{subfigure}[b]{0.9\textwidth}
        \centering
        \includegraphics[width=0.45\linewidth]{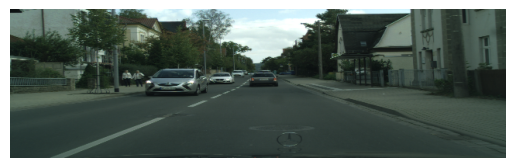}
        \includegraphics[width=0.45\linewidth]{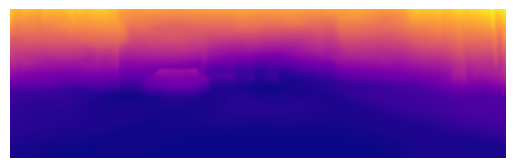}
        \caption{Example 1: Input image from Cityscapes~\cite{cordts2016cityscapes} (left) and corresponding $\gamma$ output (right)}
    \end{subfigure}

    \vspace{0.5cm} 
    
    \begin{subfigure}[b]{0.9\textwidth}
        \centering
        \includegraphics[width=0.45\linewidth]{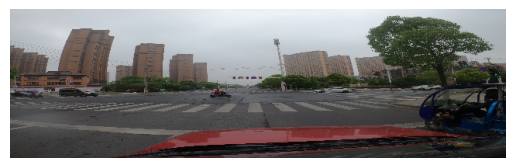}
        \includegraphics[width=0.45\linewidth]{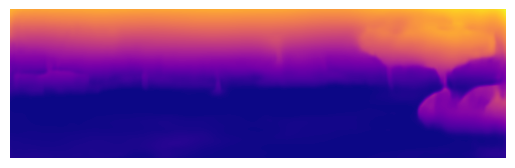}
        \caption{Example 2: Input image from~\cite{cloudy_day_dashcam} (left) and corresponding $\gamma$ output (right)}
    \end{subfigure}

    \vspace{0.5cm} 
    

    \caption{\textbf{Zero-shot evaluation on new datasets and camera setups.} Although our model is trained exclusively on KITTI, we evaluate its zero-shot generalization on different dataset examples with unseen camera setups and camera intrinsics. The model has never encountered these specific distortions or intrinsic parameters during training. Each row presents an input image (left) and its corresponding $\gamma$ output (right).}

    \label{fig:zero_shot_results_appendix}
\end{figure*}

To further analyze our method, \Cref{fig:qualtiative_comparison_appendix} directly compares our predictions with GroCo~\cite{GroCo2024} and DepthPro~\cite{DepthPro2024}, illustrating differences in \(\gamma\) estimation, depth predictions, and error maps. For the RSRD dataset, \Cref{fig:rsrd_pointcloud_comparison_appendix} visualizes 3D point clouds, showing how our method better preserves road geometry, whereas DepthPro misinterprets road slopes. In addition, we evaluated recent metric foundation models (MoGe2-L~\cite{wang2025moge2} and UniDepthV2-L~\cite{unidepthv2}) on RSRD. As illustrated in \cref{fig:height_compare}, these models tend to impose an overly planar prior on the drivable surface, visibly flattening local slopes, camber, and shallow undulations, yielding near-zero $\gamma$ where subtle elevation exists. We did not conduct a dedicated generalization study for our $\gamma$ model, just qualitative examples as presented in~\cref{fig:zero_shot_results_appendix}. However, a simple online fine-tuning pass (few frames, self-supervised) already recovers much of the near-field relief and reduces $\gamma$ errors.

\begin{figure*}[t]
    \centering

    \begin{subfigure}[b]{0.9\textwidth}
        \centering
        \includegraphics[width=0.45\linewidth]{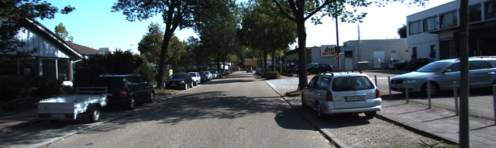}
        \includegraphics[width=0.45\linewidth]{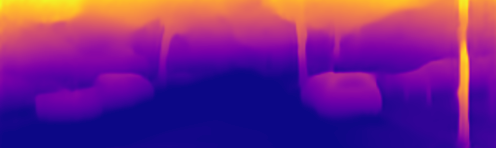}
        \caption{Original input (left) and corresponding $\gamma$ output (right)}
    \end{subfigure}

    \vspace{0.5cm} 
    
    \begin{subfigure}[b]{0.9\textwidth}
        \centering
        \includegraphics[width=0.45\linewidth]{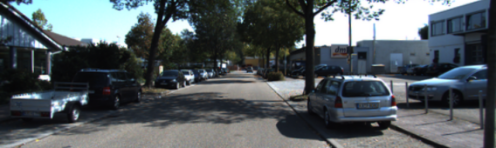}
        \includegraphics[width=0.45\linewidth]{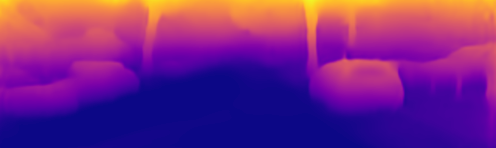}
        \caption{X-axis rotation (\(-2^\circ\)): Input (left) and $\gamma$ output (right)}
    \end{subfigure}

    \vspace{0.5cm} 

    \begin{subfigure}[b]{0.9\textwidth}
        \centering
        \includegraphics[width=0.45\linewidth]{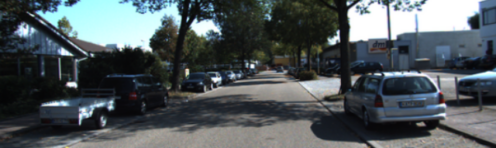}
        \includegraphics[width=0.45\linewidth]{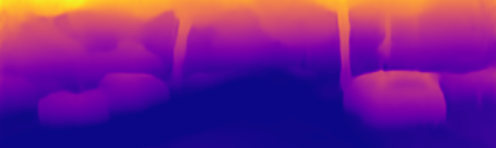}
        \caption{Y-axis rotation (\(-5^\circ\)): Input (left) and $\gamma$ output (right)}
    \end{subfigure}

    \vspace{0.5cm} 
    
    \begin{subfigure}[b]{0.9\textwidth}
        \centering
        \includegraphics[width=0.45\linewidth]{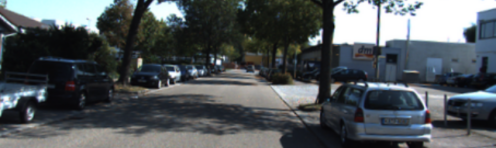}
        \includegraphics[width=0.45\linewidth]{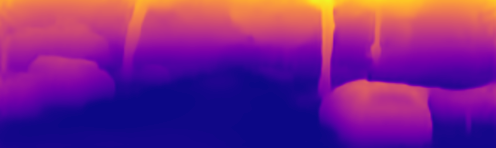}
        \caption{Z-axis rotation (\(-5^\circ\)): Input (left) and $\gamma$ output (right)}
    \end{subfigure}

    \caption{\textbf{Effect of synthetic rotations on $\gamma$ estimation.} Although our model was never trained on such transformations, we evaluate its behavior under synthetic rotations following~\cite{GroCo2024}. Each row presents an input image (left) with a rotation around the X, Y, or Z axis, alongside its corresponding $\gamma$ output (right).}

    \label{fig:rotation_effect_appendix}
\end{figure*}

We assess the model's robustness to synthetic rotations in \Cref{fig:rotation_effect_appendix} and its zero-shot generalization in \Cref{fig:zero_shot_results_appendix}, showing that it can adapt to unseen datasets and camera setups despite being trained only on KITTI. These results validate the model's performance across various conditions while indicating areas for improvement in robustness and domain adaptation.

\begin{table*}[t]
    \centering
    \small
    \renewcommand{\arraystretch}{1.2}
    \setlength{\tabcolsep}{4pt}
    \begin{tabular}{m{0.17\textwidth} m{0.17\textwidth} m{0.17\textwidth} m{0.17\textwidth} m{0.20\textwidth}}
        \toprule
        \textbf{Input Image} & \textbf{Gamma Output} & \textbf{Gamma Error} & \textbf{Depth Output} & \textbf{Depth Error} \\
        \midrule
        \includegraphics[width=\linewidth]{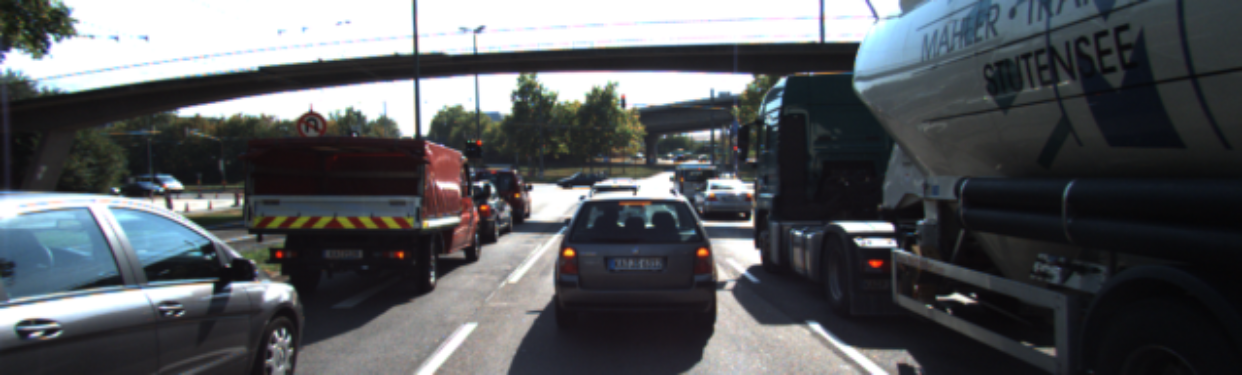} &
        \includegraphics[width=\linewidth]{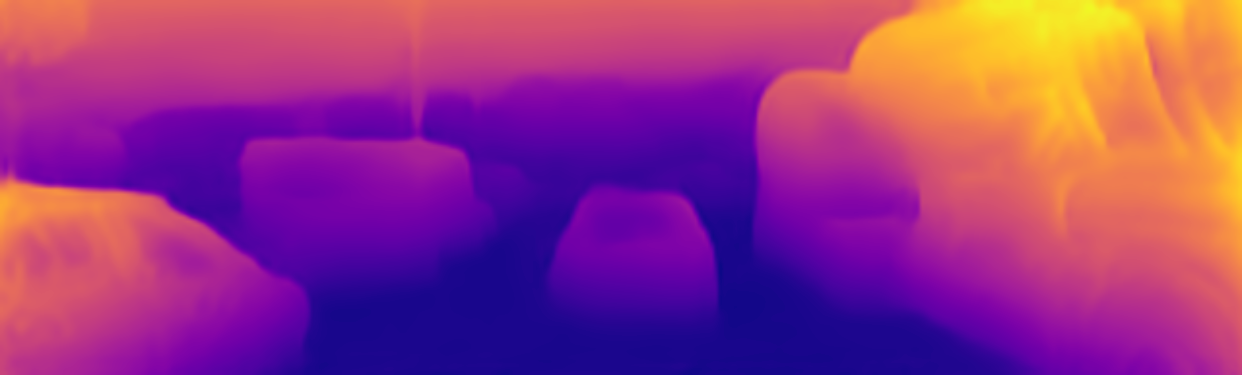} &
        \includegraphics[width=\linewidth]{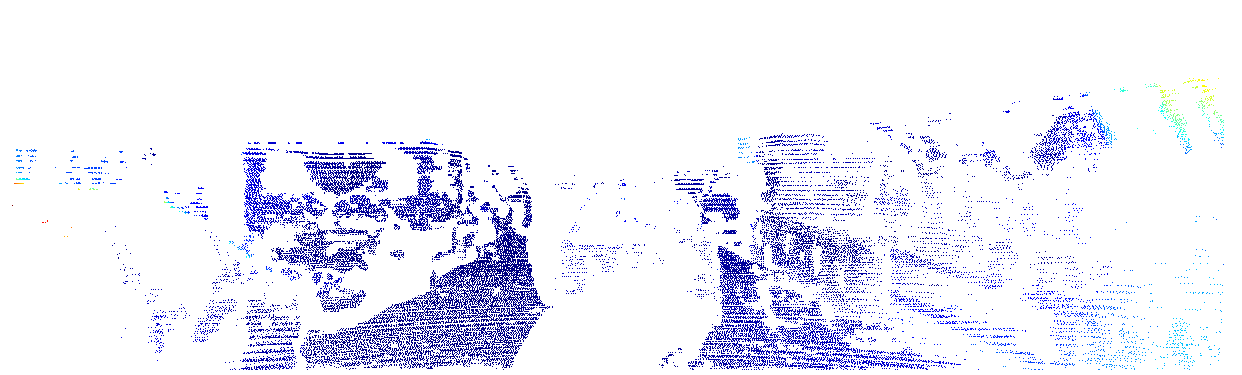} &
        \includegraphics[width=\linewidth]{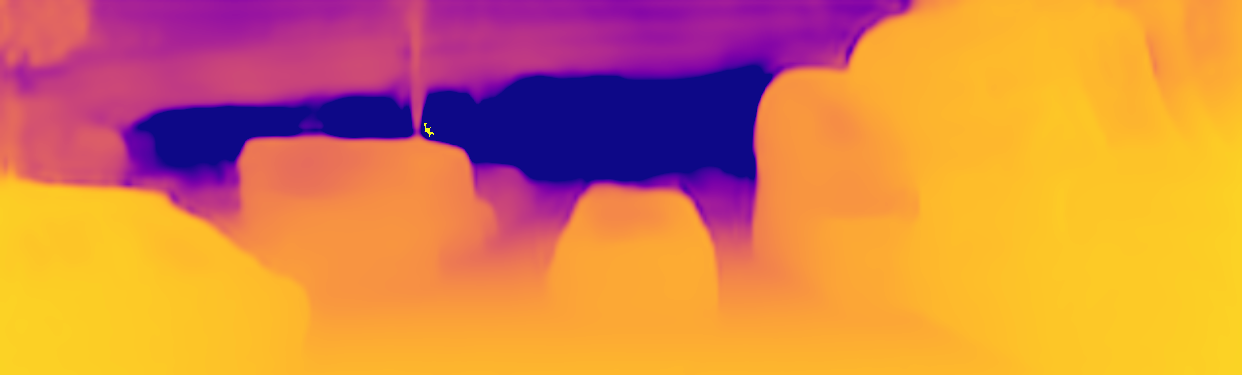} &
        \includegraphics[width=\linewidth]{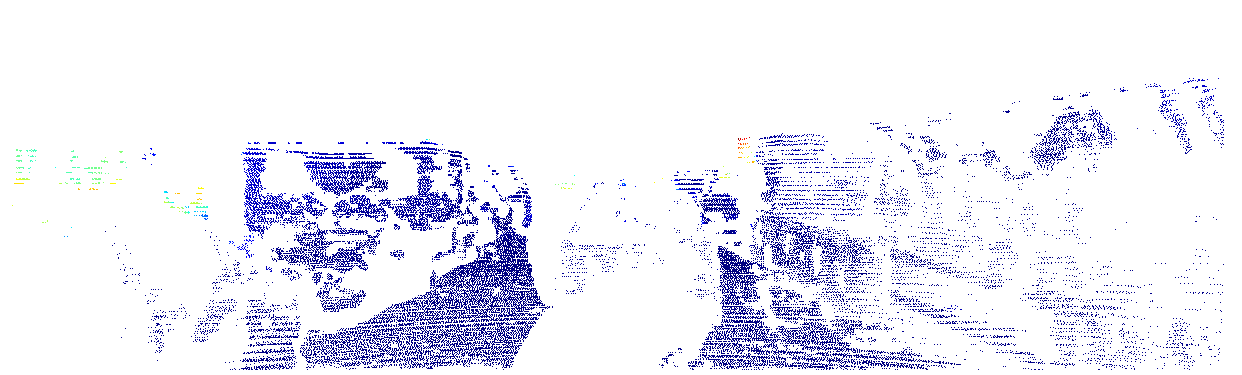} \\

        \includegraphics[width=\linewidth]{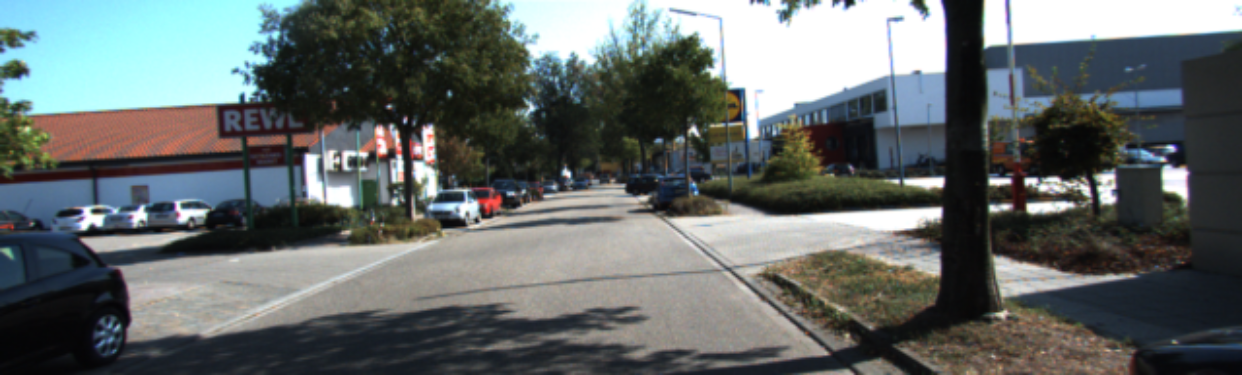} &
        \includegraphics[width=\linewidth]{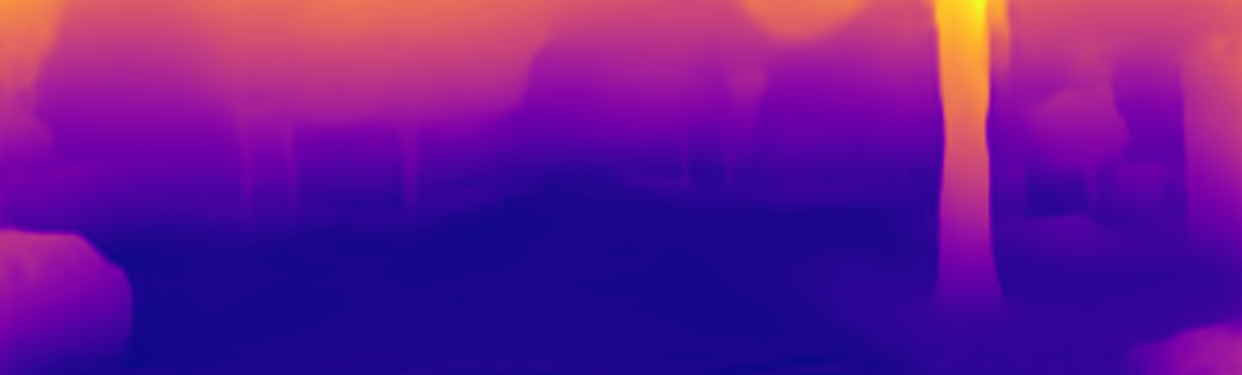} &
        \includegraphics[width=\linewidth]{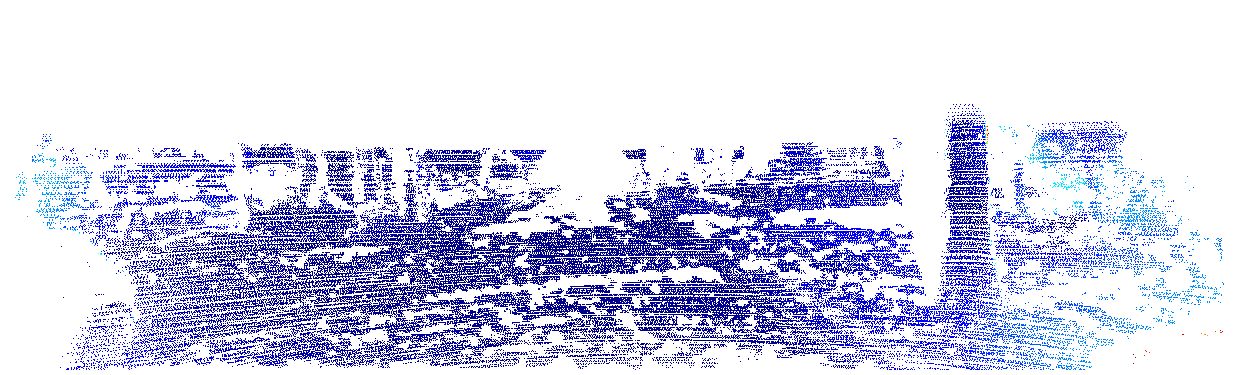} &
        \includegraphics[width=\linewidth]{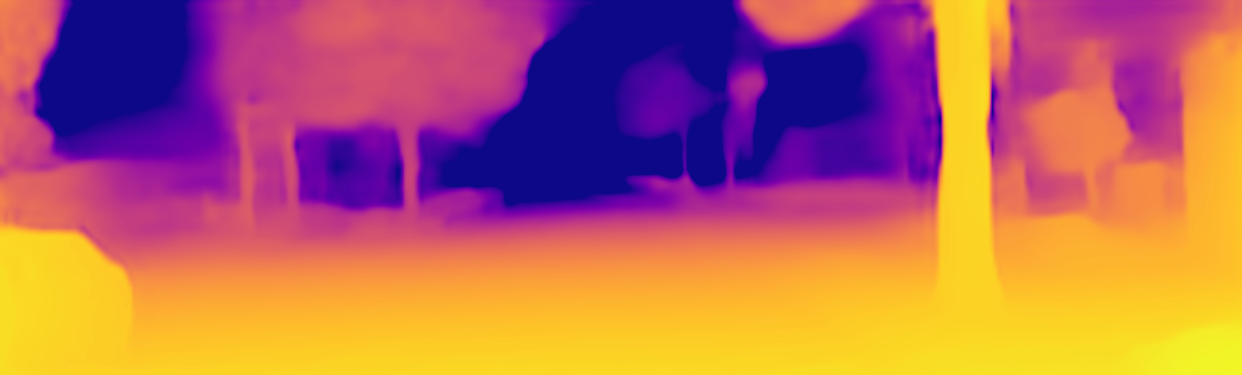} &
        \includegraphics[width=\linewidth]{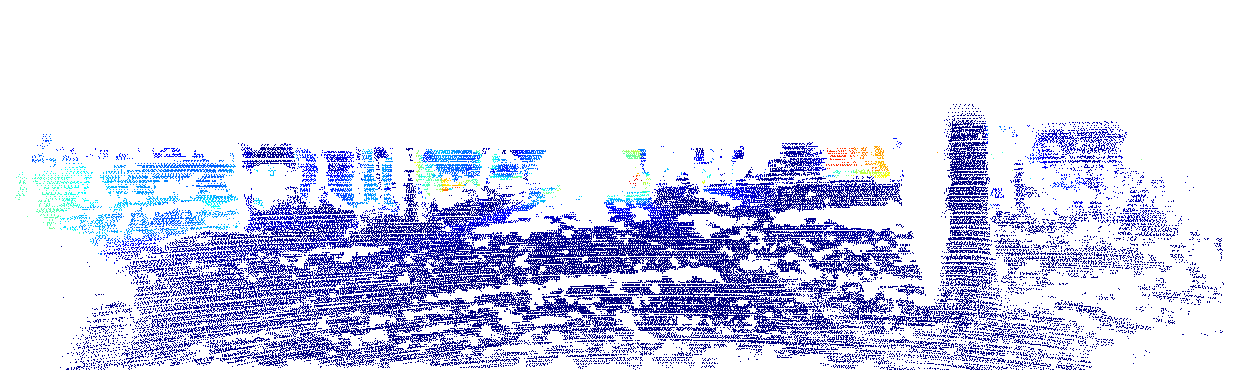} \\
        
        \includegraphics[width=\linewidth]{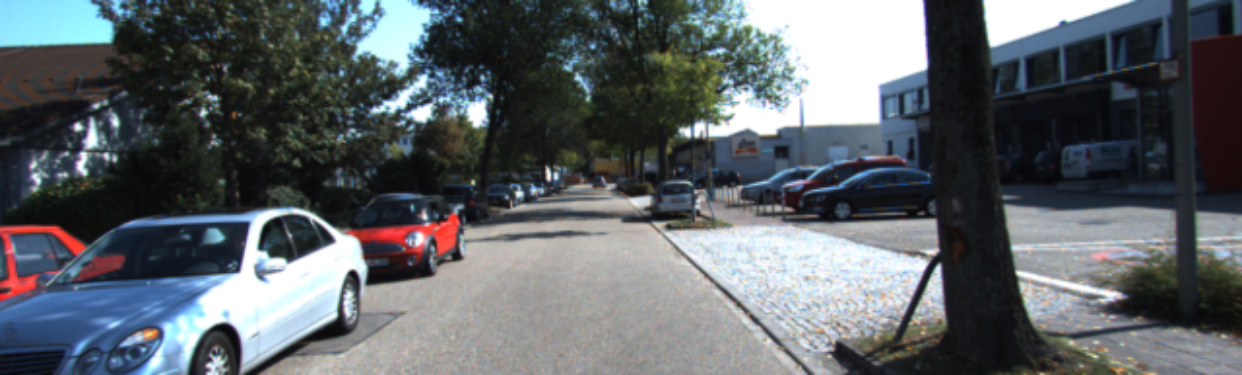} &
        \includegraphics[width=\linewidth]{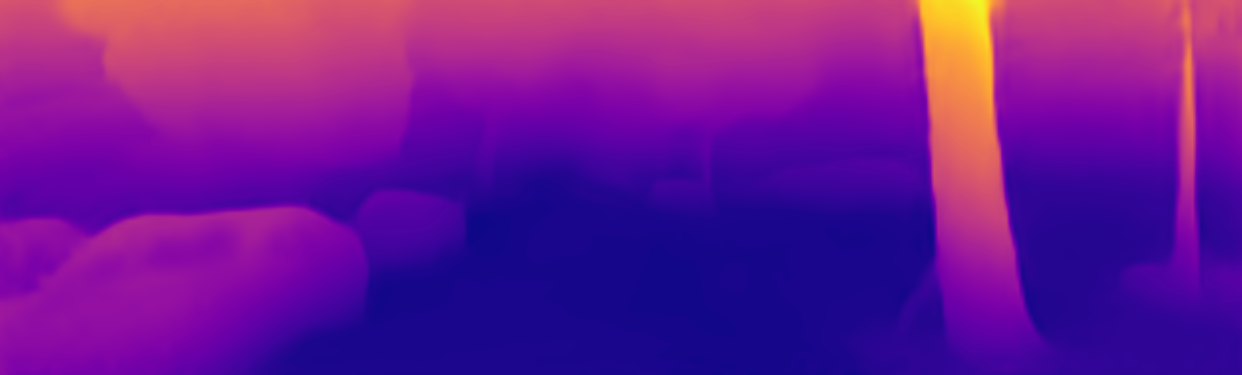} &
        \includegraphics[width=\linewidth]{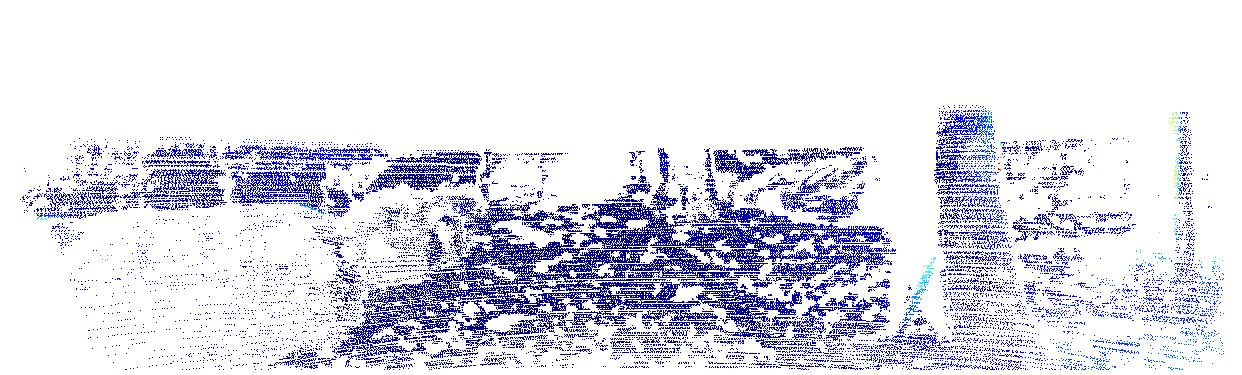} &
        \includegraphics[width=\linewidth]{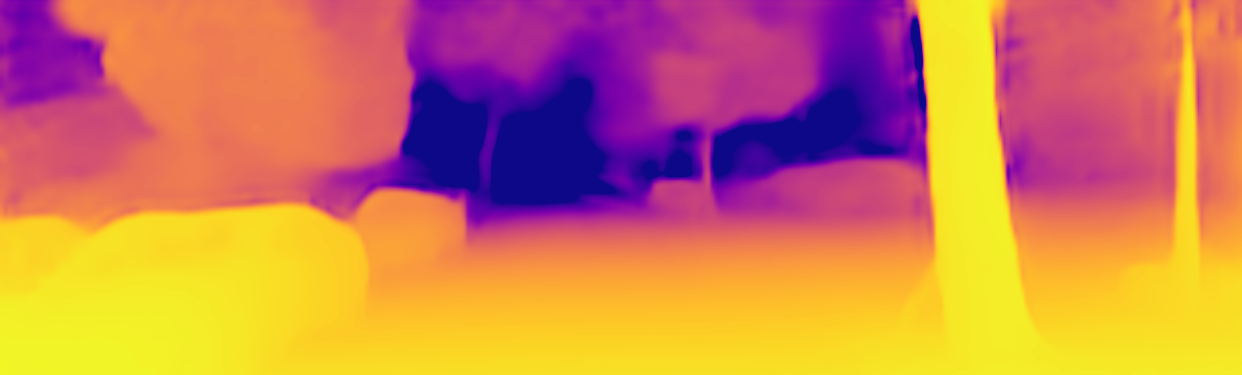} &
        \includegraphics[width=\linewidth]{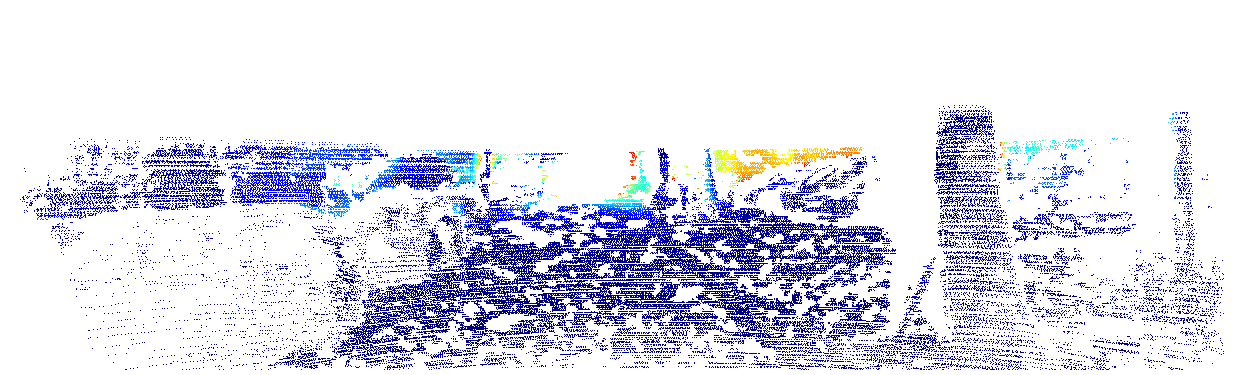} \\
        
        \includegraphics[width=\linewidth]{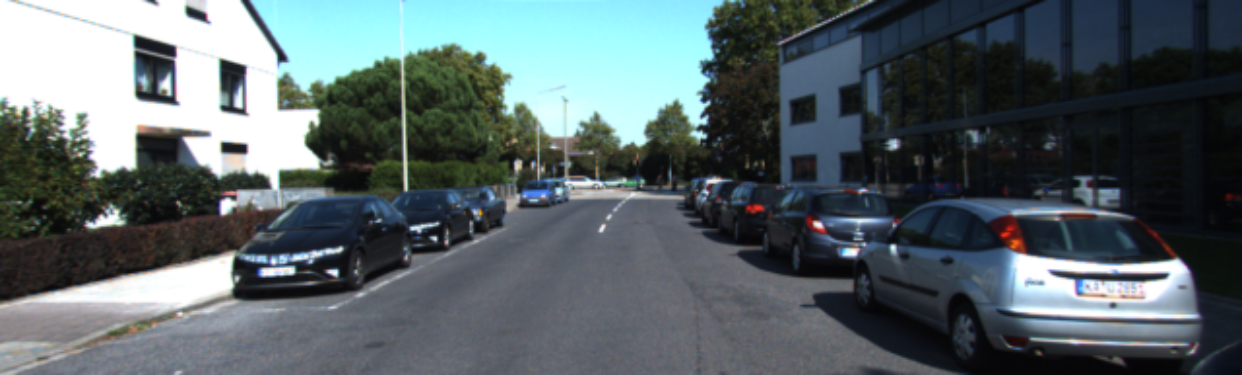} &
        \includegraphics[width=\linewidth]{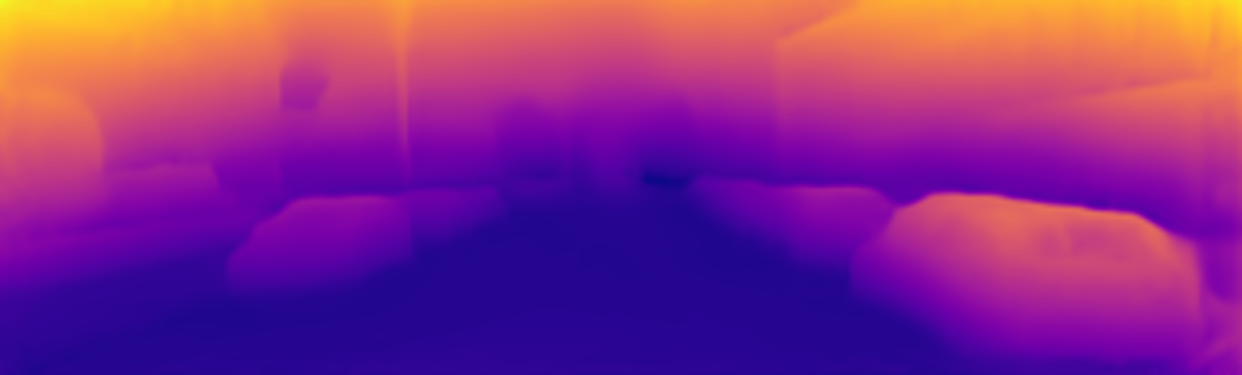} &
        \includegraphics[width=\linewidth]{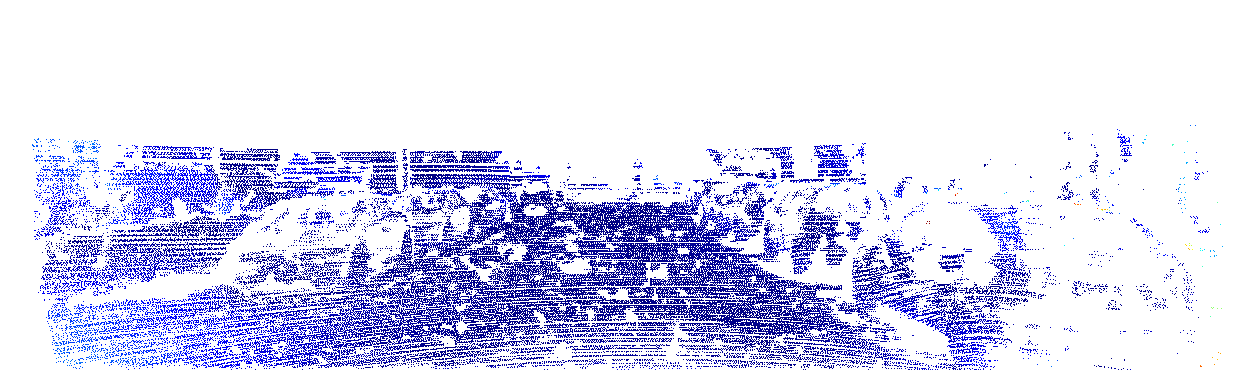} &
        \includegraphics[width=\linewidth]{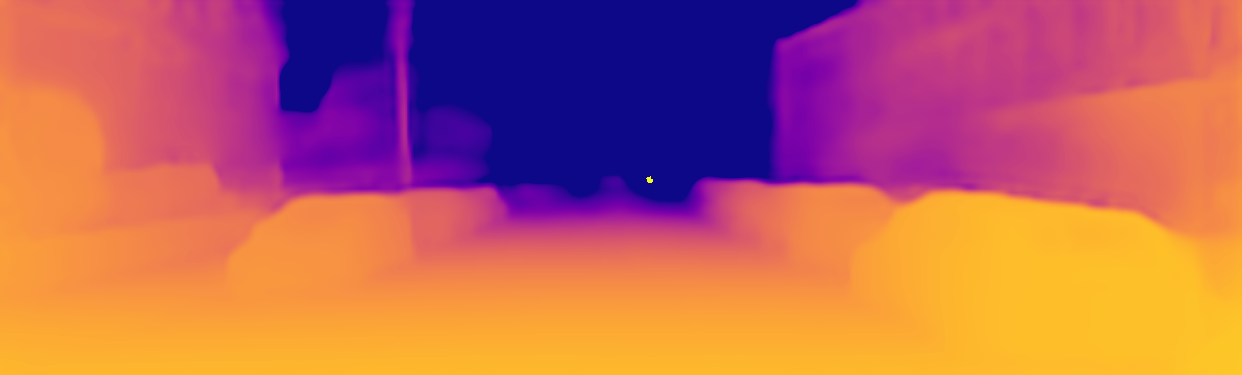} &
        \includegraphics[width=\linewidth]{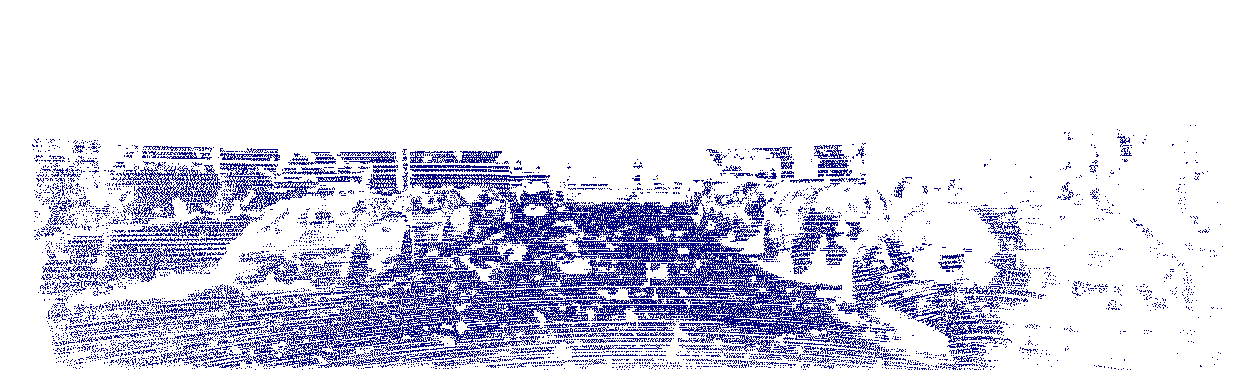} \\
        
        \includegraphics[width=\linewidth]{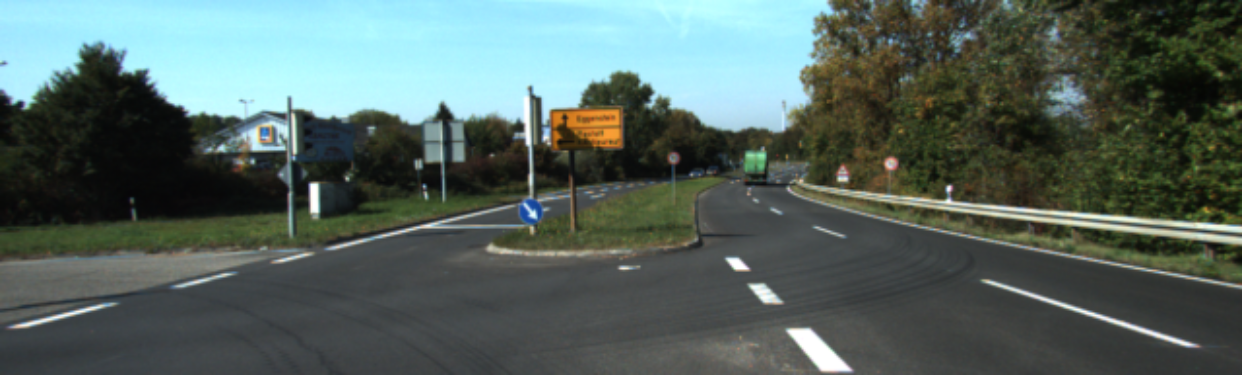} &
        \includegraphics[width=\linewidth]{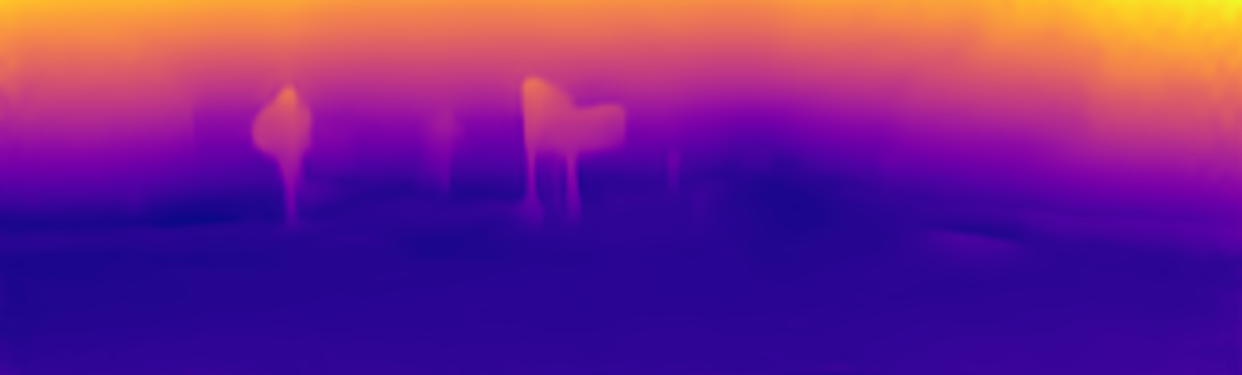} &
        \includegraphics[width=\linewidth]{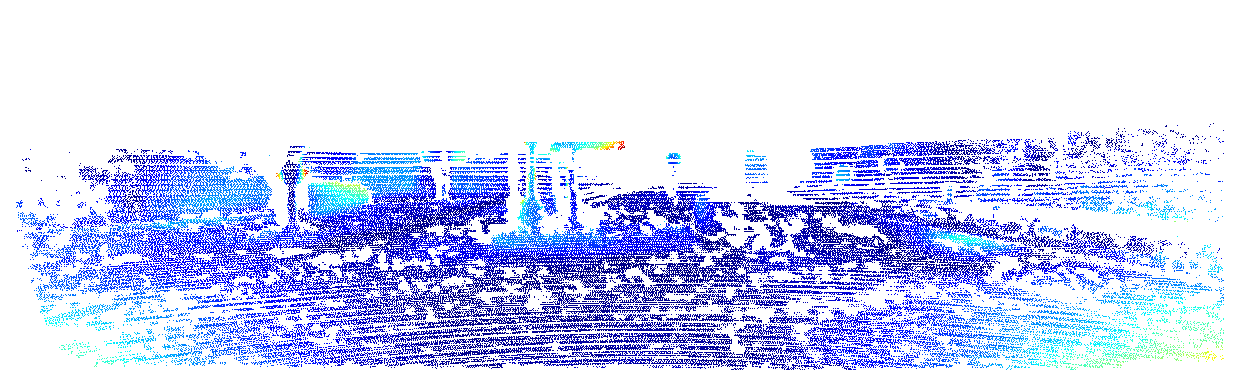} &
        \includegraphics[width=\linewidth]{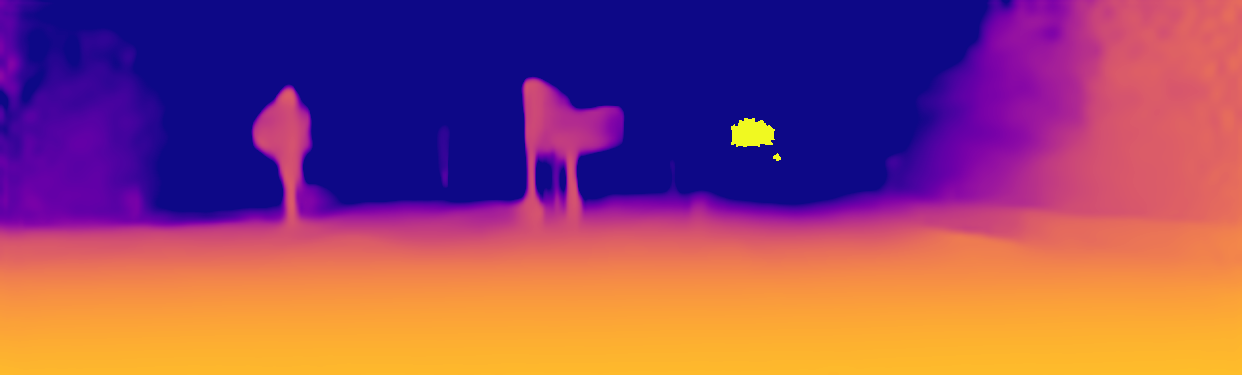} &
        \includegraphics[width=\linewidth]{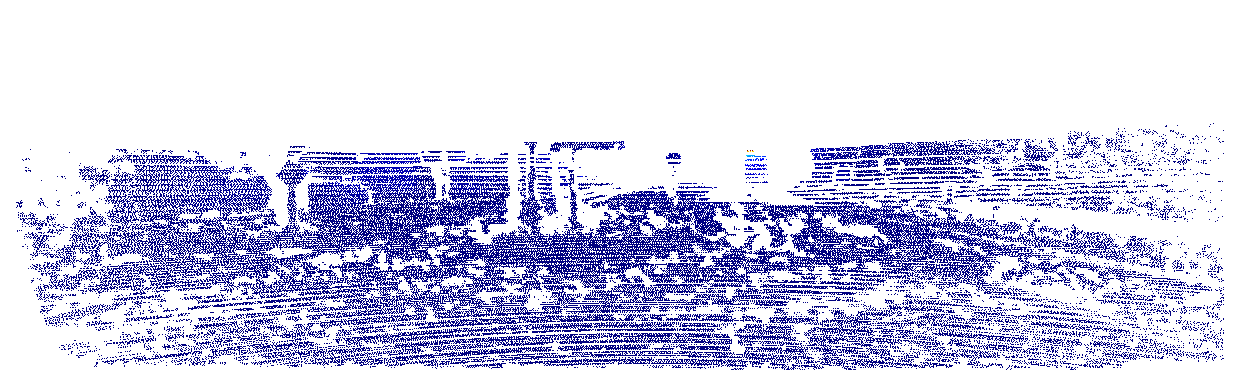} \\
        
        \includegraphics[width=\linewidth]{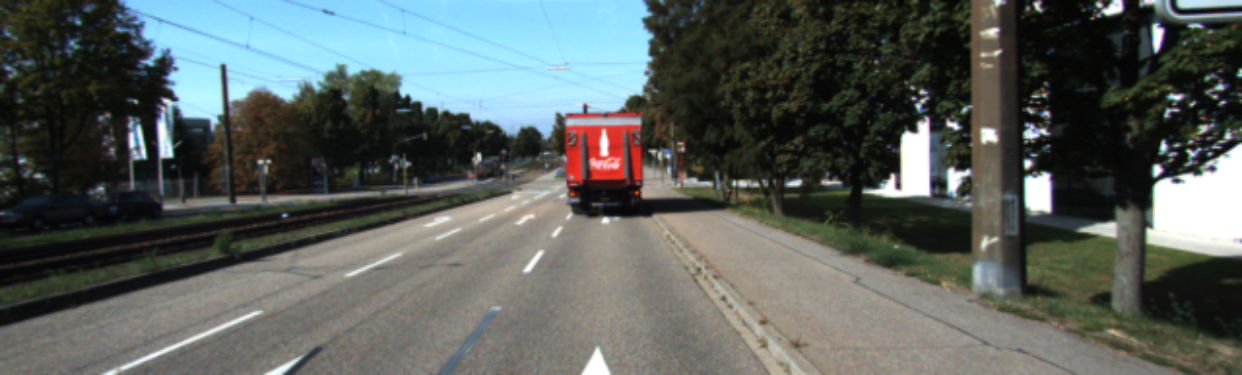} &
        \includegraphics[width=\linewidth]{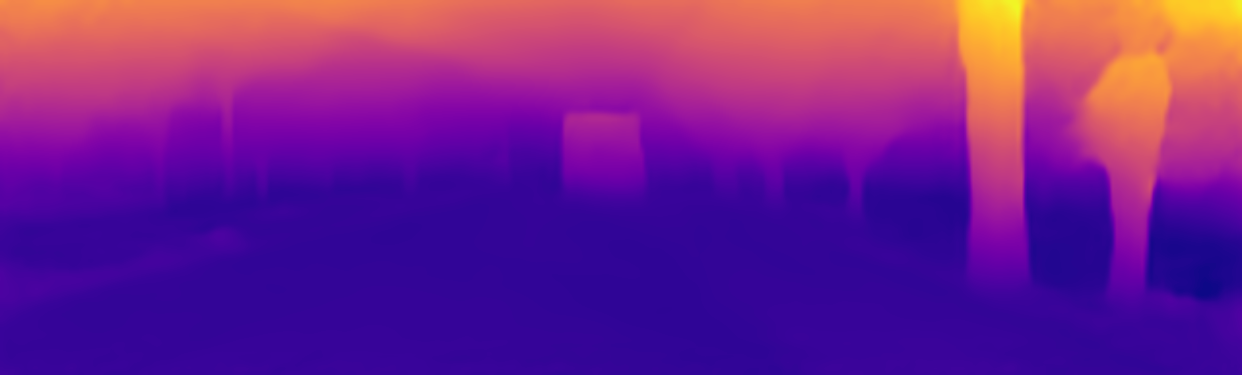} &
        \includegraphics[width=\linewidth]{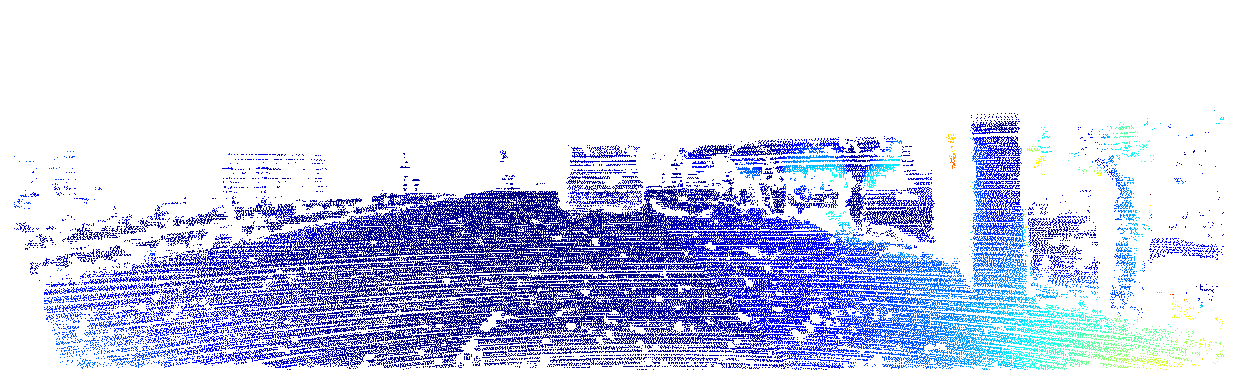} &
        \includegraphics[width=\linewidth]{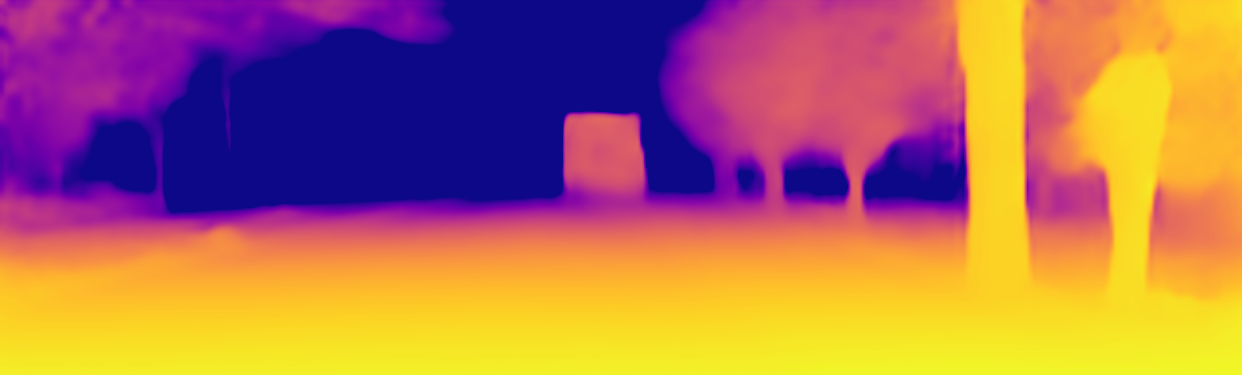} &
        \includegraphics[width=\linewidth]{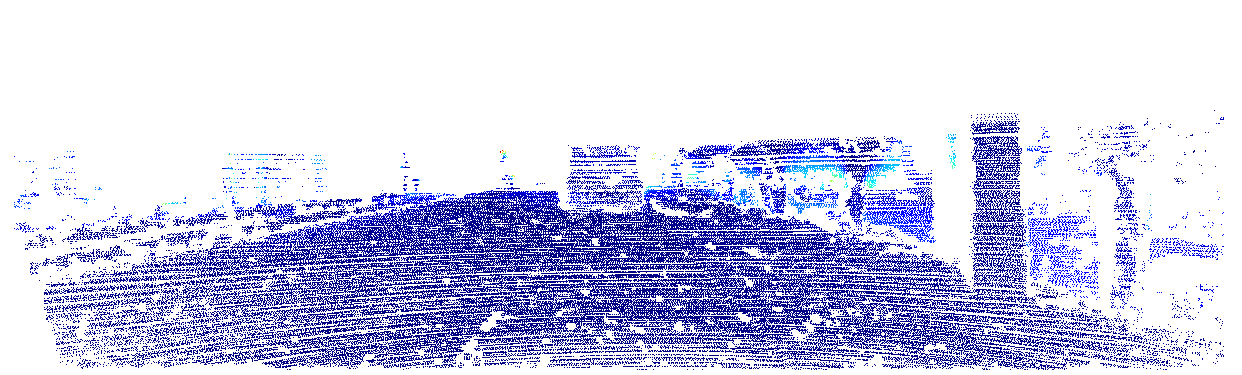} \\
        
        \includegraphics[width=\linewidth]{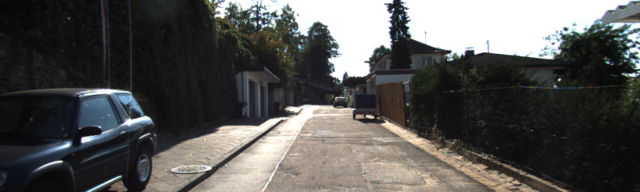} &
        \includegraphics[width=\linewidth]{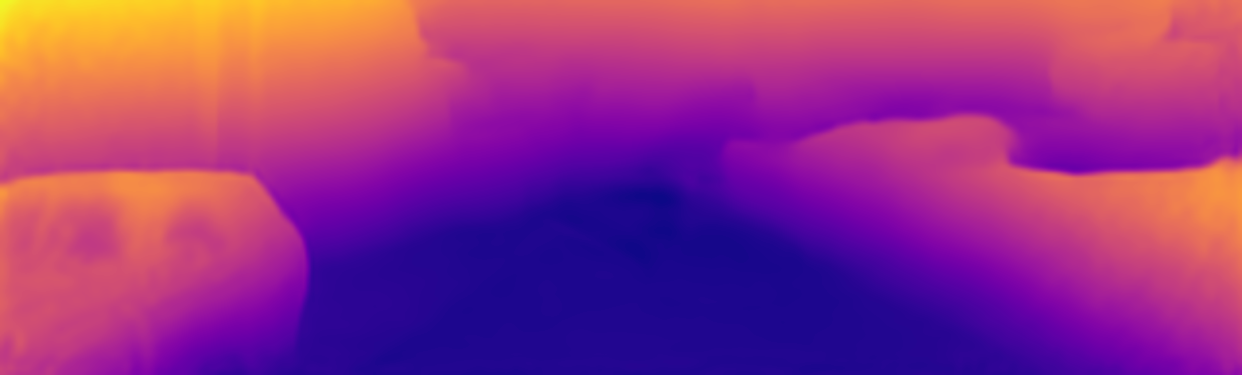} &
        \includegraphics[width=\linewidth]{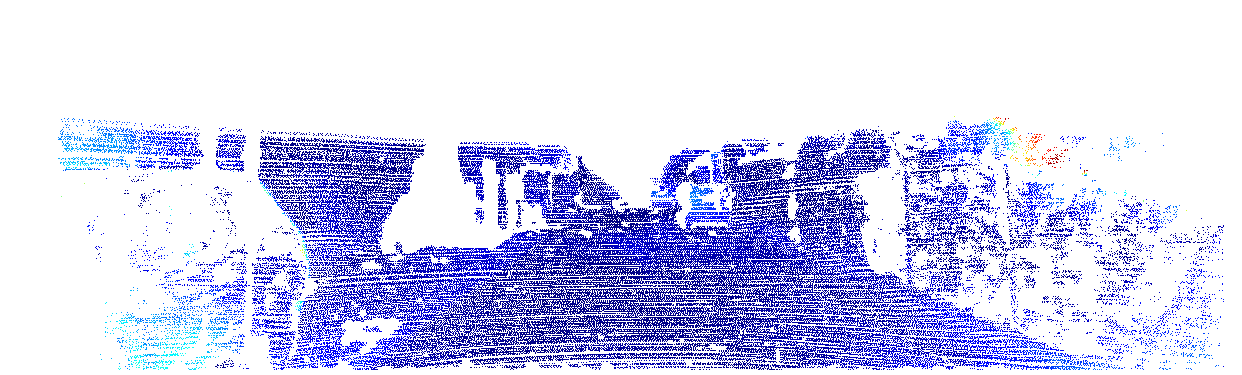} &
        \includegraphics[width=\linewidth]{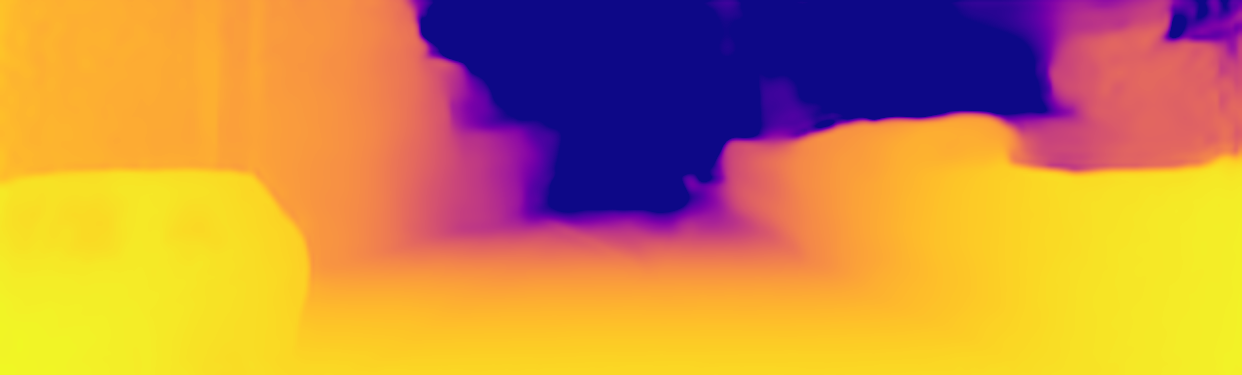} &
        \includegraphics[width=\linewidth]{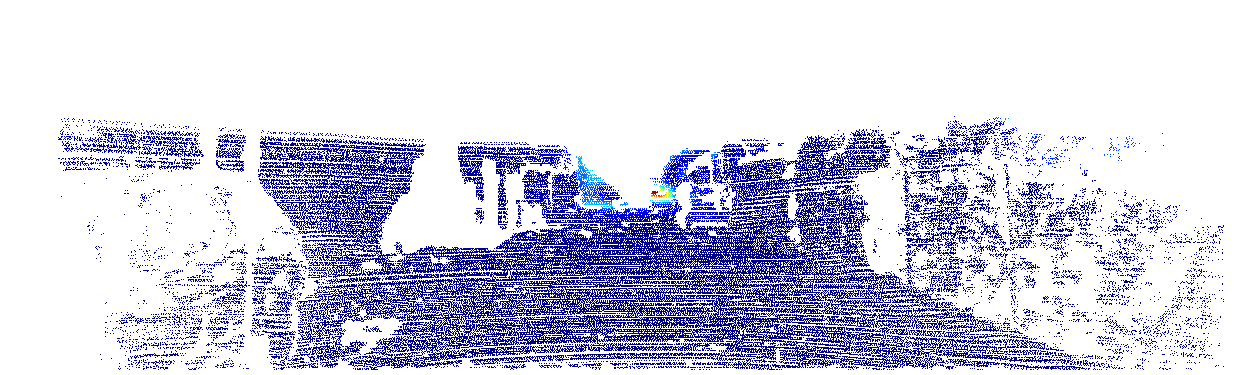} \\

        \includegraphics[width=\linewidth]{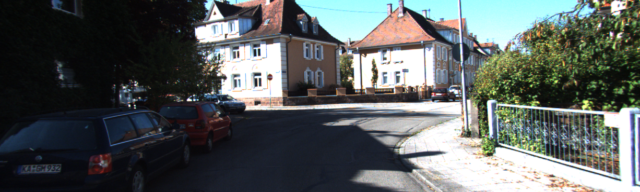} &
        \includegraphics[width=\linewidth]{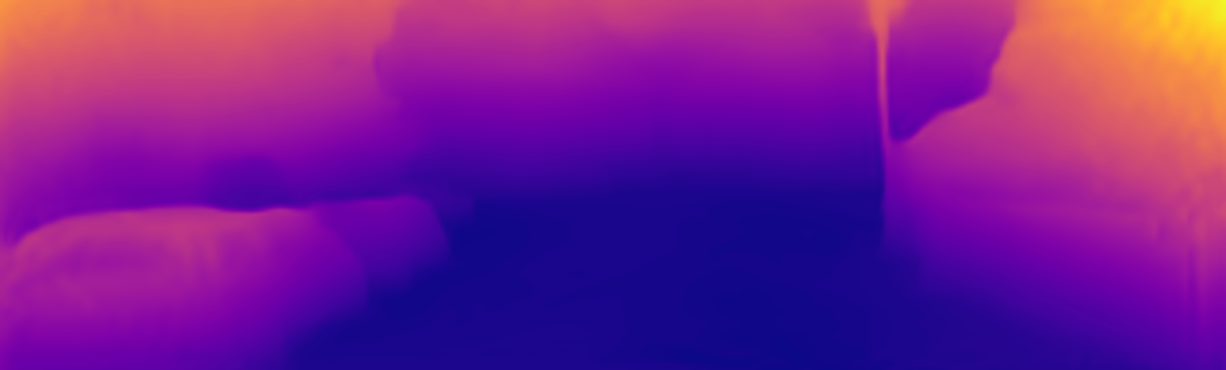} &
        \includegraphics[width=\linewidth]{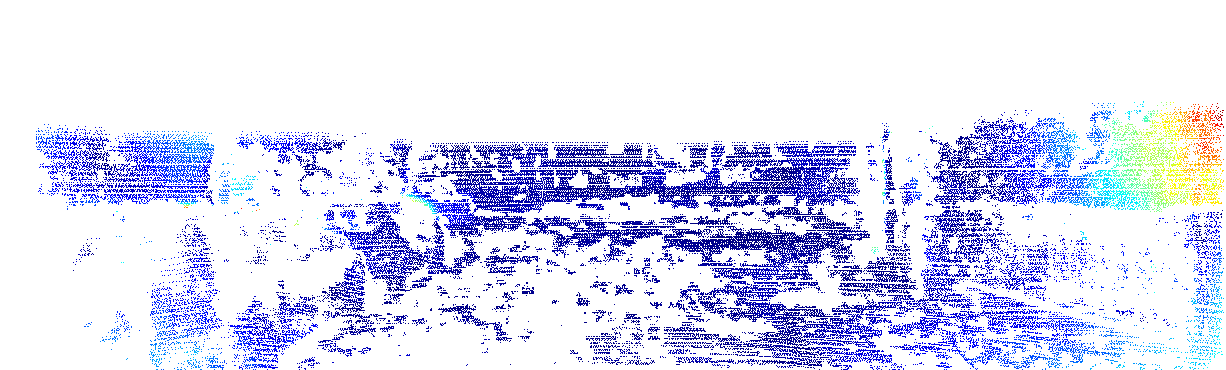} &
        \includegraphics[width=\linewidth]{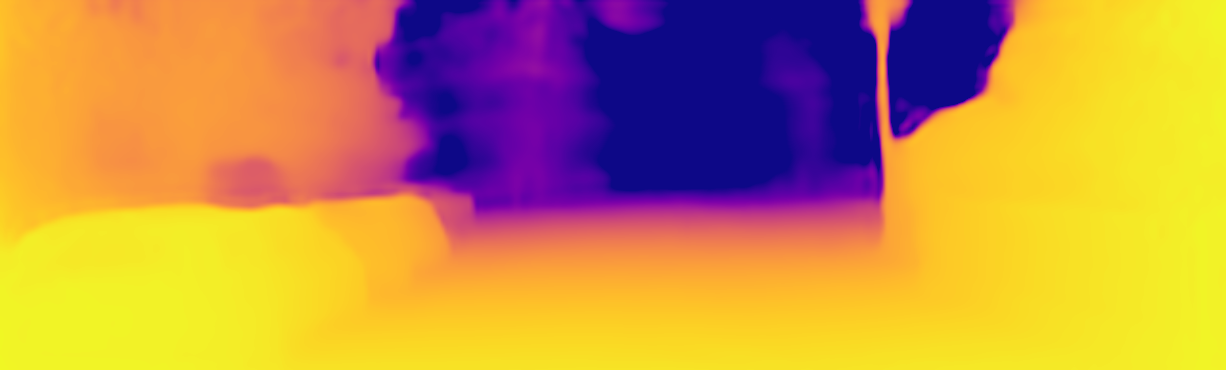} &
        \includegraphics[width=\linewidth]{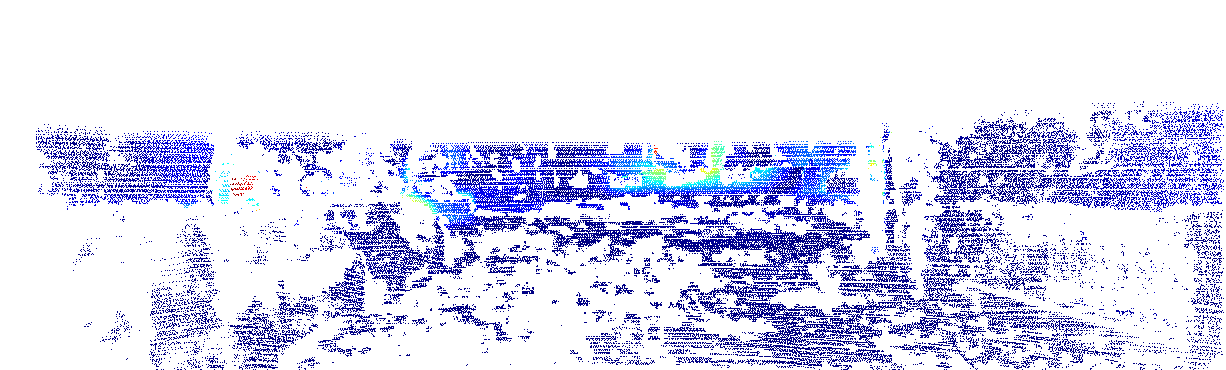} \\

        \includegraphics[width=\linewidth]{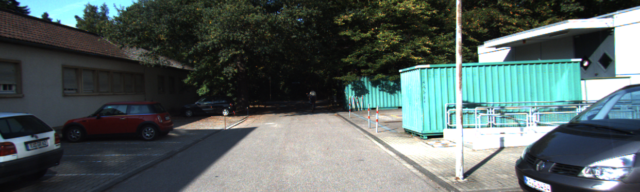} &
        \includegraphics[width=\linewidth]{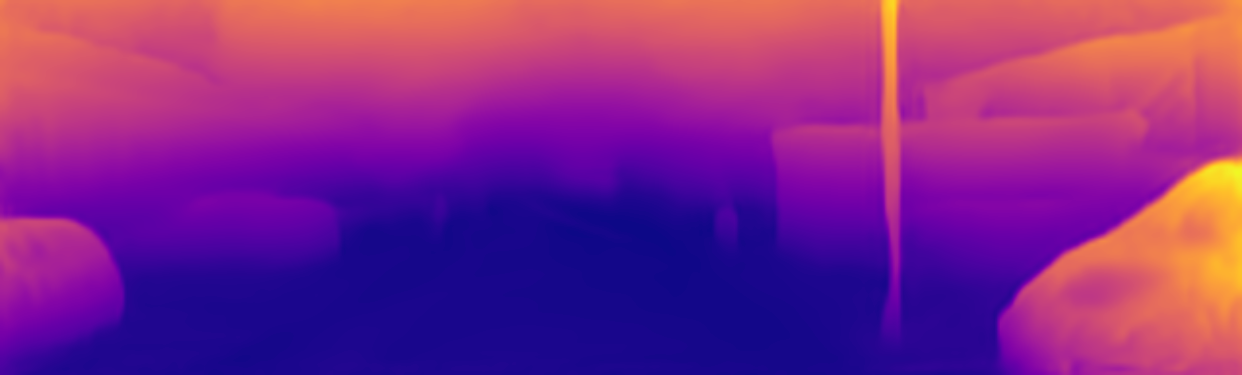} &
        \includegraphics[width=\linewidth]{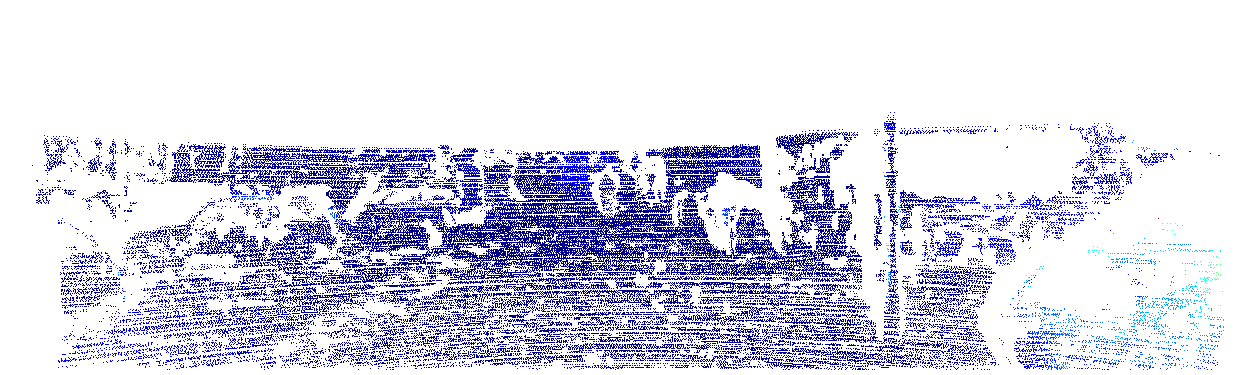} &
        \includegraphics[width=\linewidth]{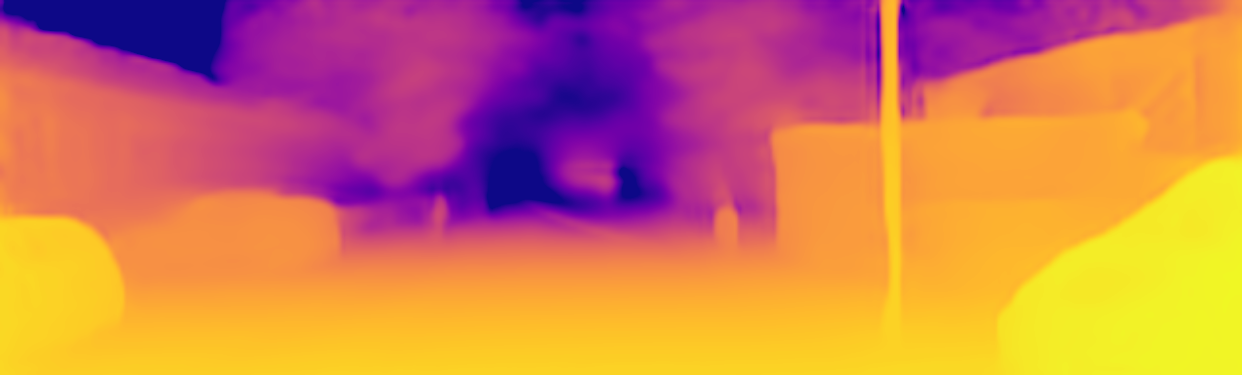} &
        \includegraphics[width=\linewidth]{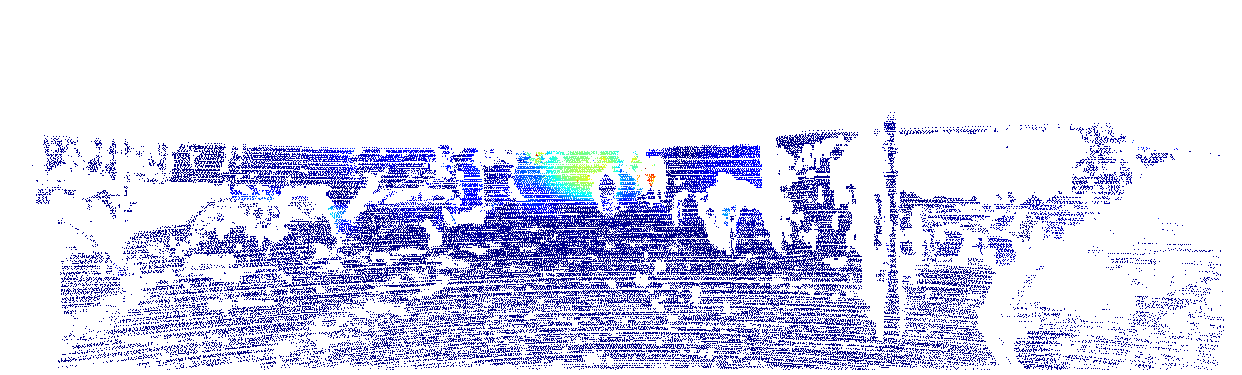} \\

        \includegraphics[width=\linewidth]{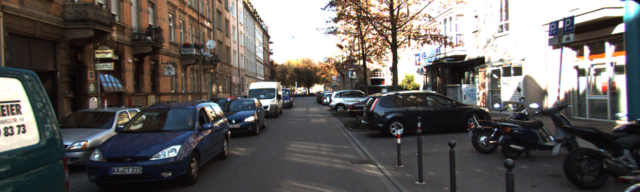} &
        \includegraphics[width=\linewidth]{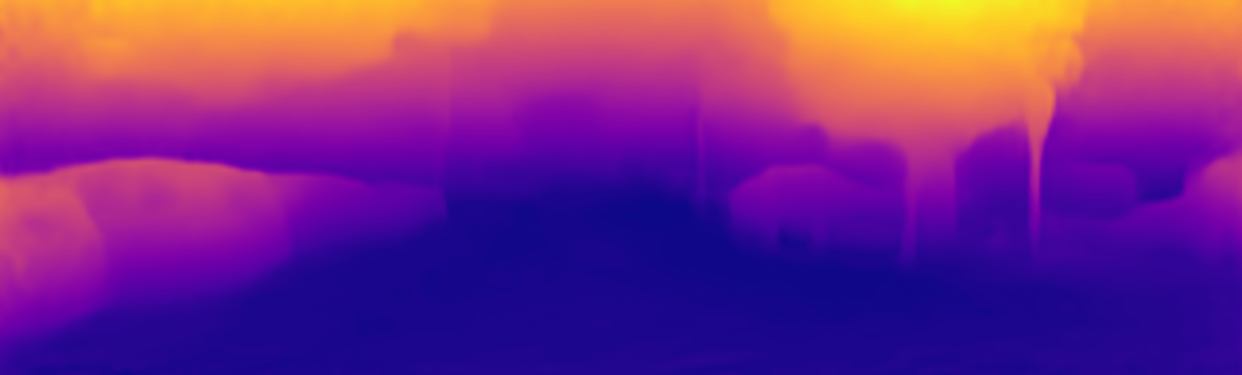} &
        \includegraphics[width=\linewidth]{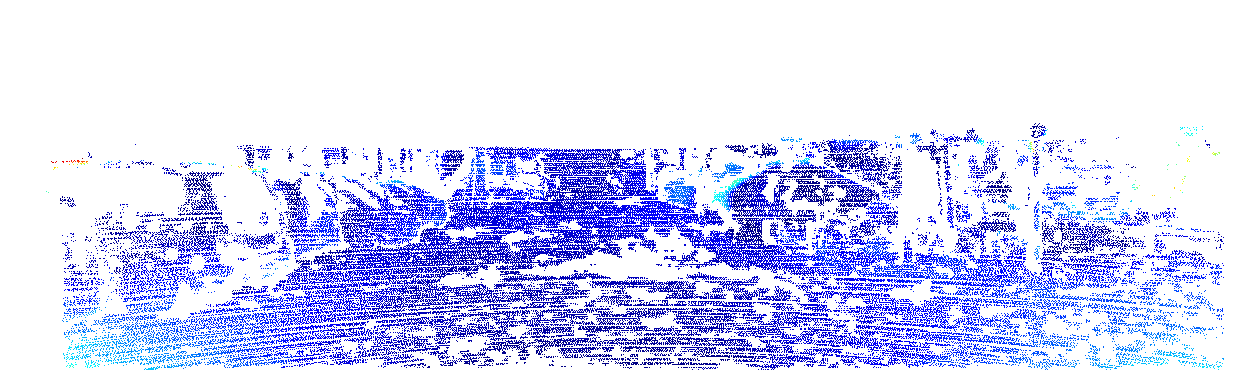} &
        \includegraphics[width=\linewidth]{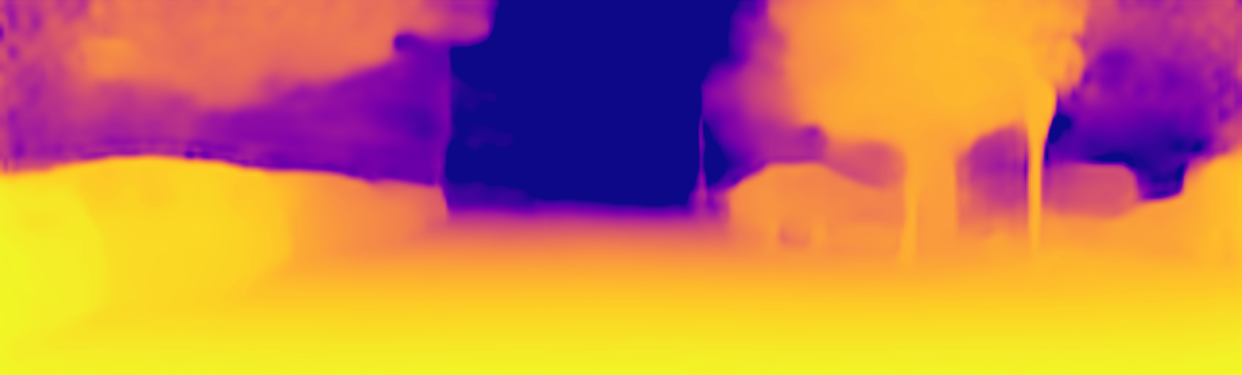} &
        \includegraphics[width=\linewidth]{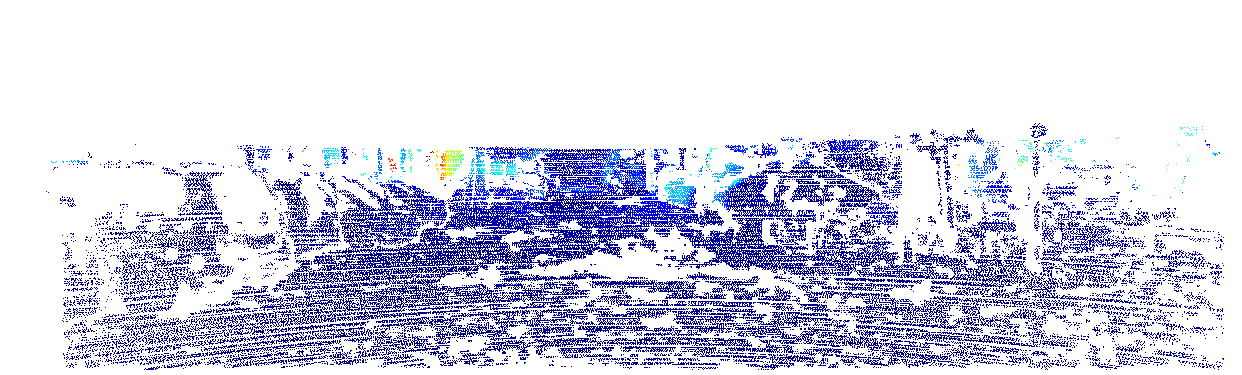} \\

        \includegraphics[width=\linewidth]{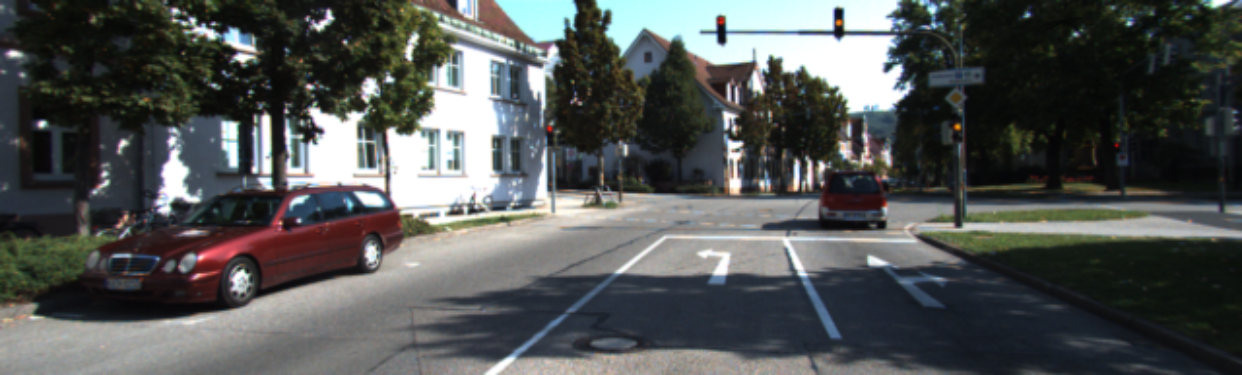} &
        \includegraphics[width=\linewidth]{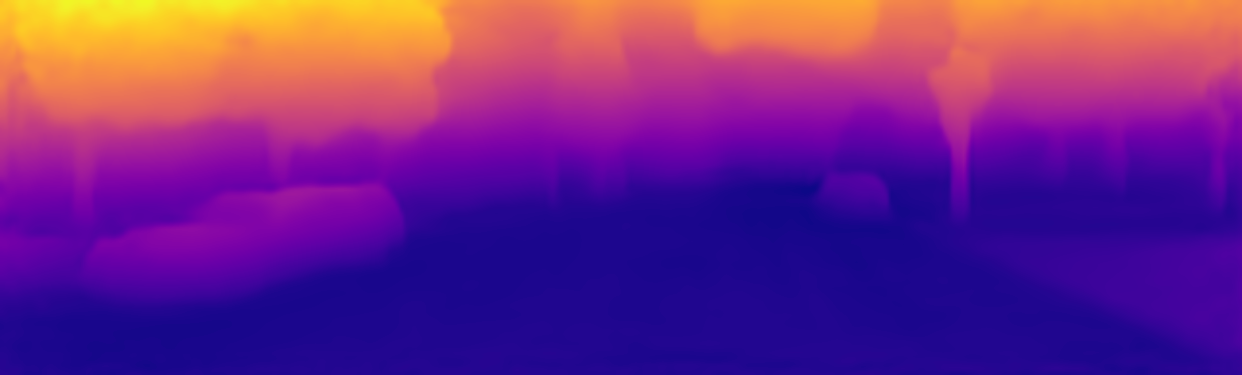} &
        \includegraphics[width=\linewidth]{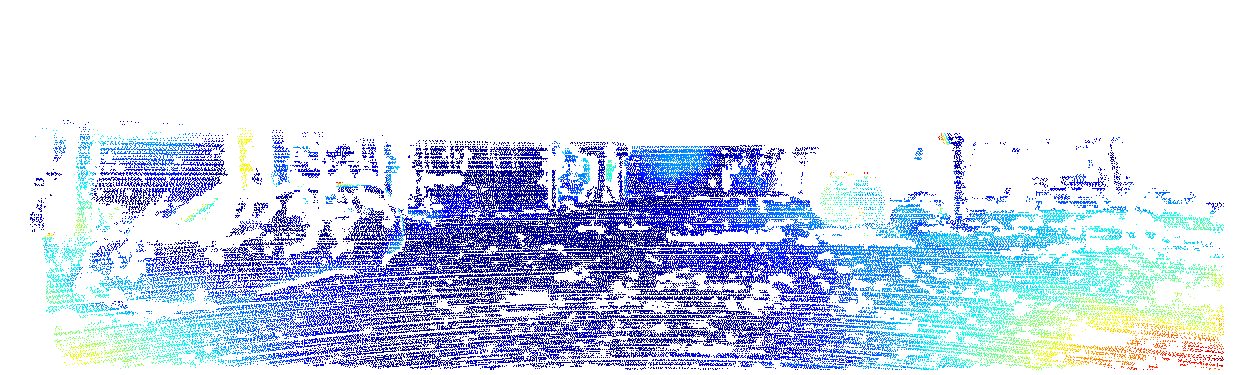} &
        \includegraphics[width=\linewidth]{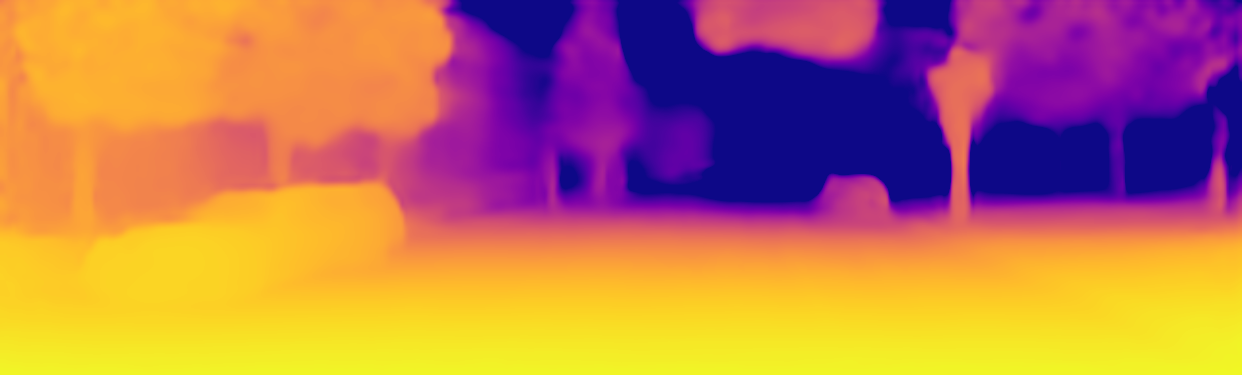} &
        \includegraphics[width=\linewidth]{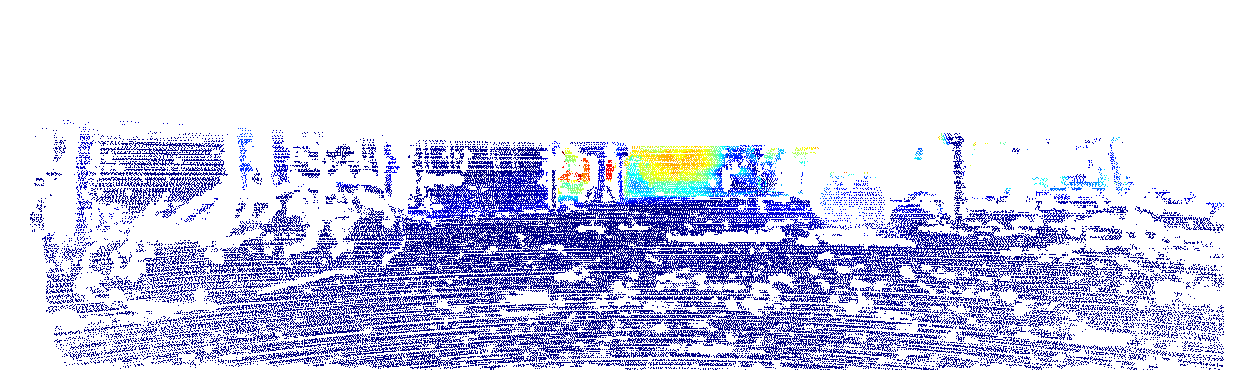} \\

        \includegraphics[width=\linewidth]{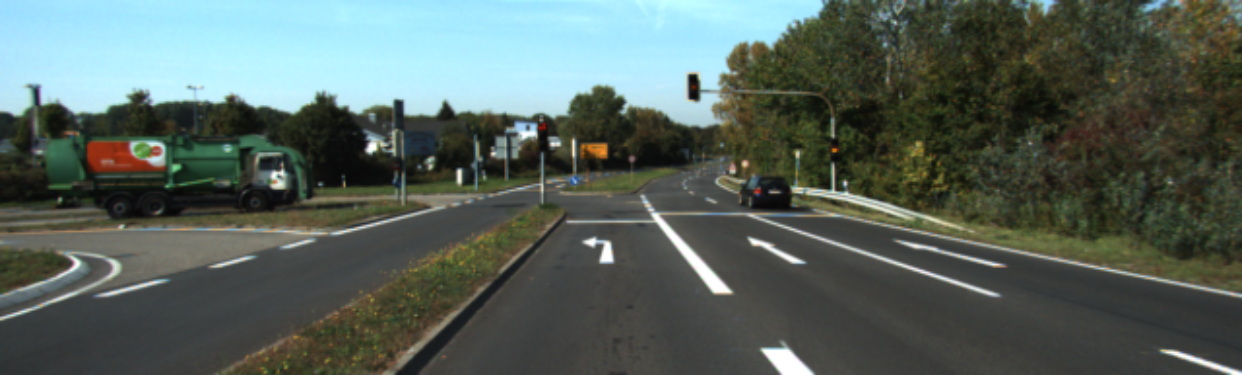} &
        \includegraphics[width=\linewidth]{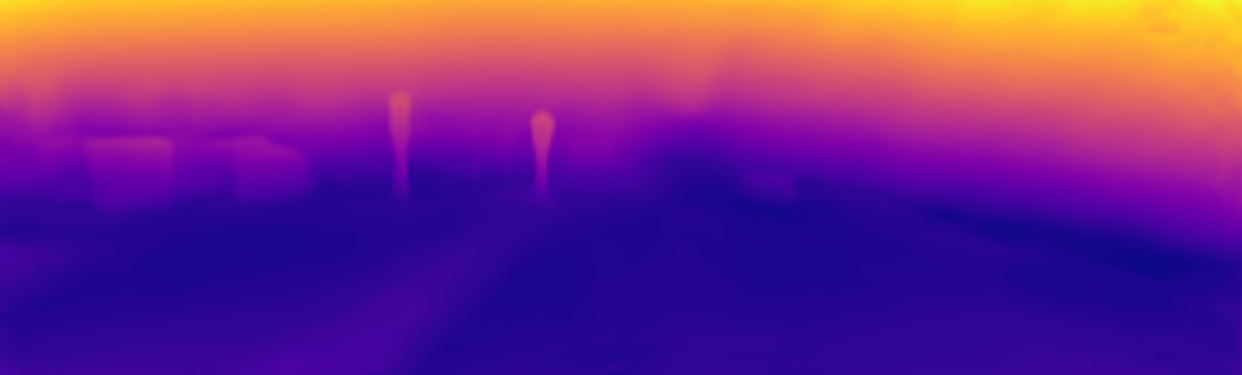} &
        \includegraphics[width=\linewidth]{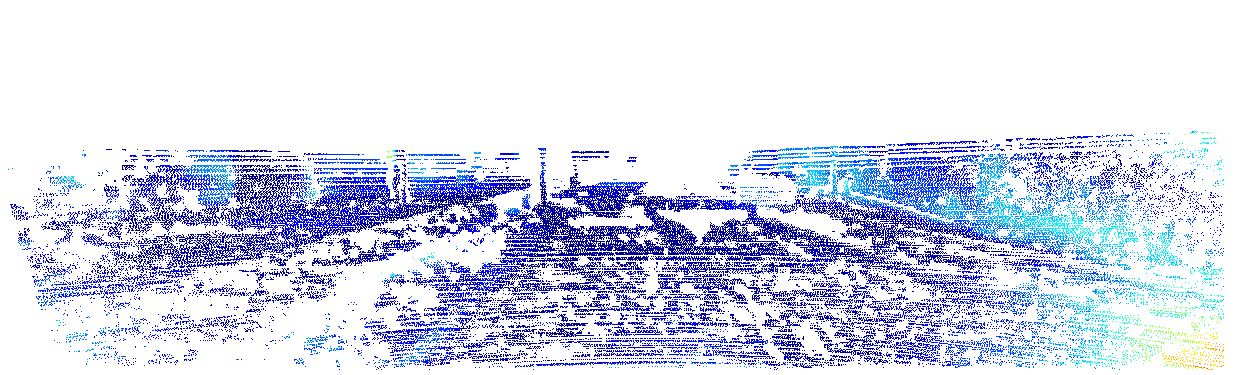} &
        \includegraphics[width=\linewidth]{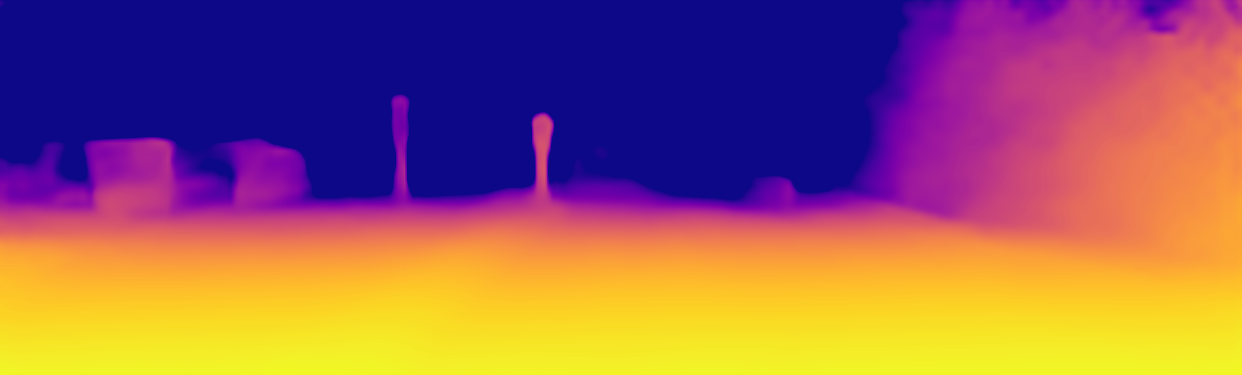} &
        \includegraphics[width=\linewidth]{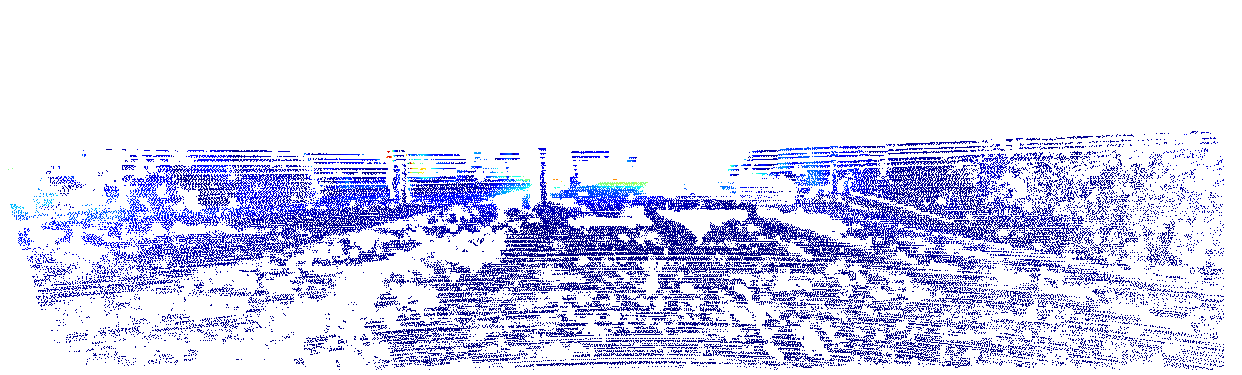} \\

        \includegraphics[width=\linewidth]{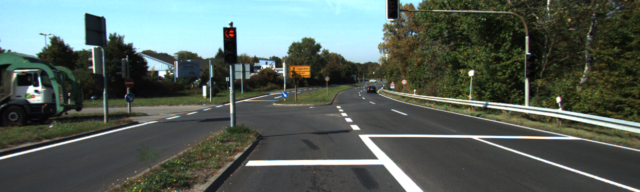} &
        \includegraphics[width=\linewidth]{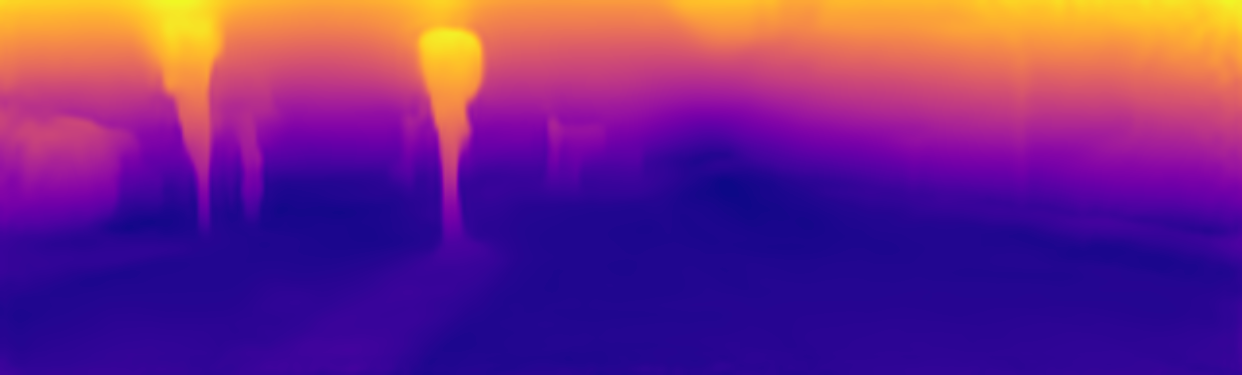} &
        \includegraphics[width=\linewidth]{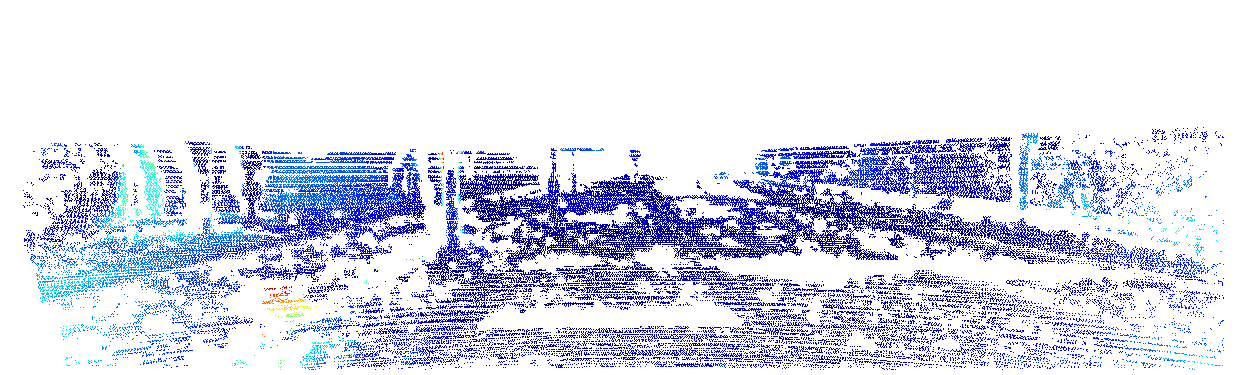} &
        \includegraphics[width=\linewidth]{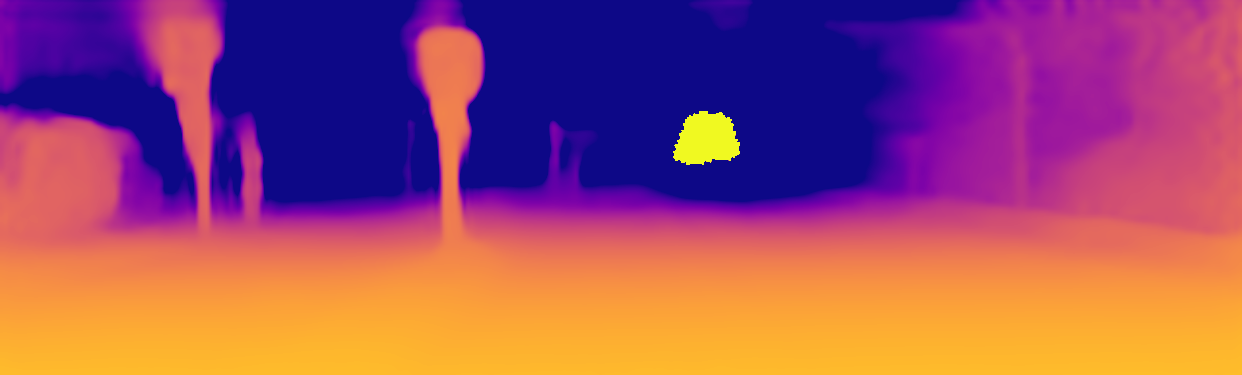} &
        \includegraphics[width=\linewidth]{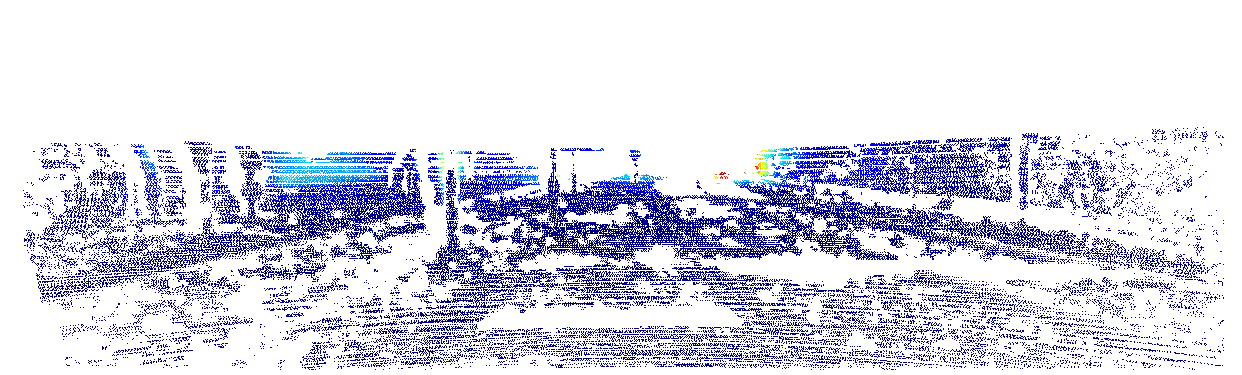} \\

        \includegraphics[width=\linewidth]{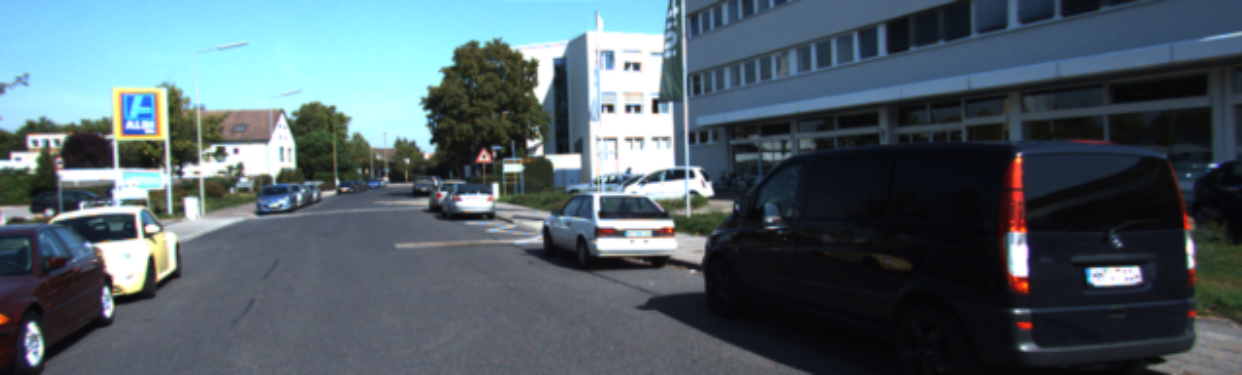} &
        \includegraphics[width=\linewidth]{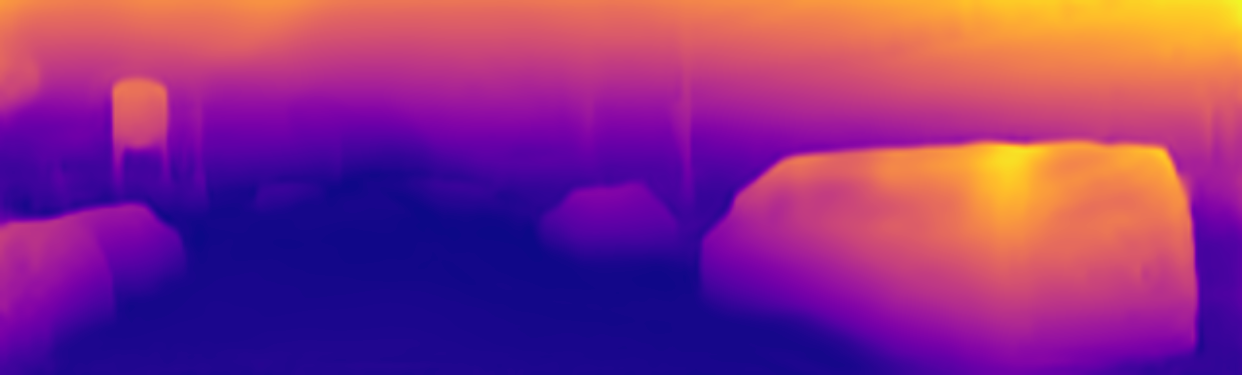} &
        \includegraphics[width=\linewidth]{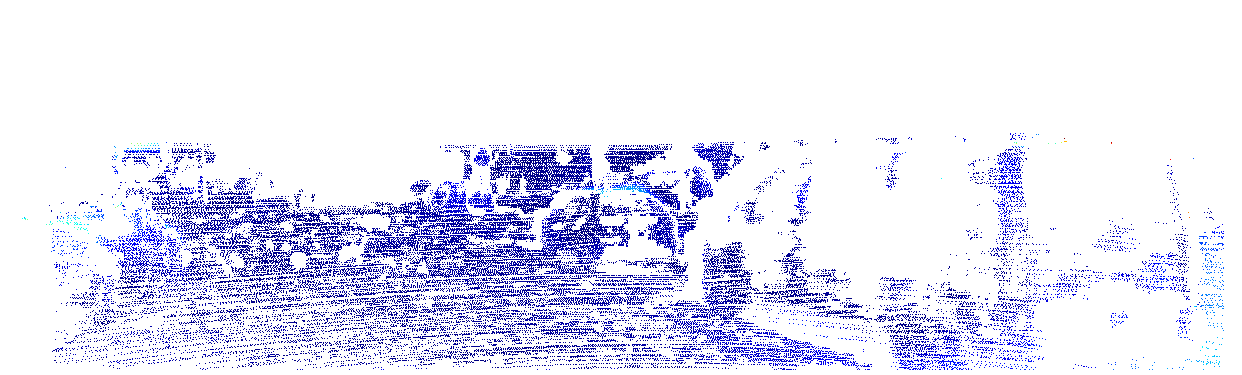} &
        \includegraphics[width=\linewidth]{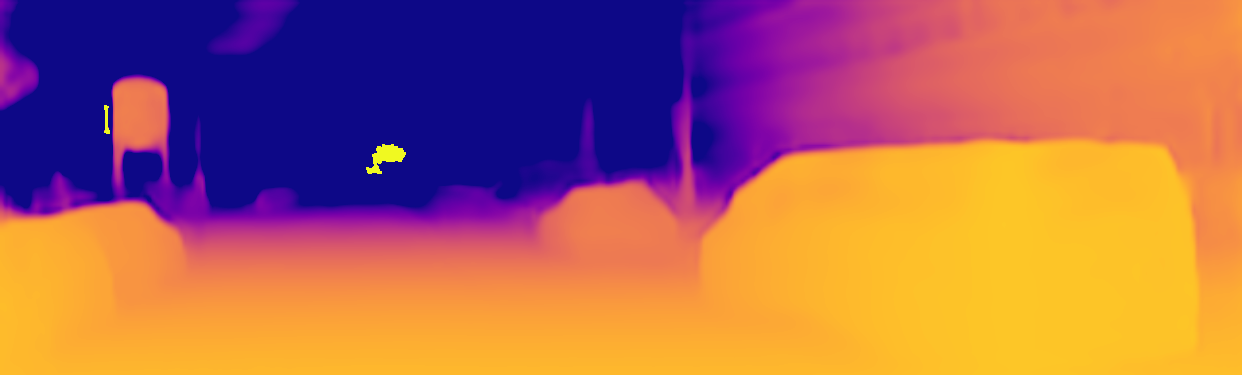} &
        \includegraphics[width=\linewidth]{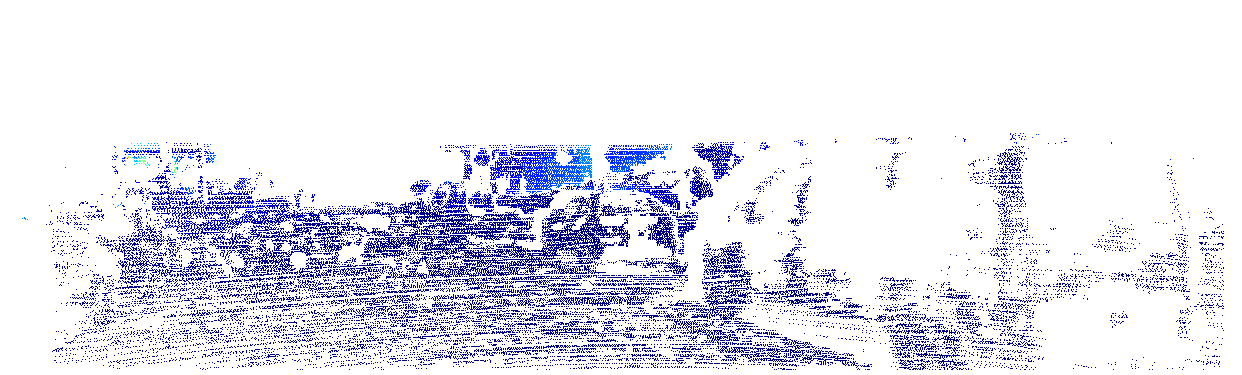} \\

        \includegraphics[width=\linewidth]{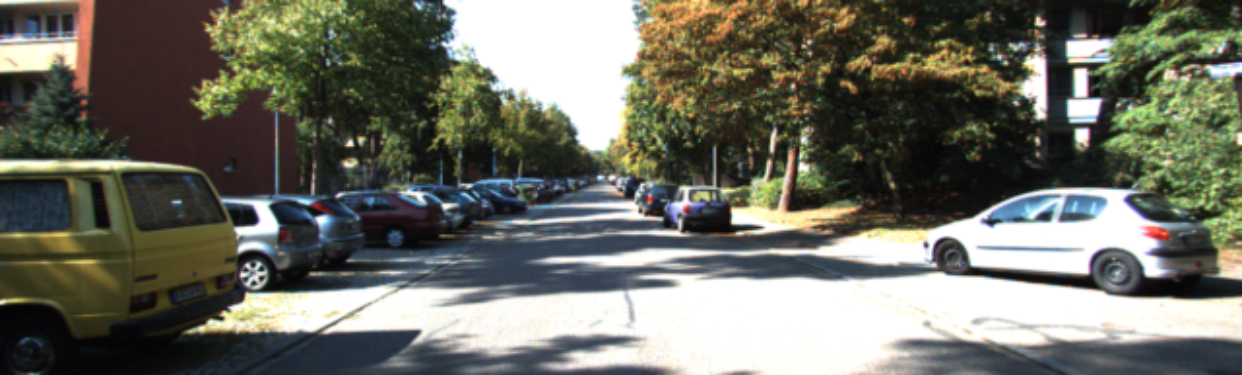} &
        \includegraphics[width=\linewidth]{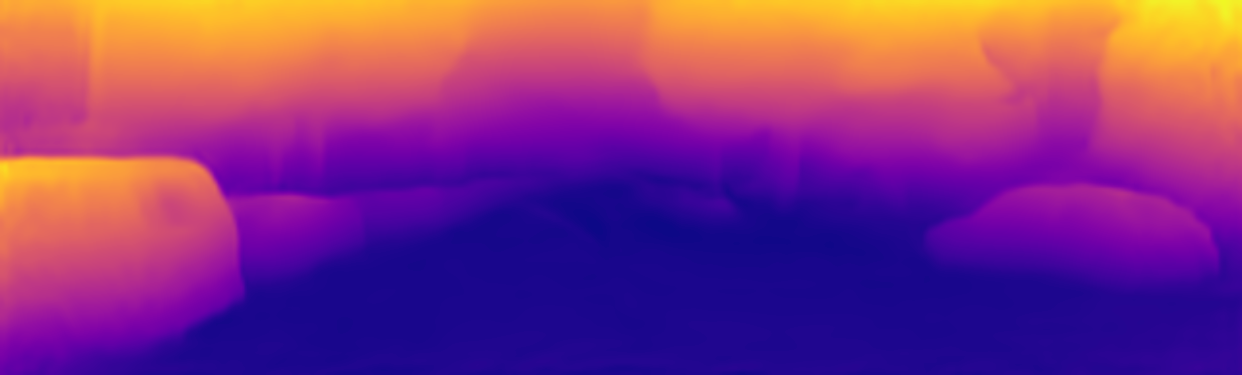} &
        \includegraphics[width=\linewidth]{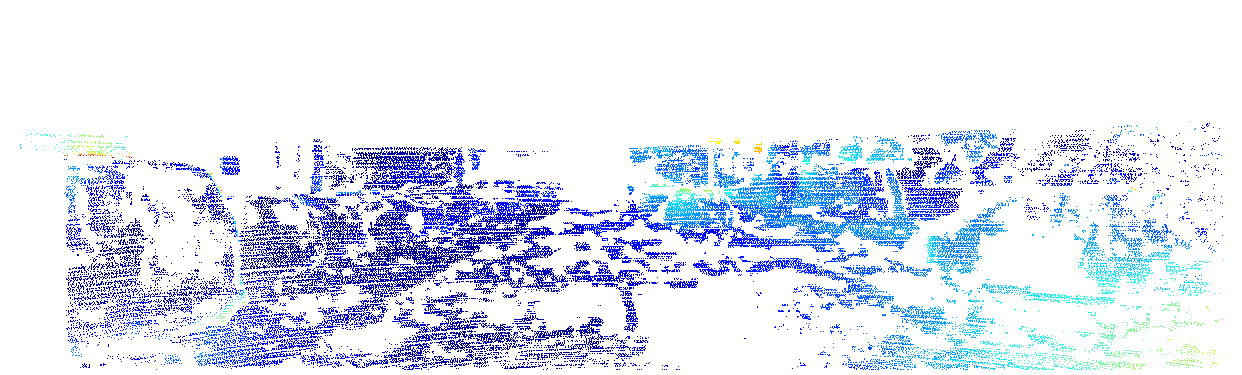} &
        \includegraphics[width=\linewidth]{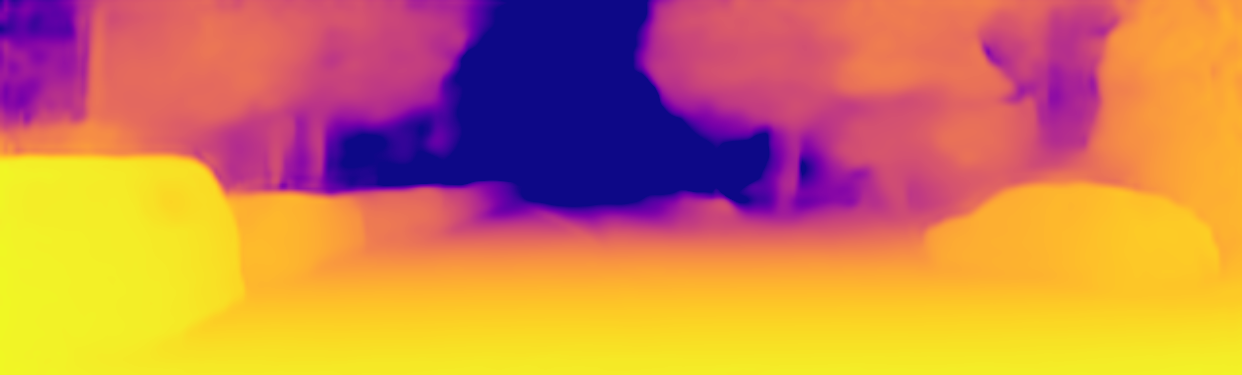} &
        \includegraphics[width=\linewidth]{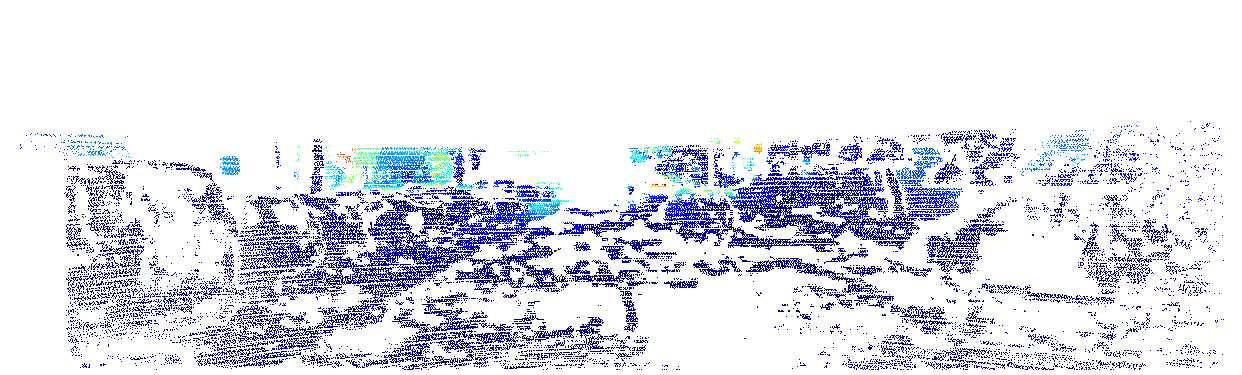} \\

        \includegraphics[width=\linewidth]{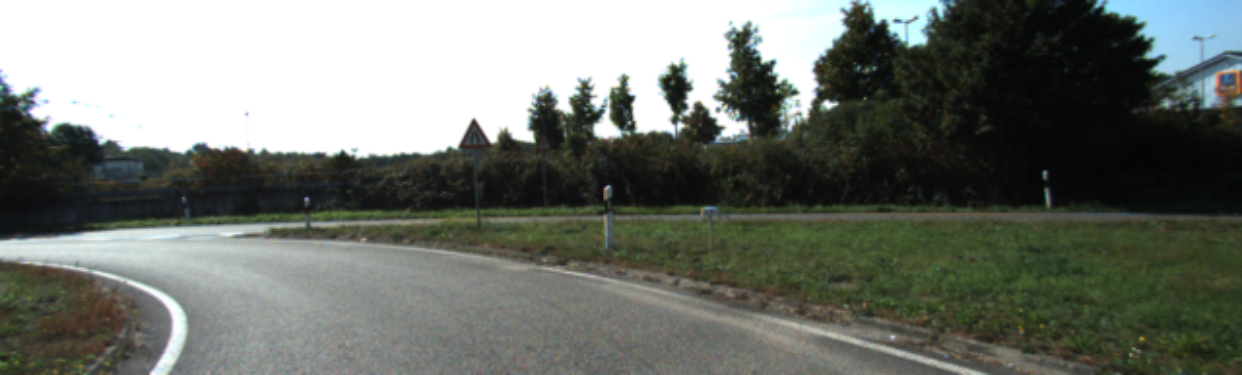} &
        \includegraphics[width=\linewidth]{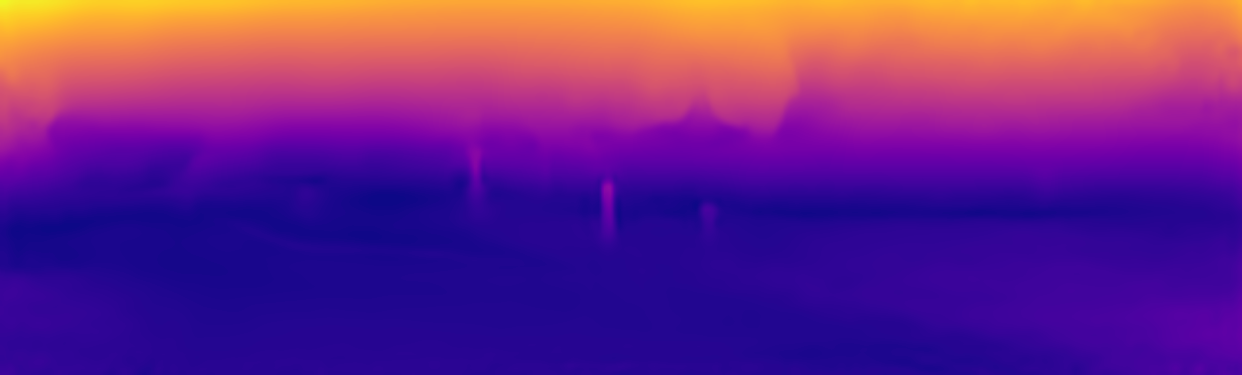} &
        \includegraphics[width=\linewidth]{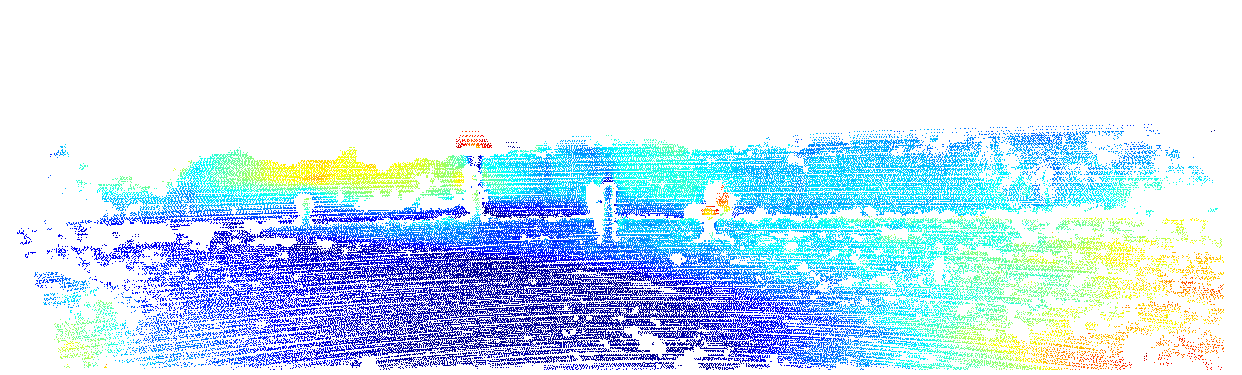} &
        \includegraphics[width=\linewidth]{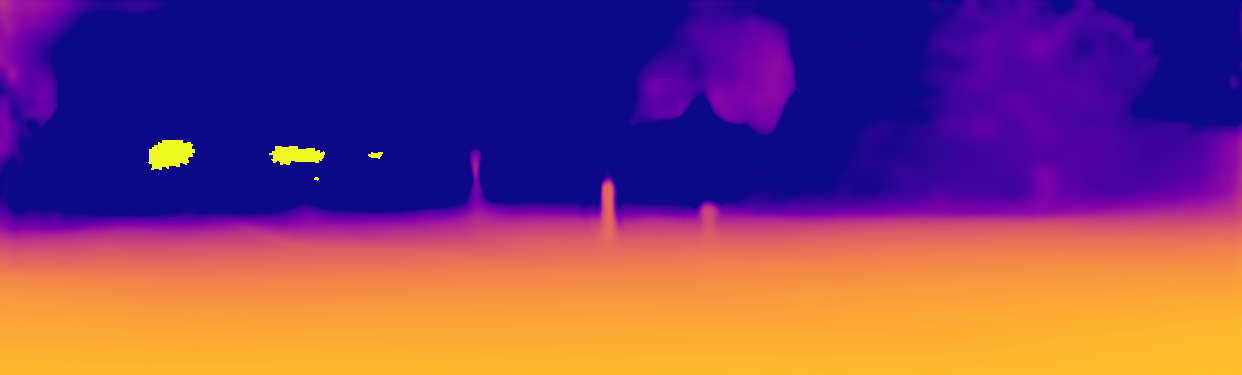} &
        \includegraphics[width=\linewidth]{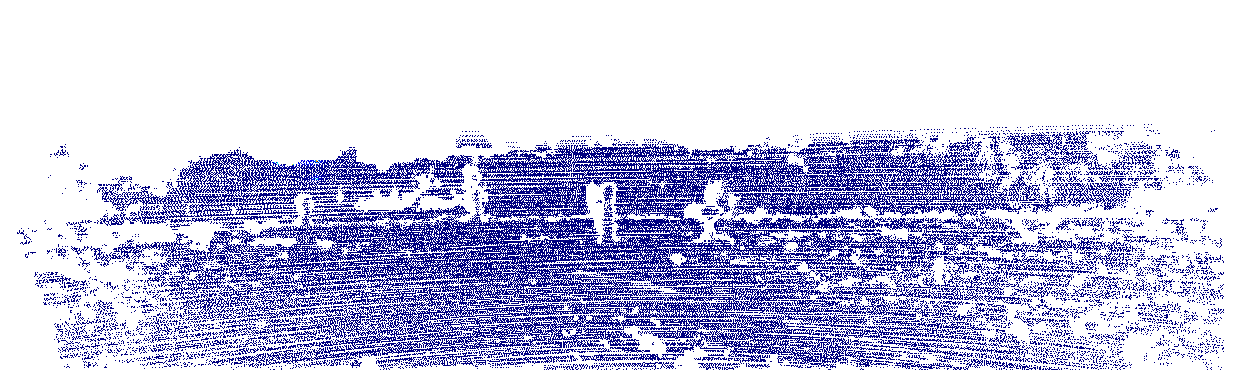} \\

        \includegraphics[width=\linewidth]{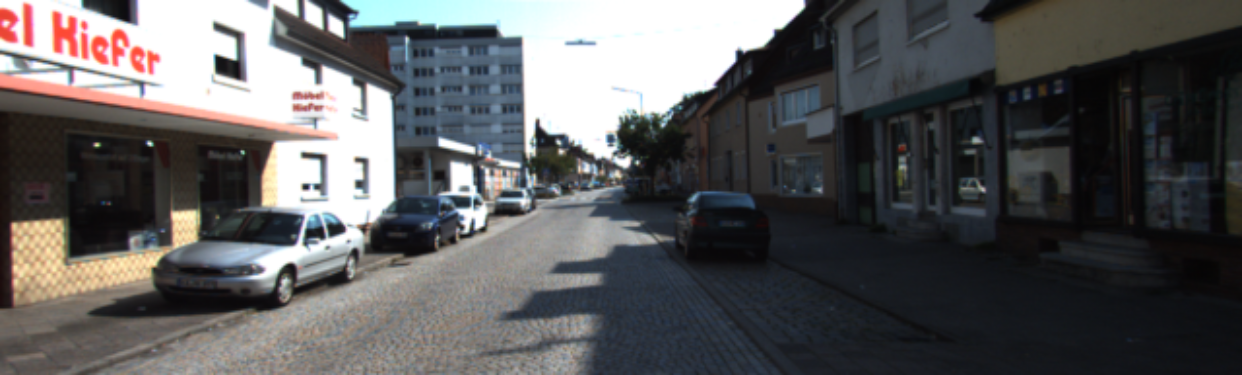} &
        \includegraphics[width=\linewidth]{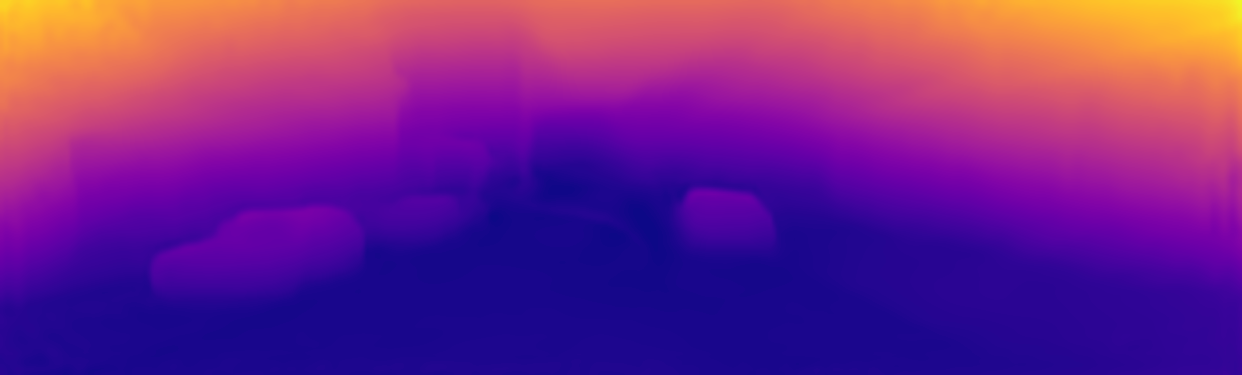} &
        \includegraphics[width=\linewidth]{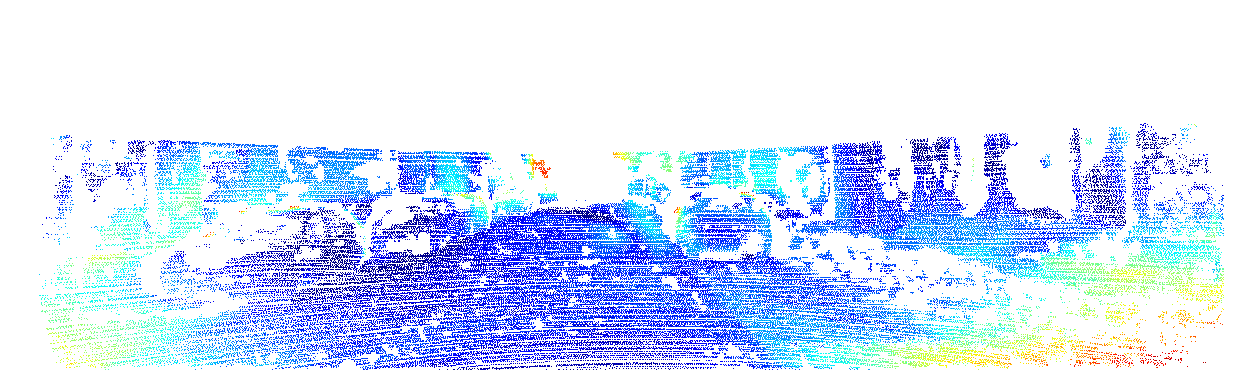} &
        \includegraphics[width=\linewidth]{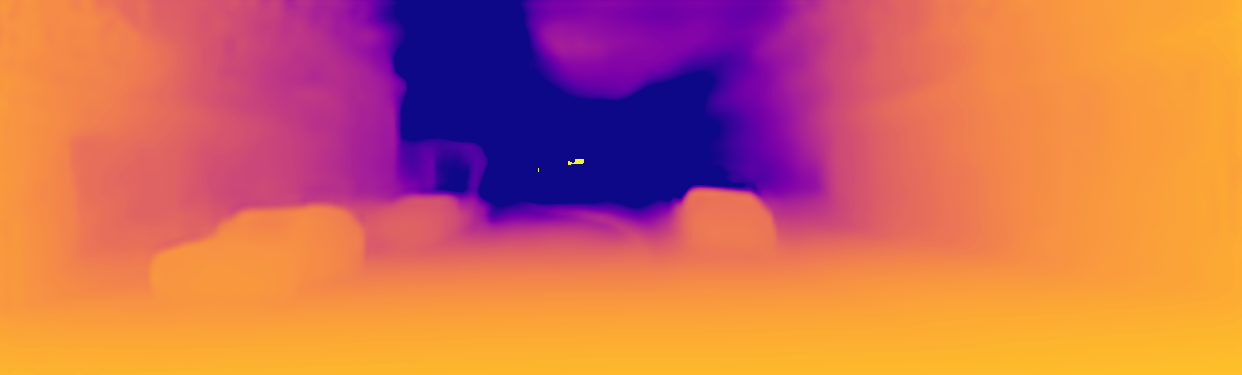} &
        \includegraphics[width=\linewidth]{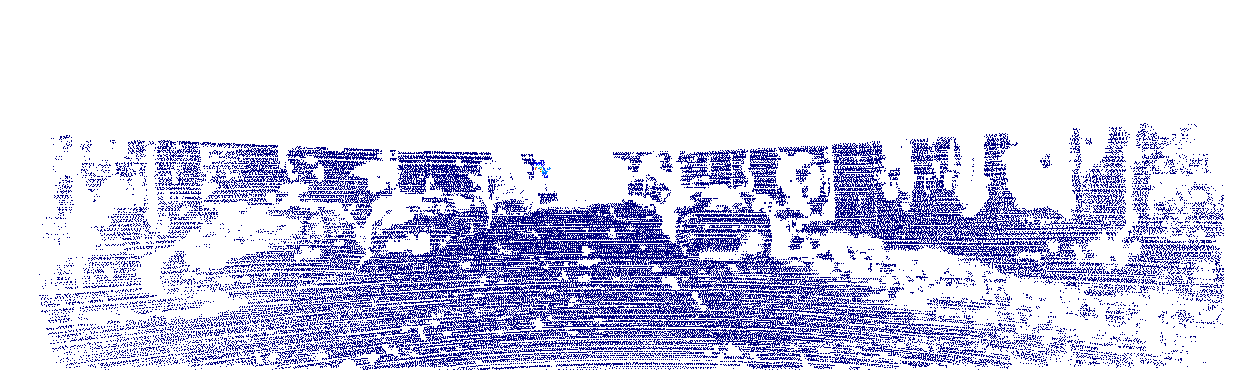} \\
        \bottomrule
    \end{tabular}
    \caption{Qualitative comparison of estimated $\gamma$ values and depth outputs on the KITTI test set benchmark. The first column shows input images, followed by their respective $\gamma$ outputs, $\gamma$ error maps, depth predictions, and depth error maps. Yellow artifacts in the depth maps result from instability around the epipole.}

    \label{tab:qualitative_results_appendix_good}
\end{table*}

\begin{table*}[t]
    \centering
    \small
    \renewcommand{\arraystretch}{1.2}
    \setlength{\tabcolsep}{4pt}
    \begin{tabular}{m{0.20\textwidth} m{0.17\textwidth} m{0.17\textwidth} m{0.17\textwidth} m{0.17\textwidth}}
        \toprule
        \textbf{Input Image} & \textbf{Gamma Output} & \textbf{Gamma Error} & \textbf{Depth Output} & \textbf{Depth Error} \\
        \midrule

        \includegraphics[width=\linewidth]{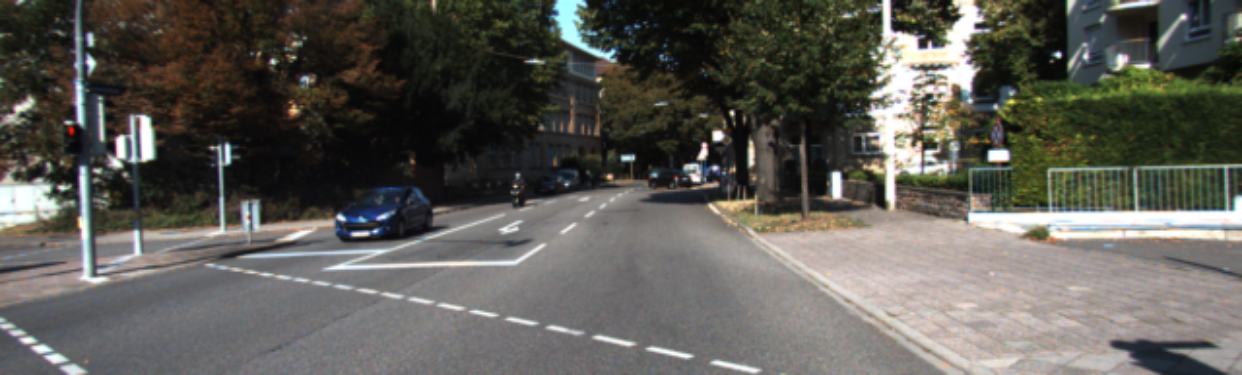} &
        \includegraphics[width=\linewidth]{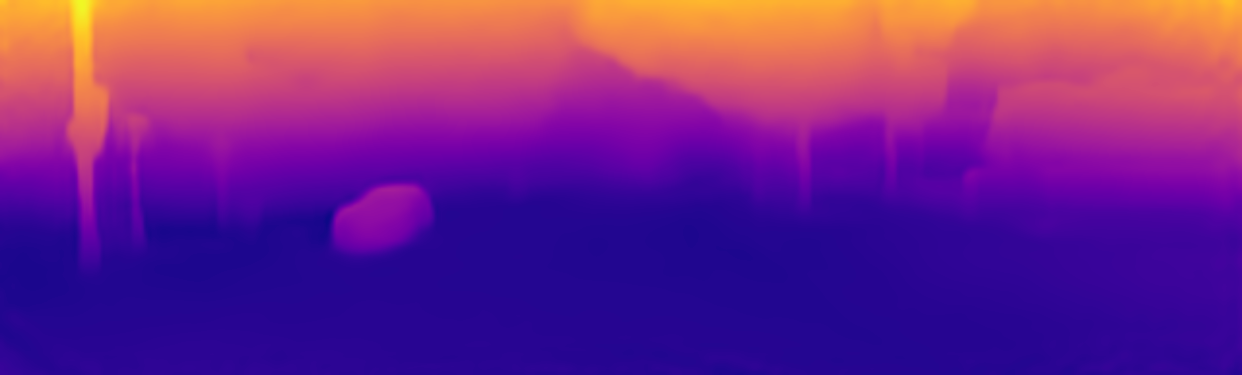} &
        \includegraphics[width=\linewidth]{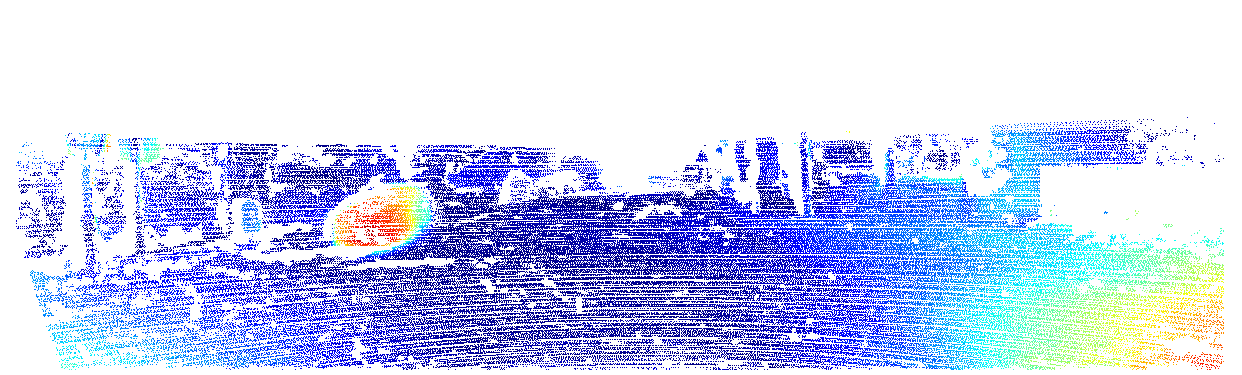} &
        \includegraphics[width=\linewidth]{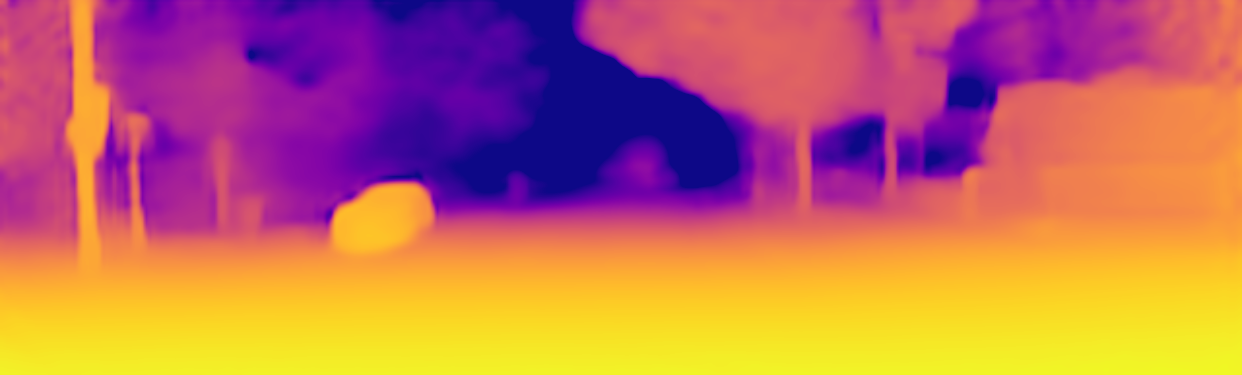} &
        \includegraphics[width=\linewidth]{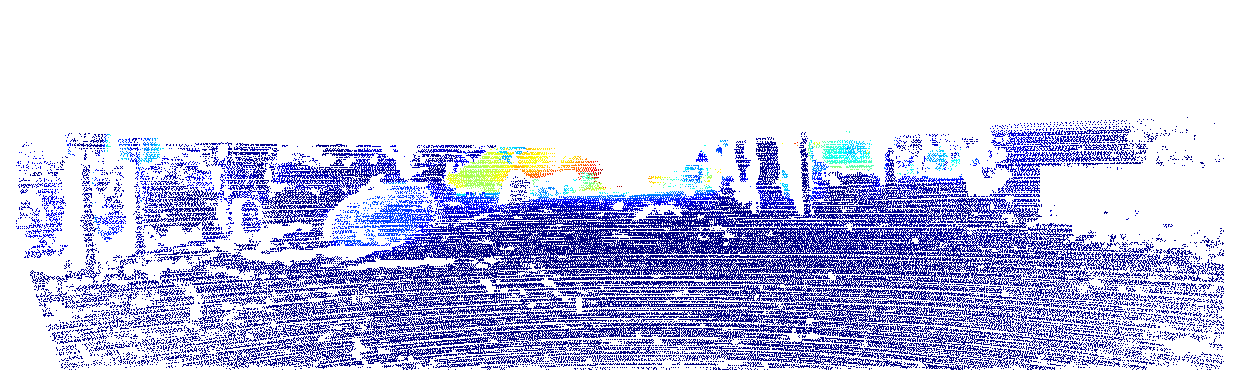} \\

        \includegraphics[width=\linewidth]{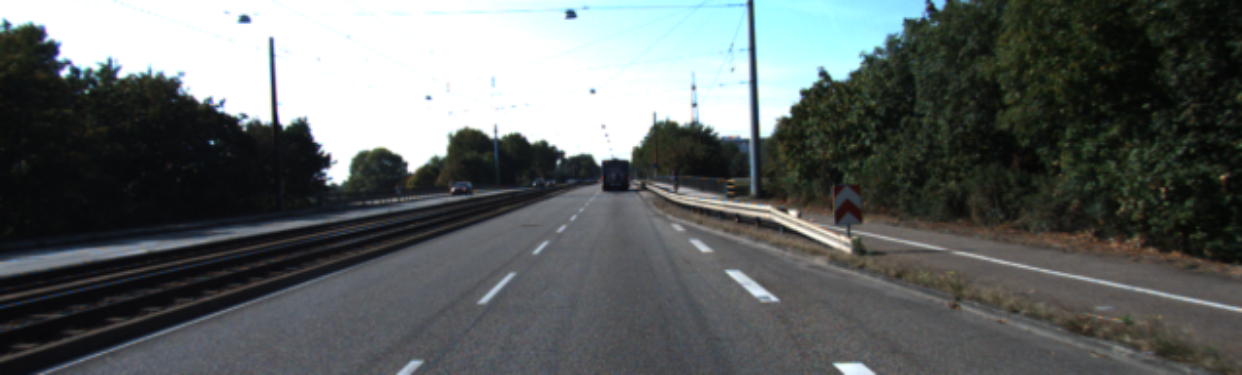} &
        \includegraphics[width=\linewidth]{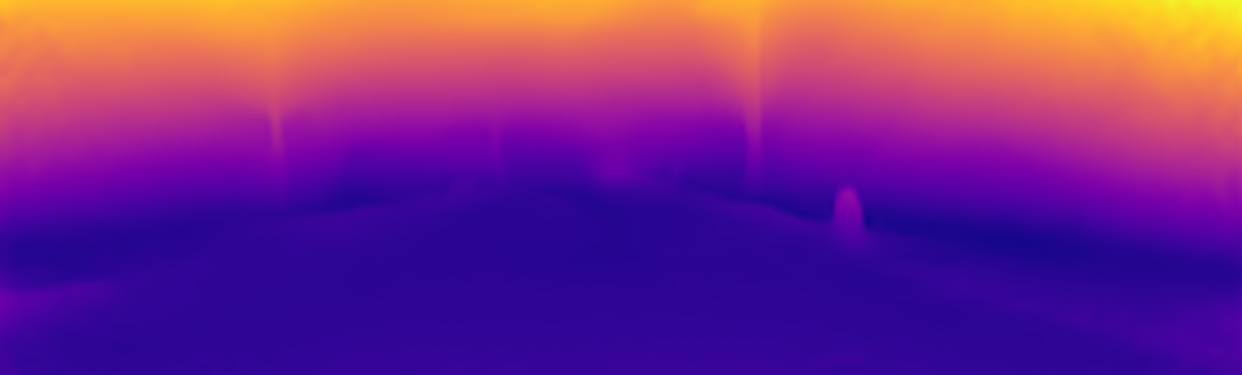} &
        \includegraphics[width=\linewidth]{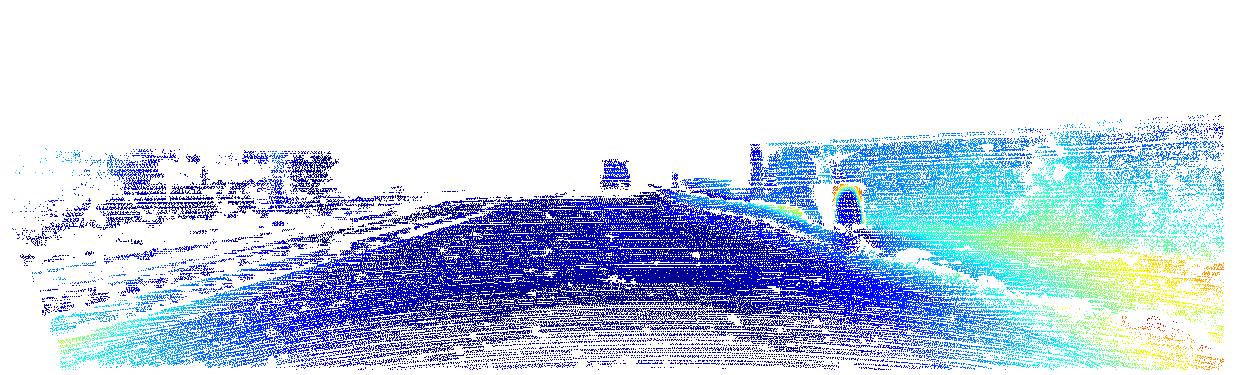} &
        \includegraphics[width=\linewidth]{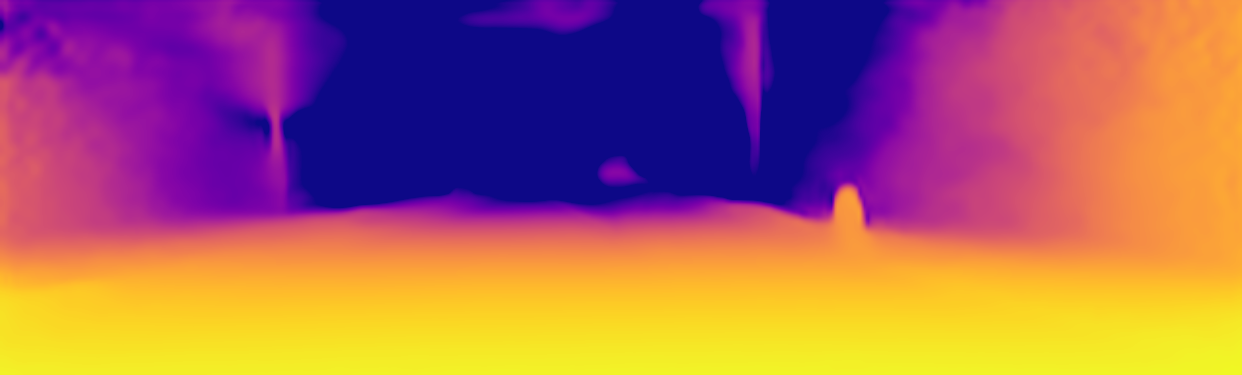} &
        \includegraphics[width=\linewidth]{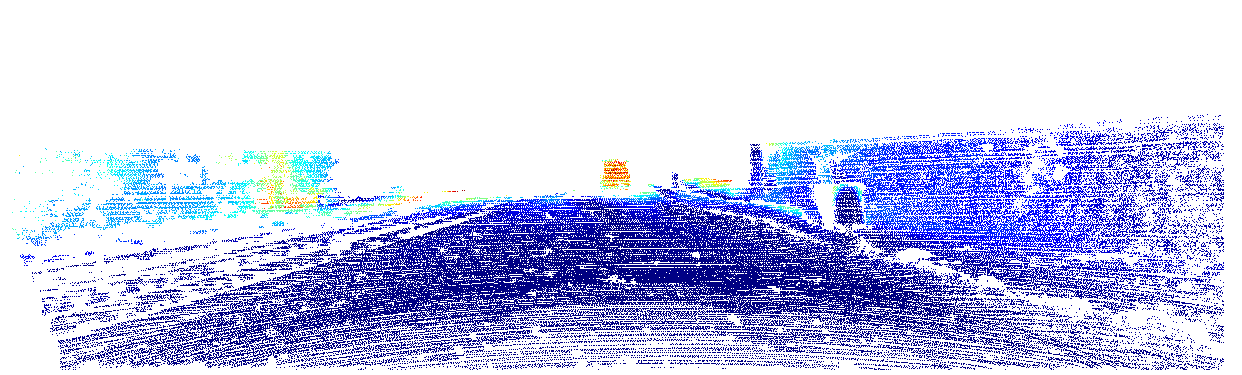} \\

        \includegraphics[width=\linewidth]{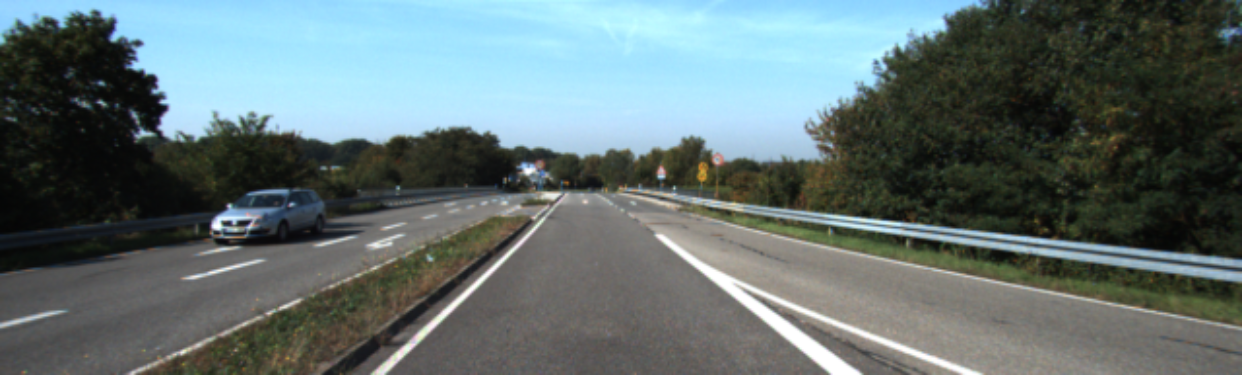} &
        \includegraphics[width=\linewidth]{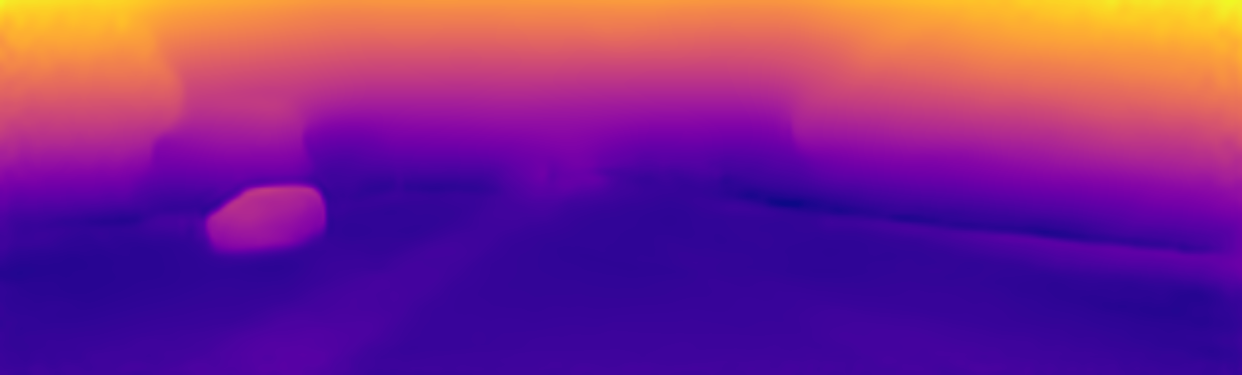} &
        \includegraphics[width=\linewidth]{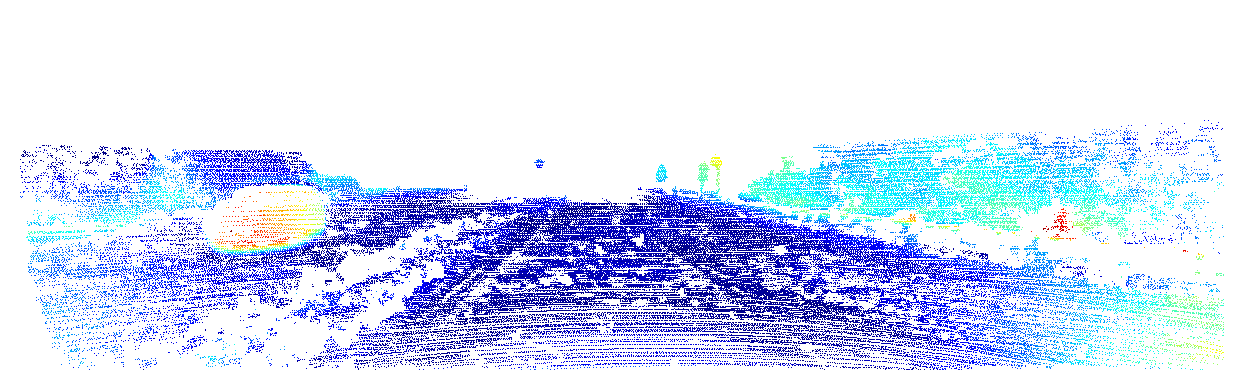} &
        \includegraphics[width=\linewidth]{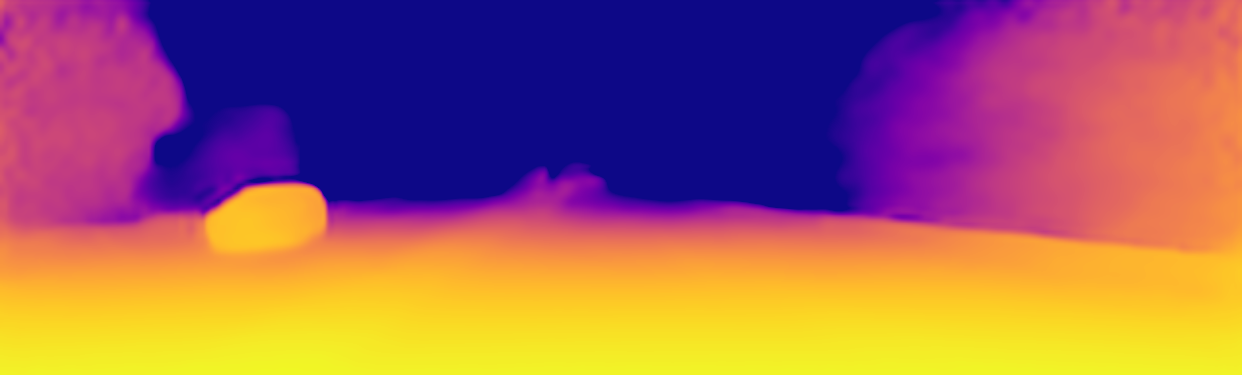} &
        \includegraphics[width=\linewidth]{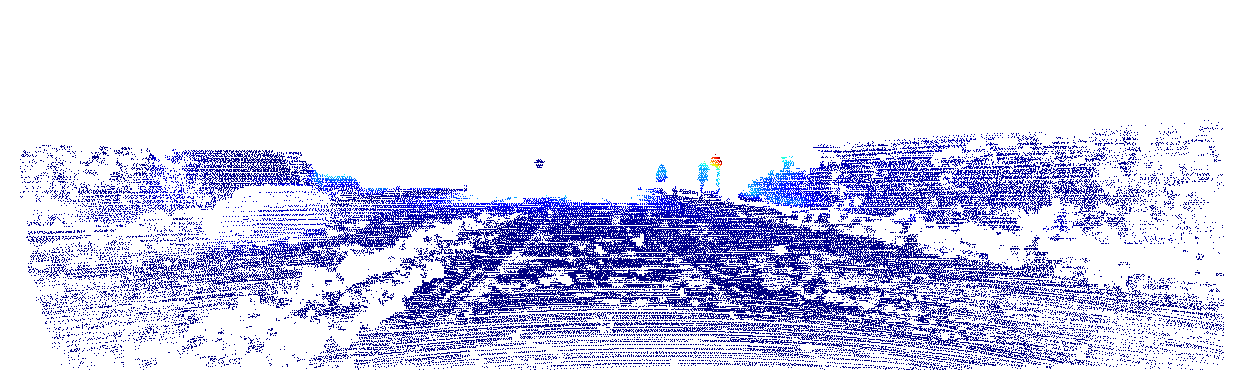} \\

        \includegraphics[width=\linewidth]{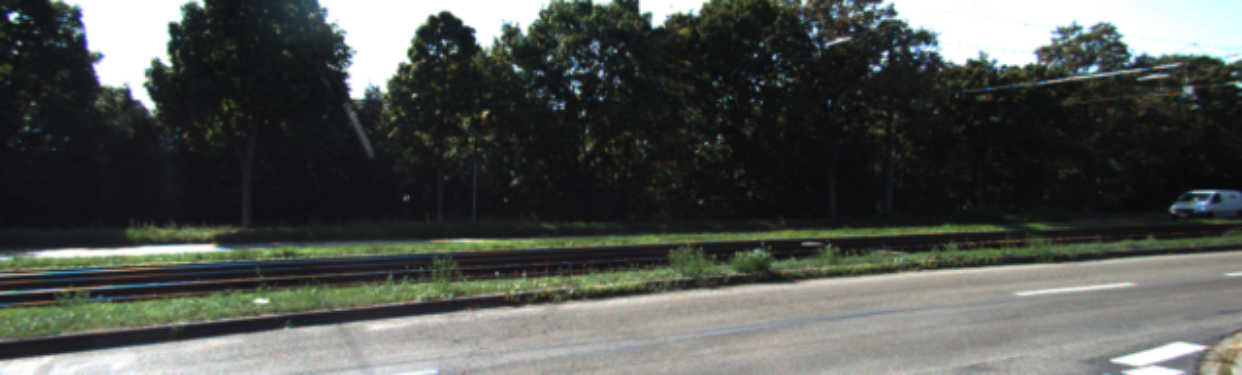} &
        \includegraphics[width=\linewidth]{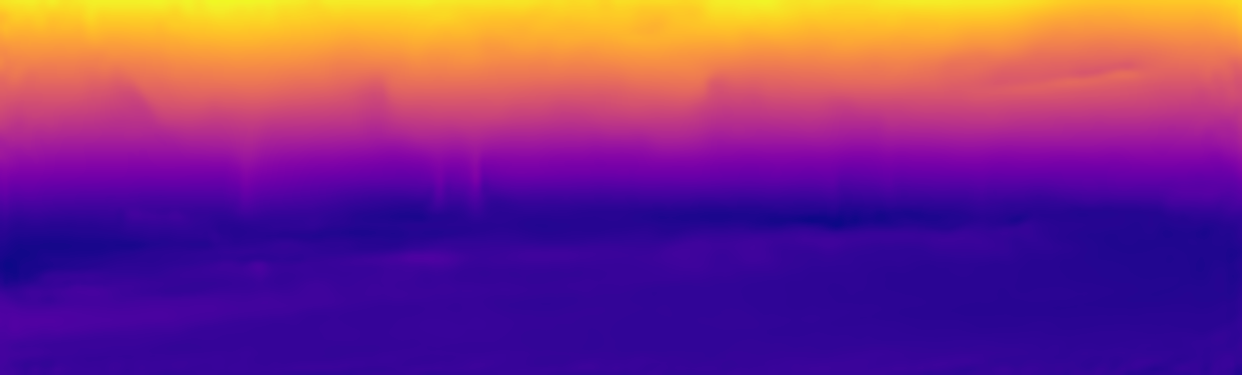} &
        \includegraphics[width=\linewidth]{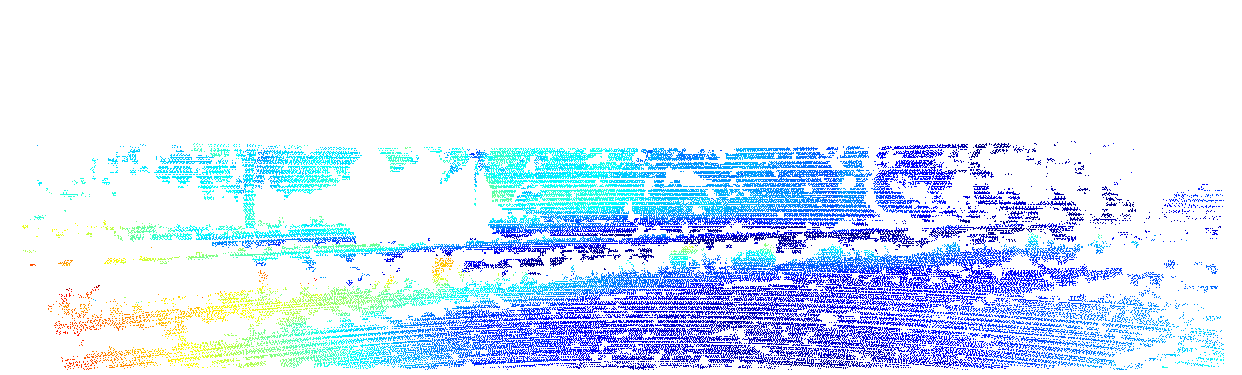} &
        \includegraphics[width=\linewidth]{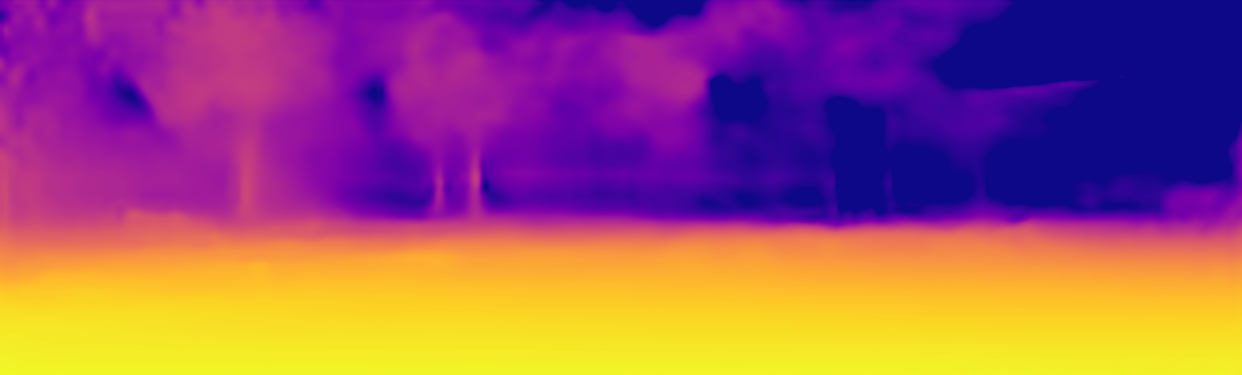} &
        \includegraphics[width=\linewidth]{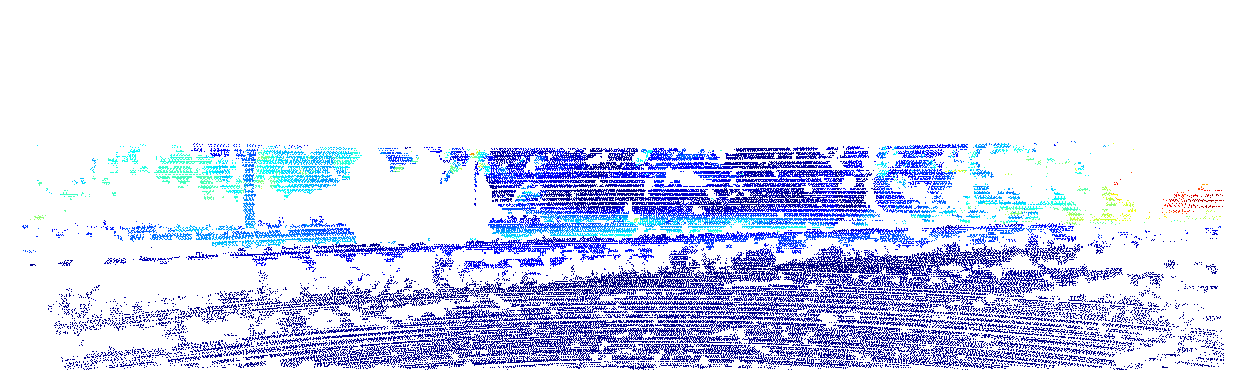} \\
        
        \bottomrule
    \end{tabular}
    \caption{Qualitative examples of some failure examples for KITTI, as seen most of our failure cases are faraway dynamic objects, and that is why our model exhibited better than the baselines in the \textit{Abs. Rel.} metric for depth evaluation. The first column contains input images, followed by their respective gamma outputs, gamma error, depth outputs, and depth error maps.}
    \label{tab:qualitative_results_appendix_bad}
\end{table*}

\begin{figure*}[t]
    \centering
    %
    %
    \begin{minipage}[c]{0.23\textwidth}  
        \centering
        \vspace{4em}
        \includegraphics[width=1\linewidth,height=1.5cm]{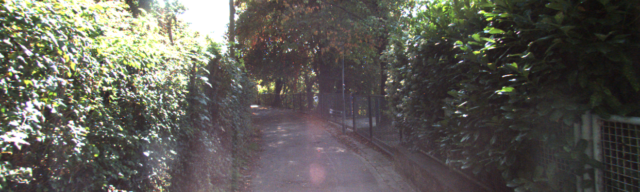}\\
        \vspace{0.3em}
        \small (a)~Input (Example 1)

        \vspace{7.2em}

        \includegraphics[width=\linewidth,height=1.5cm]{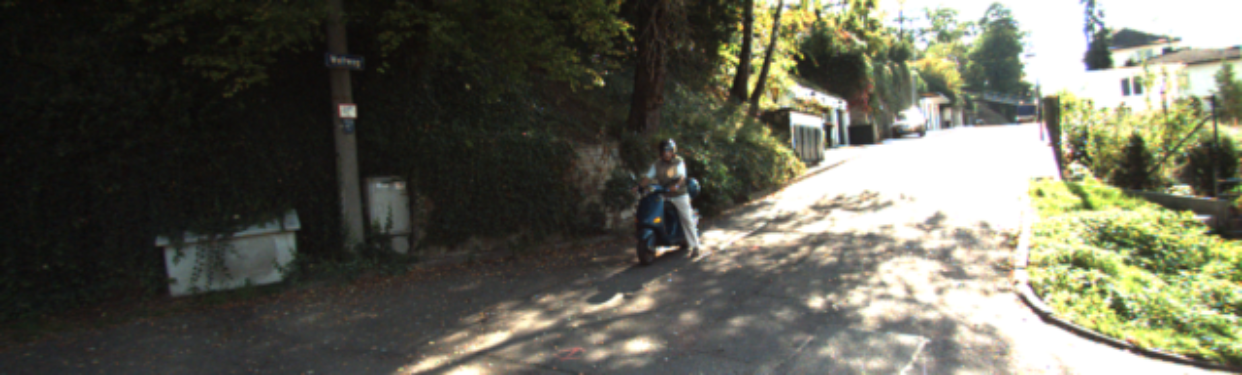}\\
        \vspace{0.3em}
        \small (b)~Input (Example 2)
    \end{minipage}
    %
    %
    \begin{minipage}[c]{0.75\textwidth}  
        \centering
        \small
        \setlength{\tabcolsep}{2pt}   
        \begin{tabular}{
            >{\centering\arraybackslash}m{0.12\textwidth}  
            >{\centering\arraybackslash}m{0.20\textwidth}  
            >{\centering\arraybackslash}m{0.20\textwidth}  
            >{\centering\arraybackslash}m{0.20\textwidth}  
            >{\centering\arraybackslash}m{0.03\textwidth}  
        }
            \multicolumn{5}{c}{\textbf{Qualitative Results for Two Examples (a) and (b)}} \\
            \toprule
            \textbf{Description} 
             & \textbf{GfM (ours)} 
             & \textbf{GroCo~\cite{GroCo2024}} 
             & \textbf{DepthPro~\cite{DepthPro2024}} 
             &  \\
            \midrule

            $\gamma$ &
            \includegraphics[width=\linewidth,height=1cm]{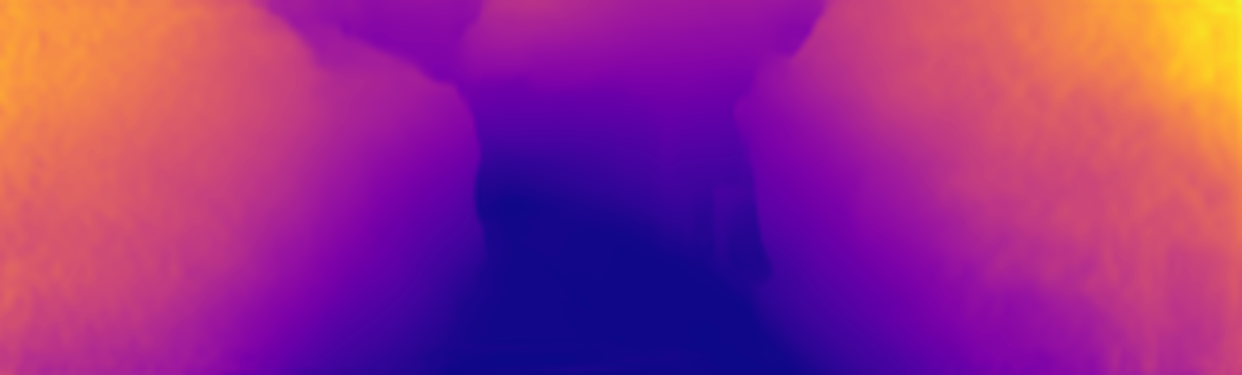} &
            \includegraphics[width=\linewidth,height=1cm]{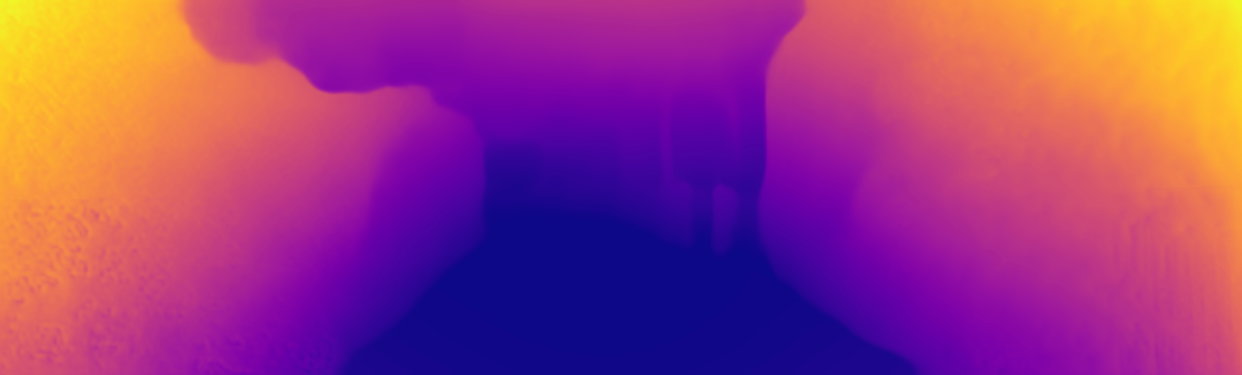} &
            \includegraphics[width=\linewidth,height=1cm]{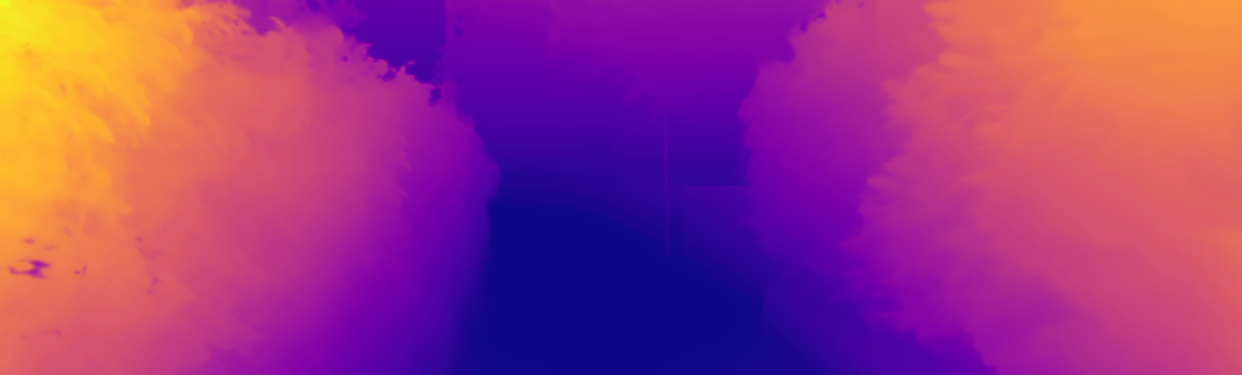} &
            \includegraphics[width=0.9\linewidth,height=1cm]{figures/colormaps/new_colormap.pdf} \\

            $\gamma$ error &
            \includegraphics[width=\linewidth,height=1cm]{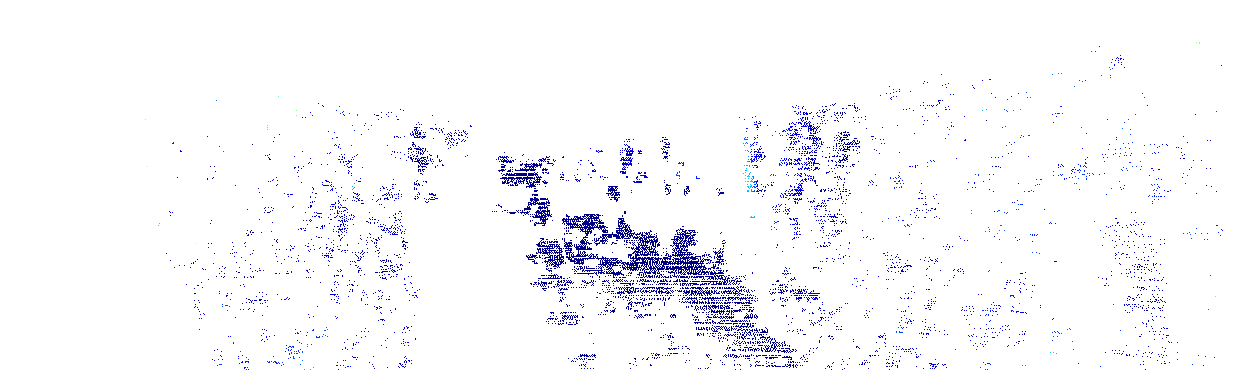} &
            \includegraphics[width=\linewidth,height=1cm]{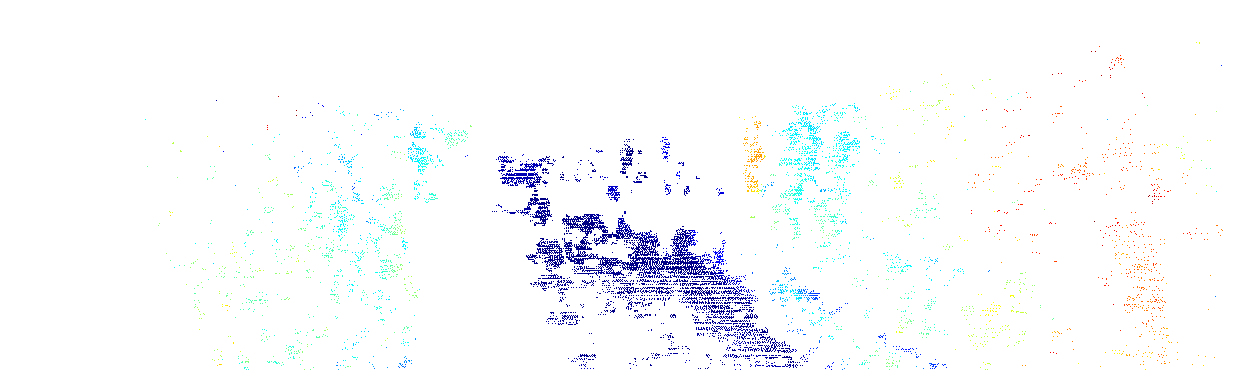} &
            \includegraphics[width=\linewidth,height=1cm]{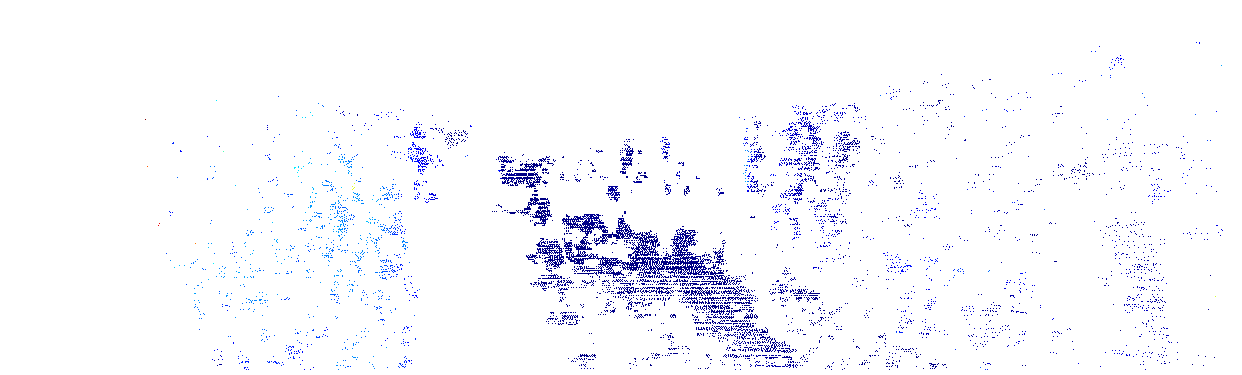} &
            \includegraphics[width=0.9\linewidth,height=1cm]{figures/colormaps/colormaps_E.pdf} \\

            Depth &
            \includegraphics[width=\linewidth,height=1cm]{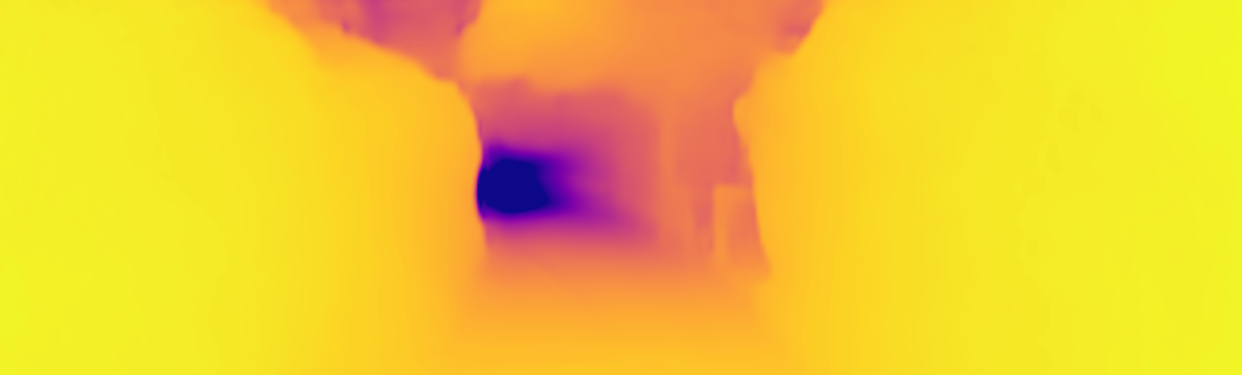} &
            \includegraphics[width=\linewidth,height=1cm]{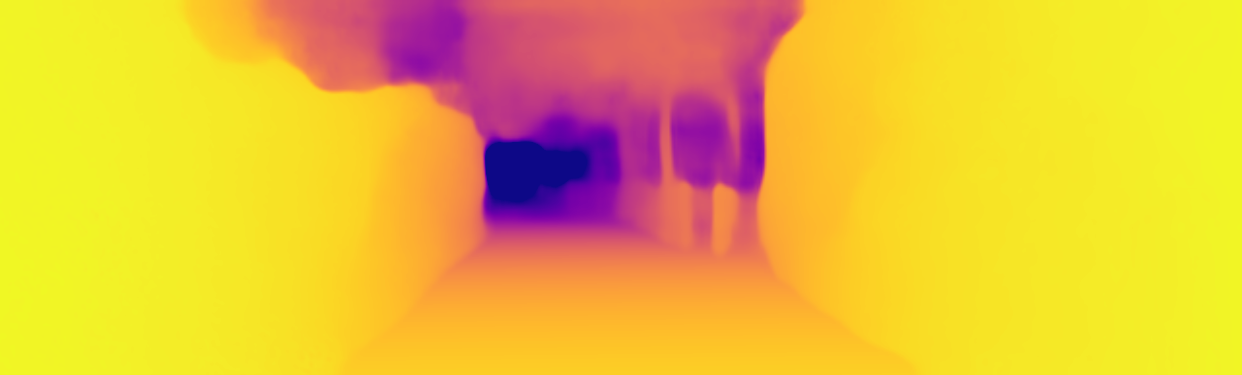} &
            \includegraphics[width=\linewidth,height=1cm]{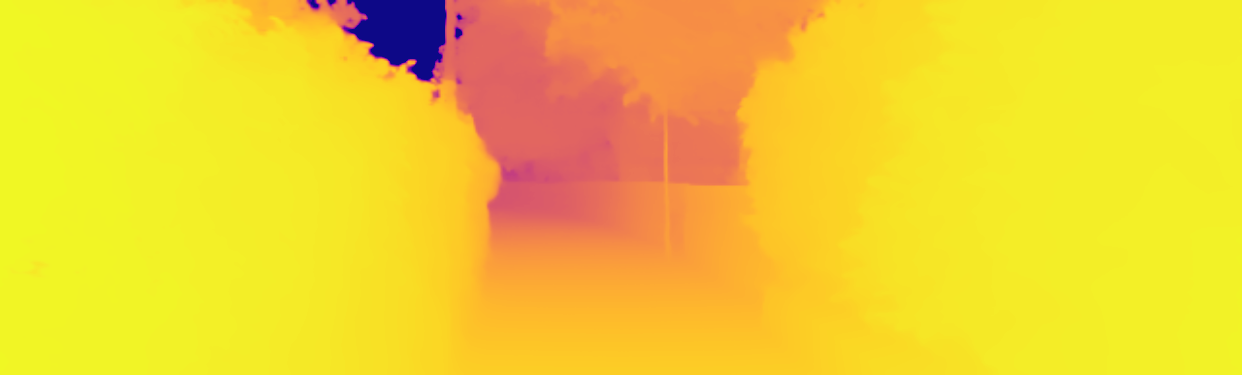} &
            \includegraphics[width=0.9\linewidth,height=1cm]{figures/colormaps/colormaps_D.pdf} \\

            Depth error &
            \includegraphics[width=\linewidth,height=1cm]{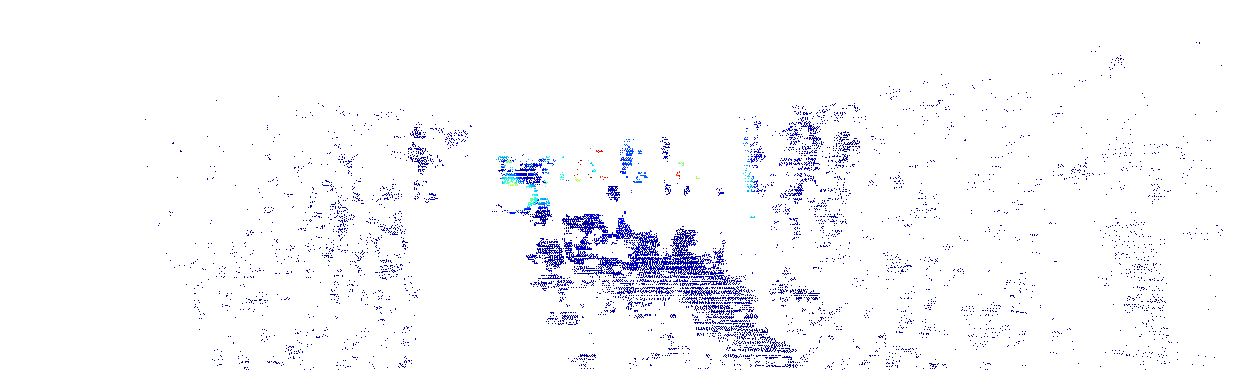} &
            \includegraphics[width=\linewidth,height=1cm]{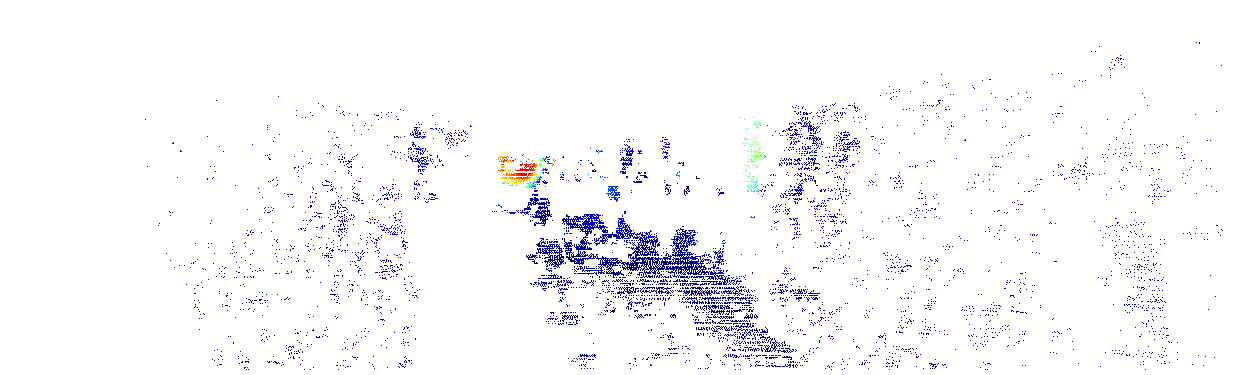} &
            \includegraphics[width=\linewidth,height=1cm]{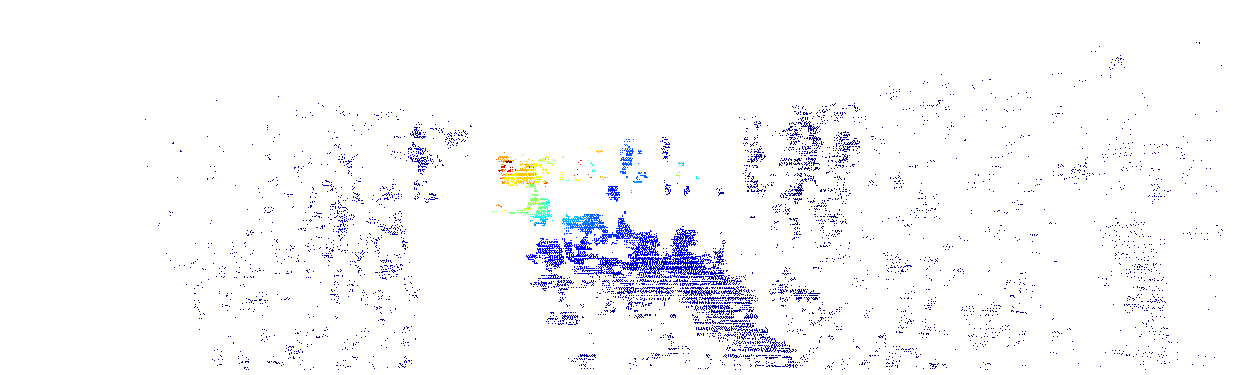} &
            \includegraphics[width=0.9\linewidth,height=1cm]{figures/colormaps/colormaps_E.pdf} \\
            \midrule

            $\gamma$ &
            \includegraphics[width=\linewidth,height=1cm]{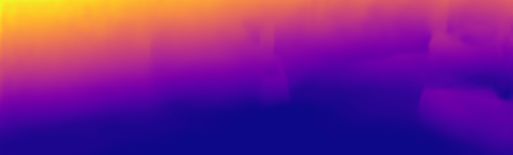} &
            \includegraphics[width=\linewidth,height=1cm]{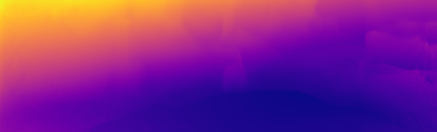} &
            \includegraphics[width=\linewidth,height=1cm]{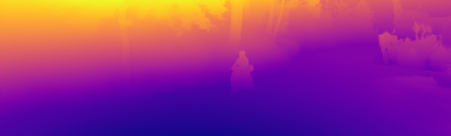} &
            \includegraphics[width=0.9\linewidth,height=1cm]{figures/colormaps/new_colormap.pdf} \\

            $\gamma$ error &
            \includegraphics[width=\linewidth,height=1cm]{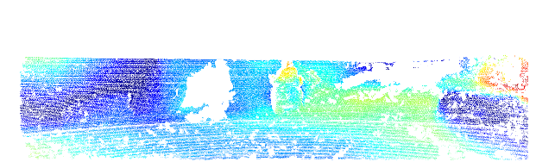} &
            \includegraphics[width=\linewidth,height=1cm]{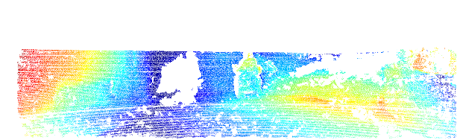} &
            \includegraphics[width=\linewidth,height=1cm]{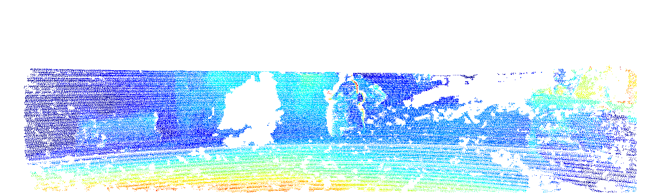} &
            \includegraphics[width=0.9\linewidth,height=1cm]{figures/colormaps/colormaps_E.pdf} \\

            Depth &
            \includegraphics[width=\linewidth,height=1cm]{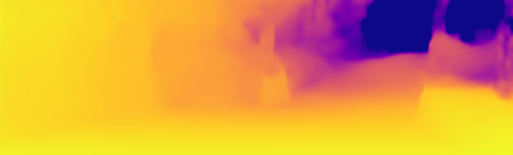} &
            \includegraphics[width=\linewidth,height=1cm]{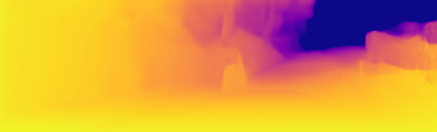} &
            \includegraphics[width=\linewidth,height=1cm]{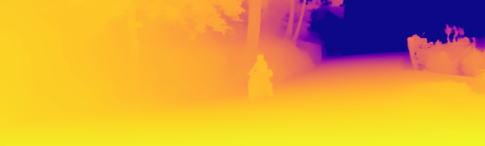} &
            \includegraphics[width=0.9\linewidth,height=1cm]{figures/colormaps/colormaps_D.pdf} \\

            Depth error &
            \includegraphics[width=\linewidth,height=1cm]{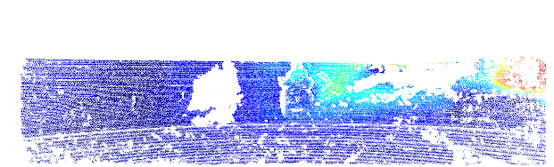} &
            \includegraphics[width=\linewidth,height=1cm]{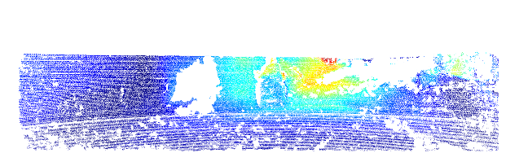} &
            \includegraphics[width=\linewidth,height=1cm]{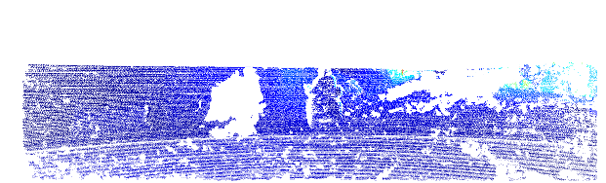} &
            \includegraphics[width=0.9\linewidth,height=1cm]{figures/colormaps/colormaps_E.pdf} \\
            \bottomrule
        \end{tabular}
    \end{minipage}

    \caption{%
        \textbf{Qualitative comparison for two examples from KITTI.}
        \textbf{Left:} (a)~Input (Example~1) and (b)~Input (Example~2). 
        \textbf{Right:} Comparison of our method (GfM), GroCo~\cite{GroCo2024}, 
        and DepthPro~\cite{DepthPro2024} for each metric, with the utilized colormap 
        shown in the last column.
    }
    \label{fig:qualtiative_comparison_appendix}
\end{figure*}

\begin{figure*}[t]
    \centering
    \renewcommand{\arraystretch}{1.2}
    \setlength{\tabcolsep}{2pt}
    \begin{tabular}{ccc}
        \textbf{Input Image} & \textbf{Point Cloud (GfM)} & \textbf{Point Cloud (DepthPro)} \\
        
        \includegraphics[width=0.30\linewidth]{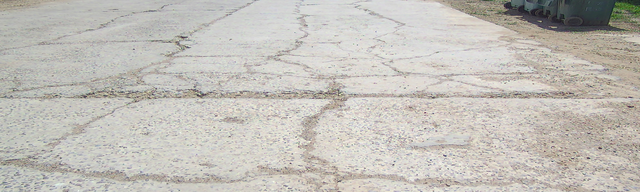} &
        \includegraphics[width=0.30\linewidth]{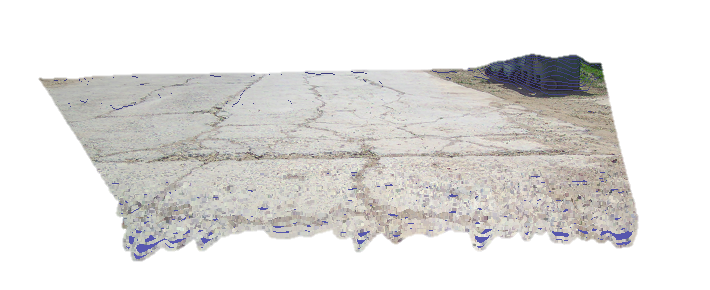} &
        \includegraphics[width=0.30\linewidth]{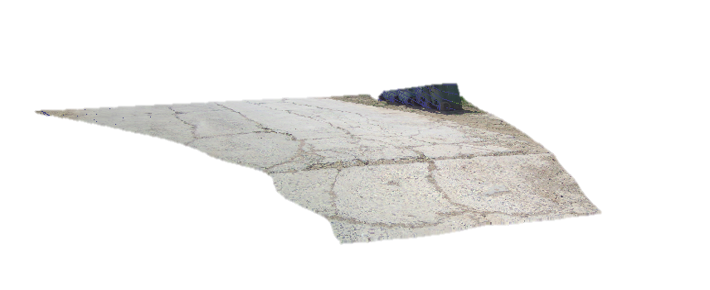} \\

        \includegraphics[width=0.30\linewidth]{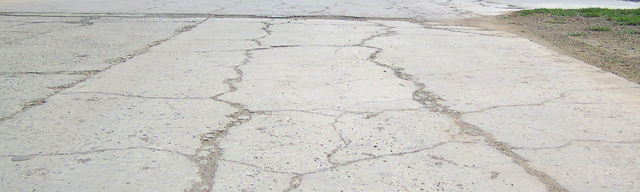} &
        \includegraphics[width=0.30\linewidth]{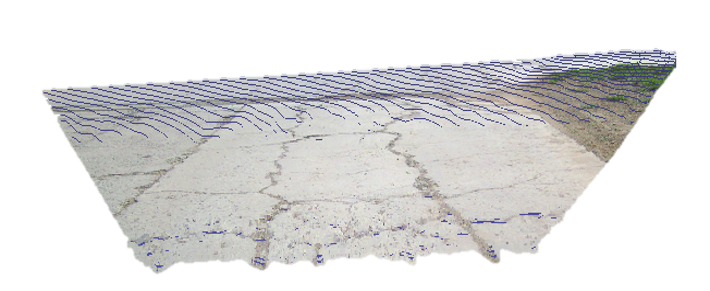} &
        \includegraphics[width=0.30\linewidth]{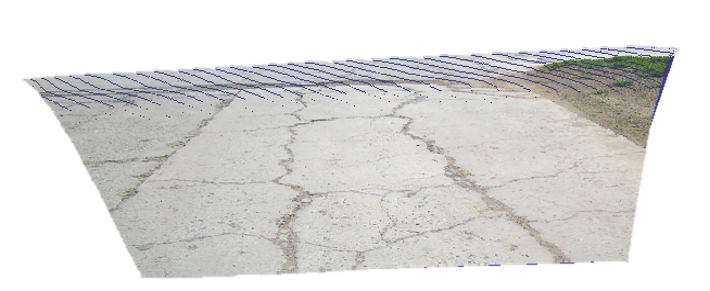} \\

        \includegraphics[width=0.30\linewidth]{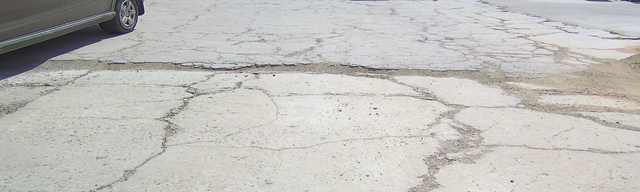} &
        \includegraphics[width=0.30\linewidth]{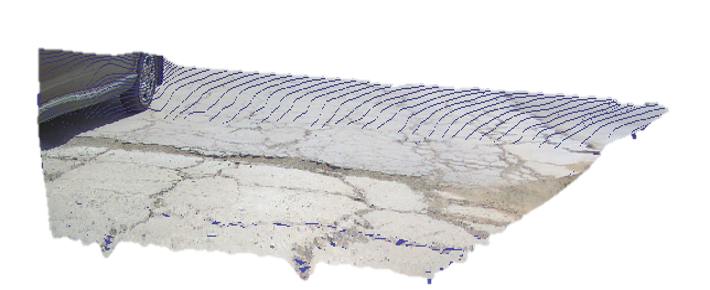} &
        \includegraphics[width=0.30\linewidth]{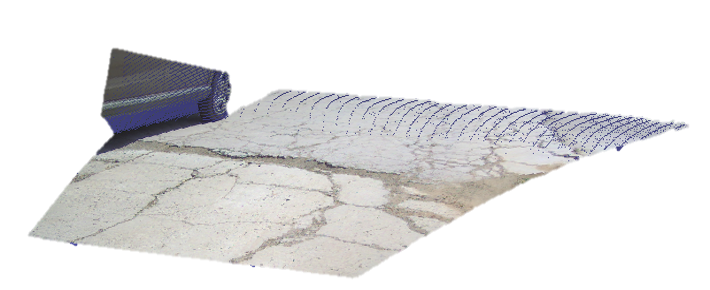} \\

        \includegraphics[width=0.30\linewidth]{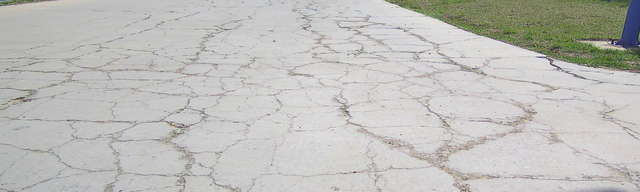} &
        \includegraphics[width=0.30\linewidth]{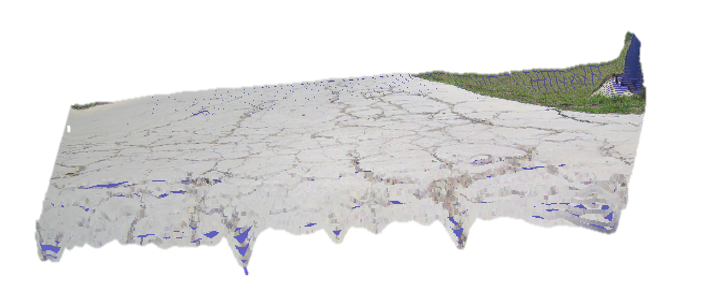} &
        \includegraphics[width=0.30\linewidth]{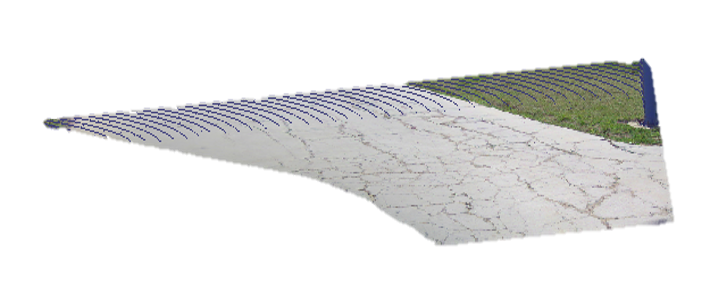} \\

        \includegraphics[width=0.30\linewidth]{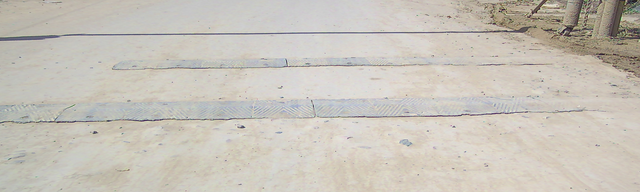} &
        \includegraphics[width=0.30\linewidth]{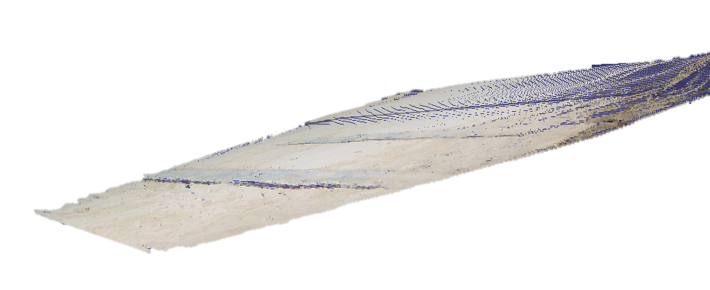} &
        \includegraphics[width=0.30\linewidth]{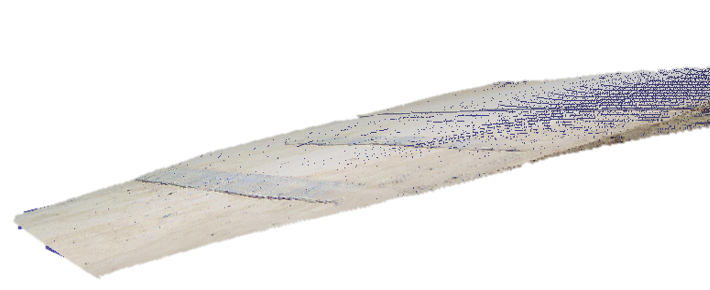} \\

        \includegraphics[width=0.30\linewidth]{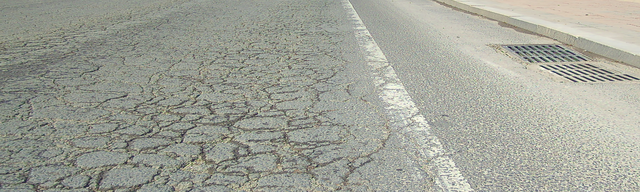} &
        \includegraphics[width=0.30\linewidth]{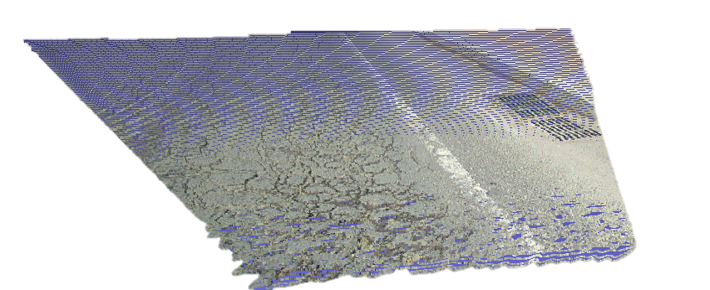} &
        \includegraphics[width=0.30\linewidth]{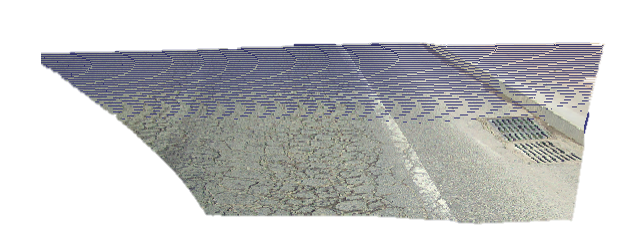} \\
    \end{tabular}
    
    \caption{
    Qualitative comparison of 3D reconstructions on the RSRD dataset test set. 
    Each row shows (left) the input image, (middle) the point cloud generated by GfM, and (right) the point cloud from DepthPro. 
    Our method preserves road geometry better, while DepthPro exhibits noticeable misinterpretations of road slopes.
    }
    \label{fig:rsrd_pointcloud_comparison_appendix}
\end{figure*}


